\documentclass{article}
\usepackage{iclr2021_workshop,times}

\usepackage{hyperref}
\usepackage{url}
\usepackage{amssymb,amsmath}
\usepackage{bbm}
\usepackage{amsthm}
\usepackage{amsfonts}
\usepackage{booktabs}
\usepackage{multirow}
\usepackage{makecell}
\usepackage{subcaption}
\usepackage{graphicx}

\title{Atlas Based Representation and Metric Learning on Manifolds}

\author{Eric O.~Korman \\
\texttt{e.korman@striveworks.us} \\
}

\DeclareMathOperator*{\argmax}{argmax}

\newcommand{\R}{\mathbb R}
\newcommand{\E}{\mathbb E}
\newcommand{\M}{\mathcal M}
\newcommand{\X}{\mathcal X}
\newcommand{\Z}{\mathcal Z}
\newcommand{\U}{\mathcal U}
\newcommand{\J}{\mathcal J}
\newcommand{\A}{\texttt A}
\newcommand{\MA}{\texttt{MA}}
\newcommand{\MMD}{\operatorname{MMD}}

\newcommand{\valstd}[2]{$#1 {\scriptstyle \,\pm\, #2}$}

\newtheorem{prop}{Proposition}

\usepackage{tikz}
\usetikzlibrary{shapes.geometric, arrows, automata, fit, positioning}

\tikzstyle{tnode} = [rectangle, rounded corners, minimum width=.75cm, minimum height=.75cm,text centered, draw=black]
\tikzstyle{node} = [circle,  minimum width=.2cm, minimum height=.2cm, draw=black]
\tikzstyle{arrow} = [thick,->,>=stealth]
\tikzstyle{edge} = [thick]
\tikzstyle{dashedarrow} = [dashed,->,>=stealth]

\definecolor{g}{RGB}{109,144,79}
\definecolor{b}{RGB}{52,138,189}
\definecolor{r}{RGB}{226,74,51}

\iclrfinalcopy
\begin{document}

\maketitle

\begin{abstract}
We explore the use of a topological manifold, represented as a collection of charts, as the target space of neural network based representation learning tasks. This is achieved by a simple adjustment to the output of an encoder's network architecture plus the addition of a maximal mean discrepancy (MMD) based loss function for regularization. Most algorithms in representation and metric learning are easily adaptable to our framework and we demonstrate its effectiveness by adjusting SimCLR (for representation learning) and standard triplet loss training (for metric learning) to have manifold encoding spaces.  Our experiments show that we obtain a substantial performance boost over the baseline for low dimensional encodings. In the case of triplet training, we also find, independent of the manifold setup, that the MMD loss alone (i.e. keeping a flat, euclidean target space but using an MMD loss to regularize it) increases performance over the baseline in the typical, high-dimensional Euclidean target spaces. Code for reproducing experiments is provided at \url{https://github.com/ekorman/neurve}.
\end{abstract}

\section{Introduction}
Representation learning algorithms typically produce encodings into a Euclidean space. However, if the \textit{manifold hypothesis} (the assumption that the data is well-approximated by a manifold) is taken seriously then a Euclidean target space is unnecessarily big for encoding this manifold: by the Whitney embedding theorem, to embed an $n$-dimensional manifold into a Euclidean space one may need up to $2n$ dimensions for the ambient space. The situation is even worse if the embedding is required to preserve distances (which is obviously important for metric learning), as the Nash embedding theorem shows that in this case the dimension of the ambient space needed is $\mathcal O(n^2)$ \cite{andrews2002notes}. Additionally, such algorithms seldom regularize the encoding space (i.e. encourage it to look like a prior distribution) to ensure that it is well-behaved.

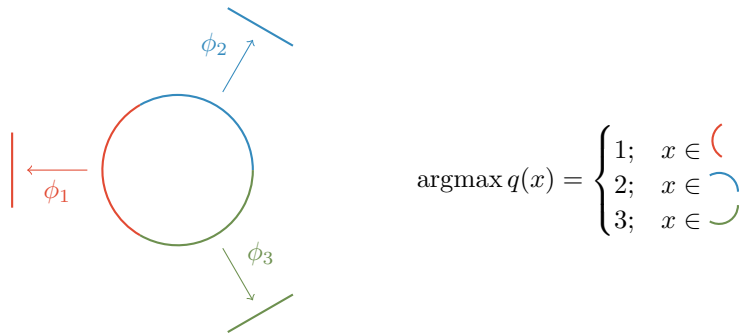
\begin{figure}[h]
\centering
\begin{subfigure}{.4\textwidth}
\centering
\begin{tikzpicture}

\draw[b, thick] (1, 0) arc (0:120:1);
\draw[r, thick] (-.5, .866025) arc (120:240:1);
\draw[g, thick] (-.5, -.866025) arc (240:360:1);

 \draw[r, ->] (-1.2, 0) -- (-2, 0) 
        node[pos=0.5, auto=left] {$\phi_1$};

 \draw[b, ->] (0.6, 1.03923048)  -- (1, 1.73205081) 
        node[pos=0.5, auto=left] {$\phi_2$};

 \draw[g, ->] (0.6, -1.03923048)  -- (1, -1.73205081) 
        node[pos=0.5, auto=left] {$\phi_3$};
        
\draw[r, thick] (-2.2, -.5) -- (-2.2, .5);
\draw[b, thick] (0.6669873 , 2.15525589) -- (1.5330127 , 1.65525589);
\draw[g, thick] (0.6669873 , -2.15525589) -- (1.5330127 , -1.65525589);
        
\end{tikzpicture}
\end{subfigure} \begin{subfigure}{.4\textwidth}
\[
\argmax q(x) = \begin{cases}
1; &x \in \begin{tikzpicture}
\draw[r, thick](-.5, .866025) arc (120:240:.25);
\end{tikzpicture} \\
2; &x \in \begin{tikzpicture}
\draw[b, thick] (1, 0) arc (0:120:.25);
\end{tikzpicture} \\
3; &x \in \begin{tikzpicture}
\draw[g, thick] (-.5, -.866025) arc (240:360:.25);
\end{tikzpicture}
\end{cases}
\]
\end{subfigure}
\caption{Example of a learned atlas (with $d=1$) of a circle. Instead of a single encoder function, we learn $n=3$ many: $\phi_1, \phi_2$ and $\phi_3$, each a map $\X \to [0, 1]$. For a given point $x\in\X$, the learned function $q: \X \to [0,1]^3$ ranks which of the $\phi_i$ to use.}
\label{S1}
\end{figure}

In this work we develop a framework for using a manifold as a target space of deep representation learning algorithms. Following our prior work \citep{korman2018autoencoding}, which uses manifolds as the latent space of an autoencoder, this is done by having the encoding network output chart embeddings and membership probabilities for an atlas of a manifold. In other words, instead of learning a single encoder $f: \X \to \R^d$ (where $\X$ is the input data and $\R^d$ is the Euclidean encoding space), our technique learns $n$-such encoders together with a scoring function $q: \X \to [0, 1]^n$ that determines which encoding output to use for a given input. Figure \ref{S1} shows such an example for $\X$ a circle. For matching the distribution of embeddings to a manifold prior, we introduce a maximal mean discrepancy (MMD) \citep{gretton2012kernel} loss for manifolds.

Our framework is flexible enough to apply to a variety of representation and deep metric learning algorithms and in this paper we do experiments with manifold generalizations of SimCLR \cite{chen2020simple} for representation learning and standard triplet training \cite{schroff2015facenet} for metric learning. Our results show that, in low dimensions, replacing the Euclidean target space of these algorithms with a manifold gives a significant improvement in the learned embeddings. Besides being of theoretical interest, low dimensional embeddings have the practical benefits of having faster distance computations, requiring less storage, providing visualizations (in the case of dimensions 2 or 3) and avoiding issues with curse of dimensionality for downstream tasks from the embedding (such as clustering). 

In metric learning, we additionally find that the MMD regularization loss alone provides a significant improvement over the baseline in the typical high-dimensional euclidean encoding spaces used for metric learning.

We note that this is extended form of our work \cite{korman2021selfsupervised}, which does not include our metric learning experiments.

Our main contributions are as follows:
\begin{itemize}
\item In section \ref{sec-manifolds} we describe the general procedure for how to adjust typical deep representation and metric learning algorithms to have a manifold as a target space. We define a (semi)-metric on this space and a two-term regularization loss that encourages the encoding distribution to be a uniform and efficient atlas.

\item In section \ref{section-MSimCLR} we show how to adapt SimCLR to encode into a manifold. We validate our approach with experiments on MNIST \cite{lecun2010mnist}, FashionMNIST \cite{xiao2017/online}, and CIFAR10 \cite{krizhevsky2009learning}. Our results show that in small dimensions (we do experiments with dimensions 2, 4, and 8), we get significant improvements over the baseline at the cost of only a few extra \textit{bits} of information (namely the chart number) in the representation space.

\item In section \ref{section-MTriplet} we show how to adapt standard triplet loss training to have a manifold as encoding space. We run experiments on the common benchmark datasets CUB-200 \cite{welinder2010caltech} and Stanford Cars \cite{krause20133d}. As in the case of SimCLR, we also see a significant improvement over the baseline in small dimensions. While, unlike in the SimCLR results, our performance likely does not close the gap enough with the results in high-dimensions to be worth the performance tradeoff in a practical setting, we find these results of theoretical interest and validating the potential of our general framework. Furthermore, we also do an experiment in dimension 64 (a common choice in the literature) and find that keeping the target space Euclidean but supplementing the triplet loss with using the (completely unsupervised) regularizing MMD loss yields a substantial improvement in performance compared to no regularization.
\end{itemize}

\subsection{Related work}
\noindent{\bf Learning an atlas}
The idea of learning a manifold as an atlas was discussed in \cite{pitelis2013learning} where they learn a collection of linear charts via a generalization of principal component analysis. The recent work \cite{kohli2021ldle}, which operates in the non-parametric setting, uses Laplacian eigenmaps to construct local charts and then stitches them together to form a single representation space. Our earlier work \cite{korman2018autoencoding}, which we build upon, learns an atlas as the latent space of an autoencoder. Another difference is that in that work the regularization of the latent space is via adversarial training instead of the MMD loss we use in this paper.

\noindent{\bf Encoding space regularization}
Regularization of a latent space is typically only done in algorithms that have a decoder/generator aspect, such as variational autoencoders \cite{kingma2013auto} and adversarial autoencoders (AAE) \cite{makhzani2015adversarial}. The effectiveness of using an MMD loss in particular on such an encoding space is used in  \cite{tolstikhin2017wasserstein} and inspired our MMD loss function. The work of \cite{grattarola2019adversarial} is of similar spirit to ours, as they extend the latent spaces of adversarial autoencoders to more geometrically interesting ones, namely constant curvature Riemannian manifolds.

\noindent{\bf Metric learning}
A variety of techniques improve upon the foundational triplet training approach to metric learning established in \cite{schroff2015facenet}. These techniques typically adjust the loss function in some way, whether by focusing on triplet sampling methods \cite{schroff2015facenet, harwood2017smart, xuan2020improved}, using proxies as elements of a triplet \cite{movshovitz2017no, teh2020proxynca++} or approximating the triplet loss \cite{do2019theoretically}. There have also been non-triplet based lossed functions \cite{ustinova2016learning, sohn2016improved, liu2017sphereface, zhai2018classification} and ensemble methods \cite{kim2018attention, xuan2018deep}. We refer to \cite{musgrave2020metric} and references therein for a broader overview of current approaches.

Our work, in contrast, comes from the different direction of changing the shape of and regularizing the encoding space. As far as we know there have not been techniques that try to control the embedding distribution (in a similar way to how the latent space of an autoencoder is regularized) in metric learning via an unsupervised regularization term.

Though for simplicity we supplement standard triplet training in this paper, many of the aforementioned algorithms could be supplemented with our technique, and we leave it to future work to investigate the effectiveness of this. 

\noindent{\bf Representation learning}
SimCLR, which we experiment with, is just one of many recent advancements in self-supervised representation learning. Other such algorithms include AMDIM \cite{bachman2019learning}, MoCo(v2) \cite{he2019moco, chen2020mocov2}, and BYOL \cite{grill2020bootstrap}. We leave to future work the results of adapting these to have manifold output.

\section{Manifolds as encoding spaces} \label{sec-manifolds}
\begin{figure}[h]
\centering
\begin{tikzpicture}[scale=0.7]
    \foreach \x in {90,89,...,-90} { 
    \pgfmathsetmacro\elrad{20*max(cos(\x),.1)}
    \pgfmathsetmacro\ltint{.9*abs(\x-45)/180}
    \pgfmathsetmacro\rtint{.9*(1-abs(\x+45)/180)}
    \definecolor{currentcolor}{rgb}{0, 0, \ltint}
    \draw[color=currentcolor,fill=currentcolor] 
        (xyz polar cs:angle=\x,y radius=.75,x radius=1.5) 
        ellipse (\elrad pt and 20pt);
    \definecolor{currentcolor}{rgb}{0, 0, \rtint}
    
    \draw[color=currentcolor,fill=currentcolor] 
        (xyz polar cs:angle=180-\x,radius=.75,x radius=1.5) 
        ellipse (\elrad pt and 20pt);
    }
    
    \draw[fill=cyan] (-6, -1) rectangle (-4, 0);
    
    \draw[->] (-5, -1.2) .. controls (-4.5, -2.5) and (0, -2.5) .. (0, -1.6) 
        node[pos=0.5, auto=left] {$\psi$};
   
    \pgfmathsetmacro\elrad{cos(-135)}
    \pgfmathsetmacro\xrad{1.5cm-20pt*\elrad}
    \pgfmathsetmacro\yrad{.75cm-20pt*sin(-135)}
    \path[fill=cyan, fill opacity=.35] 
        (xyz polar cs:angle=-135,radius=.75,x radius=1.5) 
        ++(20pt*\elrad,0) arc (0:45:20*\elrad pt and 20pt) 
        arc (-135:-45:\xrad pt and \yrad pt) 
        arc (45:-45:-20*\elrad pt and 20pt) 
        arc (-45:-135:\xrad pt and \yrad pt) 
        arc (-45:0:20*\elrad pt and 20pt);
\end{tikzpicture}
\caption{A singe chart on the torus}
\label{torus-chart}
\end{figure}
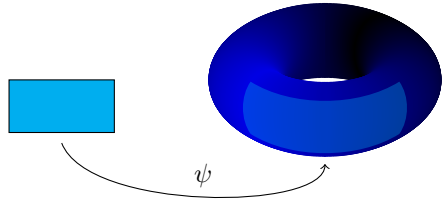

Loosely, a $d$-manifold is a space $\M$ that locally looks like $\R^d$. This notion is made rigorous by requiring the existence of \textit{charts}, which are are homeomorphisms of an open subset of $\R^d$ to an open subset of $\M$ such that every point in the manifold is contained in some chart (such a collection of charts is called an \textit{atlas}). Figure \ref{torus-chart} shows an example of a single chart on a 2-manifold.

\subsection{Representing a dataset as a manifold} \label{data-manifold}
For formally modeling a distribution of data, $\X$, as a manifold,  we use the same approach as in our earlier work \citep{korman2018autoencoding}. Namely we posit the existence of a latent space $\Z = [0, 1]^d \times \{1, \ldots, n\}$ of $n$, $d$-dimensional charts with coordinate maps $\psi_i : [0, 1]^d \to \X$ that forms an atlas of a manifold. We use the uniform distribution as the prior on $\Z$ and we let $X, Z, J$ denote the random variables on $\X, [0,1]^d,$ and $\{1, \ldots, n\}$, respectively. We will denote by $\mathbbm 1_y$ the distribution supported at a single point $y$. In our decoder-free setup, we wish to learn:

\begin{enumerate}
\item The inverse mappings of the $\psi_i$, which we denote by $\phi_i : \X\to [0,1]^d$ and which satisfy 
\[
p(z | J=i, X=x) = \mathbbm 1_{\phi_i(x)}.
\]
\item The chart membership function $q = (q_1, \ldots, q_n): \X \to [0, 1]^n$ defined by 
\begin{equation}
q(x) = \left(p(J = 1 \mid X=x), \ldots, p(J = n \mid X=x)\right). \label{q}
\end{equation}
\end{enumerate}
We can then compute the posterior in terms of $q$ and the $\phi_i$ as 
\[
p(z, j \mid x) = p(z \mid j, x) p(j \mid x) = q_j(x) \mathbbm 1_{\phi_j(x)}
\]
which gives the prior on $\Z$ as
\begin{equation}
p(z, j) = \E_x q_j(x) \mathbbm1_{\phi_j(x)}, \label{pushforward-distribution}
\end{equation}
which we wish to be uniform.

An additional desire is that we have an efficient atlas in the sense that any point $x$ should be in as few charts as possible. Thus while $p(J)$ should be uniform, we want the conditional distributions $p(J \mid X=x)$ to have low entropy: if the distributions $p(J \mid X=x)$  are mostly deterministic then encoding $x$ requires us to only keep the coordinates and chart number for the chart with highest probability for $x$. In other words, for the representation of $x$ at inference time we take
\begin{equation}
x \mapsto (\phi_i(x), i) \in \R^d \times \{1, \ldots, n\}, ~~ \text{where } i = \argmax_j q_j(x), \label{compressed-rep}
\end{equation}
which has just $\log n$-more bits of information than the Euclidean case.

\subsection{A Maximal Mean Discrepancy Loss for Manifolds}
For a given embedding task, we propose to parameterize the functions $\{\phi_1, \ldots, \phi_n, q\}$ using neural networks and optimize the parameters via gradient descent for a loss function consisting of a task-specific term (e.g. a contrastive loss function) plus a regularization term that encourages the distribution $p(z, j)$ given by (\ref{pushforward-distribution}) to be close to uniform and the discrete distribution $q(x)$ in (\ref{q}) to be close to a deterministic distribution for each $x \in \X$.

In the Euclidean case (i.e. when $n=1$), there are two popular ways for regularizing the latent space to match a prior distribution: using adversarial training \citep{makhzani2015adversarial, tolstikhin2017wasserstein} or via an MMD \citep{gretton2012kernel} loss \citep{tolstikhin2017wasserstein}. In \citet{korman2018autoencoding} we used an adversarial loss for a manifold latent space but in this work we use an MMD loss due to better training stability and less hyperparameters to tune. If $P$ and $Q$ are two distributions on a common space and $k$ is a reproducing kernel, then the MMD gives a measure of the difference between the distributions, and is defined by
\begin{equation}
\MMD_k(P, Q)^2 = \E_{y_1, y_2 \sim P\times P} k(y_1, y_2) - 2 \E_{y_1, y_2 \sim P\times Q} k(y_1, y_2) + \E_{y_1, y_2 \sim Q \times Q} k(y_1, y_2). \label{mmd-def}
\end{equation}

Let  $p$ and $q$ be given by (\ref{pushforward-distribution}) and (\ref{q}), respectively, $\U_\Z$ denote the uniform distribution on $\Z$, and $\U_{\mathcal J}$ denote the uniform distribution on $\J = \{1,\ldots, n\}$. For the loss function encouraging $p$ to be close to $\U_\Z$ we will take an approximation of $\MMD_{k_\Z}(p, \U_\Z)^2$ and for the loss function encouraging $q(x)$ to be far from $\U_{\mathcal J}$, we take $-\E_x \MMD_{k_\J}(q(x),  \U_\J)$. This will achieve the goal of encouraging $p$ to be uniform and for $q(x)$ to be deterministic.

To get a kernel $k_\Z$ on $\Z$ we can start with a kernel  $k_0$ on $[0, 1]^d$ and then define
\[
k_\Z: \Z \times \Z \to \R, ~ ((z_1, i), (z_2, j)) \mapsto \delta_{ij} k_0(z_1, z_2)
\]
where $\delta_{ii} = 1, \delta_{ij} = 0$ if $i\ne j$.  For $k_0$ we take, as in \citet{tolstikhin2017wasserstein}, the inverse multiquadratics kernel  but pulled back via the sigmoid function\footnote{this avoids having to compute the final sigmoid activation.}: $k_0(x,y)=\frac{d/6}{d/6 + |\sigma^{-1}(x) - \sigma^{-1}(y)|^2}$. We approximate $\MMD_{k_\Z}(p, \U_\Z)^2$ using the U-statistic estimator in \citet{gretton2012kernel}, adjusted to our manifold setting. Explicitly:

\begin{prop} \label{prop-mmd}
Let $p$ be the distribution on $\Z$ defined by (\ref{pushforward-distribution}) and let $\U_\Z$ denote the uniform distribution on $\Z$. Given a random sample $\{x_1, \ldots,  x_N\}$ of $\X$ and a random sample $\{w_1, \ldots, w_N\}$ drawn uniformly from $[0,1]^d$, an estimator for $\MMD_{k_\Z}(p, \U_\Z)^2$ is:
\begin{align}
\ell_\Z (q, \phi_1, \ldots, \phi_n) &= \frac{1}{N(N-1)} \sum_{\substack{j, k = 1 \\ j \ne k}}^N \sum_{i=1}^n q_i(x_j) q_i(x_k) k_0(\phi_i(x_j), \phi_i(x_k)) \label{ellZ-def} \\
& - \frac{2}{n N^2} \sum_{j, k = 1}^N \sum_{i=1}^n q_i(x_j) k_0(\phi_i(x_j), w_k) +  \frac{1}{nN(N-1)} \sum_{\substack{j, k = 1 \\ j \ne k}}^N k_0(w_j, w_k). \nonumber
\end{align}
\end{prop}
See section \ref{prop-proof} for the proof.

For $\E_x \MMD_{k_\J}(q(x),  \U_\J)$ we take the kernel $k_\J :  \{1, \ldots, n\} \times \{1, \ldots, n\} \to \R, (i, j) \mapsto \delta_{ij}$ and define
\begin{align*}
\ell_\J(q)  &:= -\E_x \MMD_{k_\J}(q(x),  \U_\J) = -\E_x \sum_{i=1}^n \left(q_i(x) - \frac{1}{n}\right)^2.
\end{align*}
The total regularization loss is then
\begin{equation}
\ell_{reg}(q, \phi_1, \ldots, \phi_n) = \lambda_1 \ell_\Z (q, \phi_1, \ldots, \phi_n) + \lambda_2 \ell_\J(q) \label{reg-loss}
\end{equation}
for some hyperparameters $\lambda_1, \lambda_2 \in [0,\infty)$.

\subsection{Distance function on a manifold}
To define a metric on the manifold encoding space (which is necessary for some representation learning algorithms), we start with a metric $d_0$ on $[0,1]^d$ and extend it to $\M$ via
\begin{equation}
d_\M(x, y) = \begin{cases}
\frac{\sum\limits_i q_i(x) q_i(y) d_0(\phi_i(x), \phi_i(y))}{\sum\limits_i q_i(x) q_i(y)}; & \sum\limits_i q_i(x) q_i(y) \ne 0\\
\infty; &\text{else}
\end{cases} \label{manifold-metric}
\end{equation} 
We note that technically, $d_M$ is a semi-metric since the triangle inequality may fail.

\subsection{Summary of our framework} \label{sec-framework}
We now describe our general technique of turning a deep representation learning algorithm to one that has a manifold encoding space. The typical setup for such an algorithm, $\A$, is to learn an encoder $f$ mapping the dataset $\X$ to the space $\R^d$ via optimizing some loss function $\ell_\A$ defined on embeddings of a mini-batch $\{x_1, \ldots, x_N\} \subset \X$.

To adjust the algorithm to one that has a manifold as the encoding space, we see from the discussion in \ref{data-manifold} that $f$ should be replaced by a collection of maps $\phi_1,\ldots,\phi_n, q$ where $\phi_i : \X \to [0, 1]^d$ is the $i$\textsuperscript{th} coordinate map and $q: \X\to[0,1]^n$ is the chart membership function. In practice, for the functions $\phi_1,\ldots,\phi_n, q$, we take a backbone network and attach $n + 1$ linear heads followed by a sigmoid activation on the first $n$ (which have output in $\R^d$) and a softmax activation on the head defining $q$ (which has output in $\R^n$).

In many contrastive representation learning algorithms (such as SimCLR \citep{chen2020simple}, MoCo v2 \citep{chen2020mocov2}, and BYOL \citep{grill2020bootstrap}), the loss $\ell_\A$ is a function of an auxiliary projection head $h: \R^d \to \R^{\tilde d}$ applied to the embedding vectors. In these cases, for the manifold version we use $n$-many projection heads (one for each coordinate chart), and then produce a single projection vector for every data point by taking a sum of these individual projection vectors weighted by $q$. We supplement the resulting loss $\ell_\MA$ with the regularization loss (\ref{reg-loss}).

At inference and evaluation time, we use the compressed representation (\ref{compressed-rep}). This ensures that the representation is indeed $d$-dimensional and gives a fair evaluation comparison to $\A$, which embeds into a $d$-dimensional Euclidean space. For example, if $q_i(x)$ were the uniform distribution then using the full-representation $(\phi_1(x), \ldots, \phi_n(x))$ instead of the compressed one would essentially be ``cheating" into an $nd$-dimensional Euclidean representation.

We summarize this procedure in Table \ref{summary-tables} and make note of the following special cases:
\begin{enumerate}
\item If $n = 1$ and $\lambda_1 = 0$ then this corresponds to just the algorithm $\A$ except that $f$ is forced to have range $[0, 1]^d$ (e.g. by applying a sigmoid activation at the end).

\item If $n = 1$ and $\lambda_1 > 0$ then the embedding space is still Euclidean like the previous case, but through the regularization loss the distribution of embeddings is encouraged to be close to the uniform distribution on $[0,1]^d$.

\item The case of $d=2$ yields a powerful data visualization by plotting the input data at their embedding coordinates for their most probable chart. We show such visualizations in the appendix \ref{sec-visualizations}.
\end{enumerate}

\begin{table*}[h]
\caption{Summary of our framework for elevating an algorithm $\A$ to have a manifold encoding space.}
\label{summary-tables}
\centering
\begin{small}
\begin{tabular}{c c}
\toprule
{\sc TYPICAL SETUP} & {\sc MANIFOLD VERSION} \\
\midrule

\makecell{learn a neural network encoder \\ $f : \X \to \R^d$} & \makecell{learn a neural network encoder with $n + 1$ heads \\ $f = (\phi_1,\ldots,\phi_n, q):\X\to [0, 1]^d \times \cdots \times [0, 1]^d \times [0, 1]^n$} \\\\

\makecell{euclidean distance function} & \makecell{metric $d_\M$ as defined in (\ref{manifold-metric})} \\\\

\makecell{auxiliary projection head $h: \R^d\to \R^{\tilde d}$ \\ giving projection map $h \circ f: \X \to \R^{\tilde d}$.} & \makecell{$n$ auxiliary projection heads $h_1, \ldots, h_n: \R^d\to \R^{\tilde d}$ \\ giving projection map $\X \to \R^{\tilde d}, ~ x\mapsto \sum_i q_i(x) h_i(\phi_i(x))$.} \\\\

\makecell{on a minibatch $\{x_1,\ldots,x_N\} \subset \X$ \\ optimize a loss $\ell_{\texttt A}(f(x_1), \ldots, f(x_N))$.} & \makecell{on a minibatch $\{x_1,\ldots,x_N\} \subset \X$  optimize a loss \\ $\ell_\MA(f(x_1), \ldots, f(x_N)) + \ell_{reg}(f(x_1), \ldots, f(x_N))$.} \\\\

\makecell{Representation of $x\in X$ at inference/evaluation: \\ $f(x) \in \R^d$.} & \makecell{Representation of $x\in X$ at inference/evaluation: \\ $(\phi_i(x), i) \in \R^d \times \{1, \ldots, n\}$, where $i = \argmax_j q_j(x)$.} 

\end{tabular}
\end{small}
\end{table*}

\section{Applications}
\subsection{MSimCLR} \label{section-MSimCLR}
In this section we generalize SimCLR \cite{chen2020simple} to have a manifold as embedding space and denote the resulting technique by \textit{MSimCLR}. We recall that SimCLR trains a neural network encoder $f: \X \to \R^d$ using a projection head $h: \R^d \to \R^{\tilde d}$. A mini-batch is formed by choosing $N$-images and augmenting in two different ways, producing examples $\{x_1, \ldots, x_{2N}\}$. The loss function over this batch is a function of the projection head output of the embeddings of these images:
\[
\ell_{\text{SimCLR}}(x_1, \ldots, x_n) = c(h(f(x_1)), \ldots, h(f(x_{2n}))),
\]
where $c$ is a contrastive loss. We follow our meta-procedure from the previous section to adjust SimCLR \citep{chen2020simple} to have a manifold as encoding space (with resulting algorithm denoted by \textit{MSimCLR}) and run experiments on MNIST \citep{lecun2010mnist}, FashionMNIST \citep{xiao2017/online}, and CIFAR10 \citep{krizhevsky2009learning} for $n \in \{1, 4, 16, 32\}$ and $d \in \{2, 4, 8\}$.

In \citet{chen2020simple} representations of SimCLR are evaluated based on the accuracy of a linear classifier trained on top of the representation. In our case, since we have a collection of charts, we evaluate our manifold representation by putting a linear classifier on each chart. We report the mean and standard deviation of the accuracy of our models on the hold out test sets in Table \ref{results-table}, from which we see that our method provides a significant performance boost over vanilla SimCLR, especially in dimensions two and four. Section \ref{sec-setup} outlines the details of our experiments, including hyperparameter selection. 

\begingroup
\setlength{\tabcolsep}{2.3pt}
\begin{table*}[h]
\caption{Piecewise linear evaluation accuracy (* denotes no convergence)} \label{results-table}
\centering
\begin{scriptsize}
\begin{tabular}{l c c c c c c c c c c}
\toprule
{\sc \tiny METHOD}  & {\sc \tiny \# CHARTS} & \multicolumn{9}{c}{\sc \tiny DATASET} \\
\cmidrule(l){3-11}
& & \multicolumn{3}{c}{\sc \tiny MNIST} & \multicolumn{3}{c}{\sc \tiny Fashion MNIST} & \multicolumn{3}{c}{\sc \tiny CIFAR10} \\
\cmidrule(l){3-5} \cmidrule(l){6-8} \cmidrule(l){9-11}
& & \multicolumn{3}{c}{\sc \tiny ENCODING DIMENSION} & \multicolumn{3}{c}{\sc \tiny ENCODING DIMENSION}  & \multicolumn{3}{c}{\sc \tiny ENCODING DIMENSION} \\
 & & 2 & 4 & 8 & 2 & 4 & 8 & 2 & 4 & 8 \\
\midrule
SimCLR & - & \valstd{15.3}{6.8} & \valstd{75.4}{2.1} & \valstd{94.5}{1.5} & \valstd{39.8}{9.3} & \valstd{62.7}{1.8} & \valstd{79.3}{0.2}  & * & \valstd{66.2}{0.2} & \valstd{79.6}{0.2} \\
MSimCLR & 1 & \valstd{35.4}{12.4} & \valstd{66.2}{0.5} & \valstd{90.5}{2.0} & \valstd{36.1}{10.8} & \valstd{59.1}{4.9} & \valstd{73.5}{3.9} & \valstd{30.2}{10.8} & \valstd{61.1}{2.4} & \valstd{78.0}{1.7} \\
MSimCLR & 4 & \valstd{75.1}{3.8} & \valstd{89.0}{1.7} & \valstd{94.1}{2.6} & \valstd{62.5}{1.2} & \valstd{71.9}{1.5} & \valstd{79.0}{0.3} & \valstd{54.5}{5.8} & \valstd{68.2}{4.5} & \valstd{81.0}{1.9}   \\
MSimCLR & 16 & \valstd{90.4}{3.0} & \valstd{94.3}{3.5} & \valstd{97.1}{0.2} & \valstd{68.5}{2.2} & \valstd{73.9}{1.0} & \valstd{80.8}{0.1} & \valstd{74.2}{0.8} & \valstd{73.0}{2.0} & \valstd{77.3}{0.5}\\
MSimCLR & 32 & \valstd{91.3}{2.8} & \valstd{96.4}{0.2} & \valstd{96.6}{0.0} & \valstd{72.1}{1.0} & \valstd{72.7}{0.5} & \valstd{74.6}{2.2} & \valstd{71.4}{3.5} & \valstd{64.2}{1.1} & \valstd{73.3}{2.7} \\
\bottomrule
\end{tabular}
\end{scriptsize}
\end{table*}
\endgroup

 \pagebreak

\subsection{MTriplet}\label{section-MTriplet}
In this section we apply our framework from section \ref{sec-framework} to adjust standard triplet training \cite{schroff2015facenet} to have a manifold encoding space. For $d_0$ in \ref{manifold-metric} we use $d_0(x, y) = || \sigma^{-1}(x) - \sigma^{-1}(y)||$, where $|| \cdot ||$ is the usual Euclidean norm, which amounts to computing the Euclidean distance before applying the sigmoid activation in the final chart layers. We denote the resulting technique by \textit{MTriplet}.

\subsubsection{Experiments setup}
We run experiments on the CUB-200 and Stanford Cars datasets using PyTorch. In forming the batches we follow \cite{musgrave2020metric} and sample batches that consist of four images across eight different classes. We use the ``batch-all" strategy and the standard margin choice of 0.2. As is common practice, we start with an ImageNet pretrained GoogLeNet architecture. During training we apply a random resized crop to $227\times227$ and a random horizontal flip with probability 0.5. During evaluation we do a resize to $227\times227$. We use the RMSProp optimizer (as suggested in  \cite{musgrave2020metric}) with learning rate $10^{-5}$.

For both datasets the standard practice in the literature is to use the first half of the classes for training and the second half as a test set. For doing hyperparameter selection we split the train half further, taking the first 80\% classes as training and the remaining 20\% for validation. Our grid search covered the following options:  $\lambda_1 \in \{0.1, 1, 5, 10, 20\}$ and $\lambda_1 \in \{0.05, 0.1, 0.2\}$ and we set $d = 2, n = 16$. We optimized for recall at one (R@1) and  chose the hyperparameters that gave the best average rank across the two datasets. This yielded $\lambda_1 = 1$ and $\lambda_2 = 0.1$. Using these parameters we train on the entire train dataset across $d \in \{2, 4, 8\}$ and $n \in \{1, 4, 16, 32\}$ for 300 epochs, repeating each experiment three times, and report the mean recalls at 1, 2, 4, and 8 on the test sets in tables \ref{cub-results} and \ref{cars-results}. For the baseline comparison we did standard, normalized triplet training in $\R^{d + 1}$ (i.e., for a fair comparison, we make the target space of the baseline one-dimension higher since the embedding vectors get normalized to the unit sphere, which is $d$-dimensional).

\begingroup
\setlength{\tabcolsep}{4.2pt}
\begin{table*}[h!]
\caption{Results on CUB-200} \label{cub-results}
\vskip 0.15in
\centering
\begin{scriptsize}
\begin{tabular}{c c | c c c c | c c c c | c c c c}
\toprule
{\sc method} & {\sc \# charts} & \multicolumn{4}{c|}{\sc encoding dim 2} & \multicolumn{4}{c|}{\sc encoding dim 4} & \multicolumn{4}{c}{\sc encoding dim 8} \\
\cmidrule(l){3-14}
 &  & R@1 & R@2 & R@4 & R@8 & R@1 & R@2 & R@4 & R@8 & R@1 & R@2 & R@4 & R@8 \\
\midrule
Triplet & & $7.65$ & $13.47$ & $22.40$ & $35.07$ & $14.87$ & $23.08$ & $34.72$ & $48.67$ & $26.93$ & $38.04$ & $50.89$ & $64.84$ \\
MTriplet & 1 & $7.03$ & $12.09$ & $20.12$ & $32.35$ & $14.04$ & $22.20$ & $33.67$ & $48.13$ & $25.24$ & $36.76$ & $49.91$ & $64.02$ \\
MTriplet & 4 & $9.96$ & $17.02$ & $27.13$ & $41.10$ & $17.22$ & $26.94$ & $39.03$ & $53.72$ & $28.67$ & $40.37$ & $53.74$ & $67.22$ \\
MTriplet & 16 & $12.13$ & $20.18$ & $31.33$ & $45.13$ & $19.95$ & $30.33$ & $42.98$ & $56.88$ & $29.77$ & $41.44$ & $54.64$ & $67.71$ \\
MTriplet & 32 & $12.92$ & $20.71$ & $31.75$ & $45.02$ & $19.80$ & $30.02$ & $41.74$ & $55.50$ & $29.73$ & $41.24$ & $54.57$ & $67.96$ \\

\bottomrule
\end{tabular}
\end{scriptsize}

\end{table*}
\begin{table*}[h!]
\caption{Results on Stanford Cars} \label{cars-results}
\vskip 0.15in
\centering
\begin{scriptsize}
\begin{tabular}{c c | c c c c | c c c c | c c c c}
\toprule
{\sc method} & {\sc \# charts} & \multicolumn{4}{c|}{\sc encoding dim 2} & \multicolumn{4}{c|}{\sc encoding dim 4} & \multicolumn{4}{c}{\sc encoding dim 8} \\
\cmidrule(l){3-14}
 &  & R@1 & R@2 & R@4 & R@8 & R@1 & R@2 & R@4 & R@8 & R@1 & R@2 & R@4 & R@8 \\
\midrule

Triplet & & $10.92$ & $18.73$ & $30.85$ & $46.81$ & $18.96$ & $29.91$ & $43.81$ & $59.55$ & $33.53$ & $47.20$ & $61.10$ & $74.14$ \\
MTriplet & 1 & $9.62$ & $17.02$ & $28.26$ & $44.29$ & $16.51$ & $27.15$ & $41.20$ & $56.76$ & $32.78$ & $46.24$ & $60.73$ & $74.11$ \\
MTriplet & 4 & $13.14$ & $22.00$ & $35.52$ & $52.34$ & $22.51$ & $34.56$ & $49.49$ & $64.75$ & $38.77$ & $52.54$ & $65.92$ & $77.94$ \\
MTriplet & 16 & $14.07$ & $24.12$ & $38.27$ & $54.61$ & $25.35$ & $37.89$ & $52.46$ & $67.04$ & $38.77$ & $52.43$ & $65.93$ & $77.74$ \\
MTriplet & 32 & $15.87$ & $25.81$ & $39.51$ & $55.50$ & $25.68$ & $38.52$ & $52.80$ & $67.38$ & $38.78$ & $53.05$ & $66.61$ & $78.13$ \\

\bottomrule
\end{tabular}
\end{scriptsize}

\end{table*}
\endgroup


\subsubsection{MMD loss alone helps in high-dimensions} \label{triplet-reg}
As we moved to dimensions higher than eight, we noticed that increasing the number of charts caused diminishing performance but the case of a single chart gave significant improvements over no regularization. In other words, doing standard triplet training for an embedding network $f: \X \to \R^d$ with sigmoid activation, metric defined as in the previous section, and adding the regularization loss term (\ref{ellZ-def}), which in this case simplifies to

\begin{align*}
\ell_\Z(x_1, \ldots, x_n) = \frac{1}{N(N-1)} &\sum_{\substack{j, k = 1 \\ j \ne k}}^N k_0(f(x_j), f(x_k)) - \frac{2}{N^2} \sum_{j, k = 1}^N k_0(f(x_j), w_k) \\
&+  \frac{1}{N(N-1)} \sum_{\substack{j, k = 1 \\ j \ne k}}^N k_0(w_j, w_k),
\end{align*}
where $\{x_1, \ldots, x_n\}$ is a minibatch and $\{w_1, \ldots, w_N\}$ is a random sample drawn uniformly from $[0,1]^d$.

For this experiment we trained 64-dimensional encodings, again using an ImageNet pretrained GoogLeNet backbone. As in the previous section, we weigh the regularization loss by $\lambda_1 = 1$. For baselines we take the standard, normalized triplet training (denoted ``baseline" in the plots) as well as the case of MTriplet with $n=1, \lambda_1 = 0$ (denoted ``unnormalized" in the plots) to control that the performance improvement is due to the regularization and not that we use unnormalized embedding vectors. We train for 1,000 epochs three times in each case and report in figures \ref{cub-reg} and \ref{cars-reg} the evaluation metrics on the test set as a function of epochs trained. For CUB-200, our results are slightly under the state-of-the art results (using 64 dimensional embeddings with a GoogLeNet backbone) reported in \cite{xuan2020improved}, while for Stanford Cars we outperform the other algorithms at R@4 and R@8.

\begingroup
\setlength{\tabcolsep}{-4pt}
\def\arraystretch{0}
\begin{figure}[h!] 
\begin{tabular}{cccc}
\includegraphics[width=.5\textwidth]{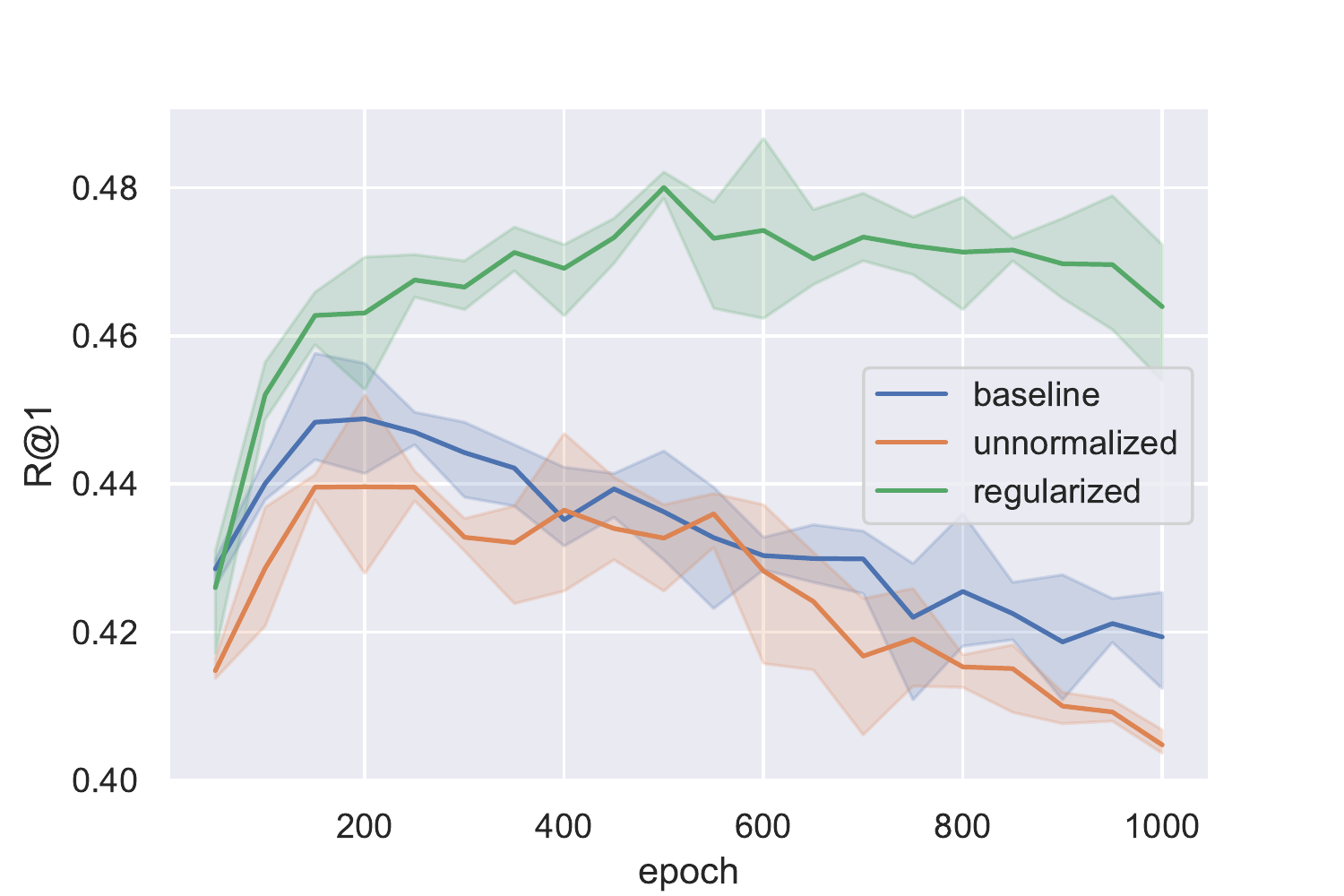} & \includegraphics[width=.5\textwidth]{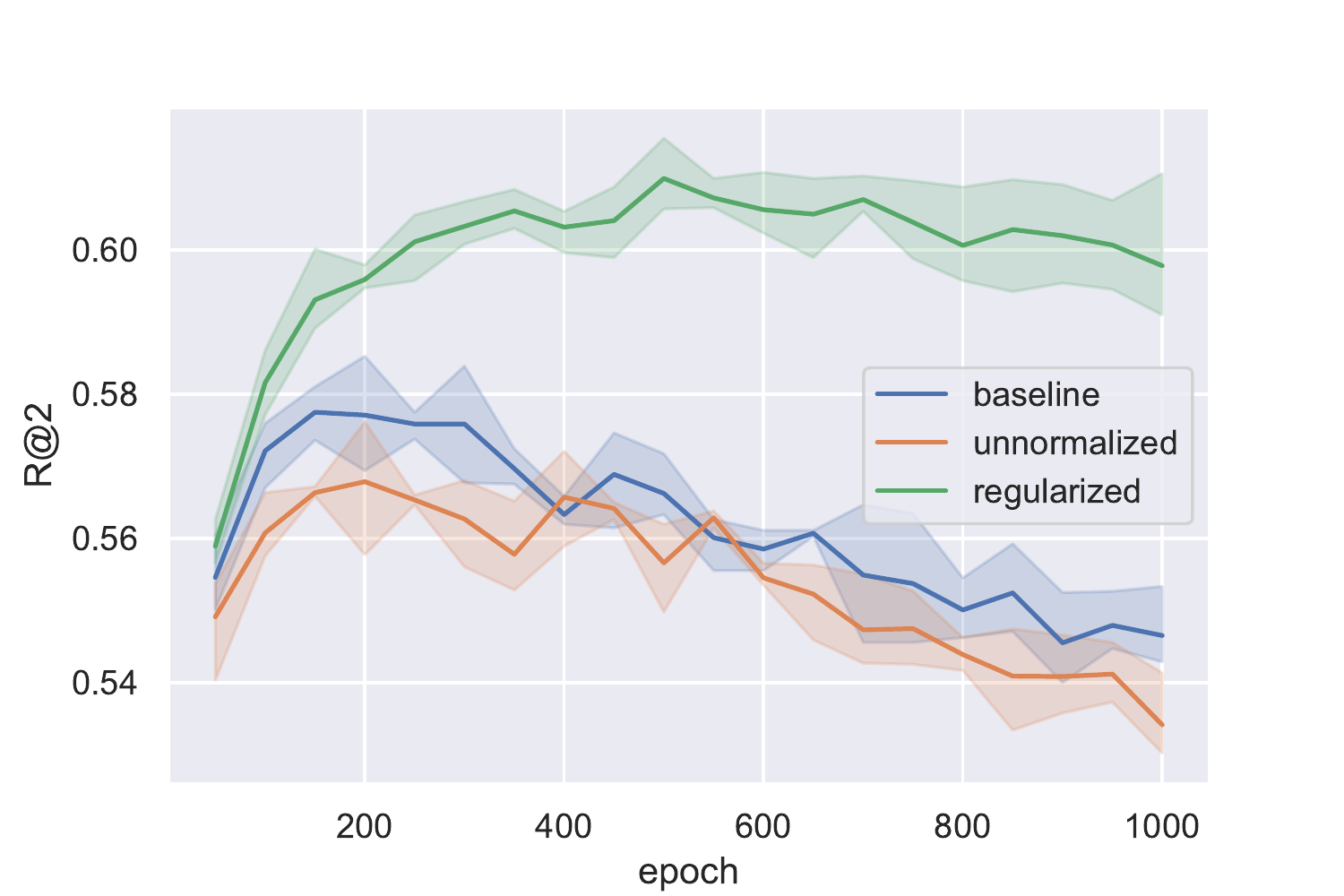} \\ \includegraphics[width=.5\textwidth]{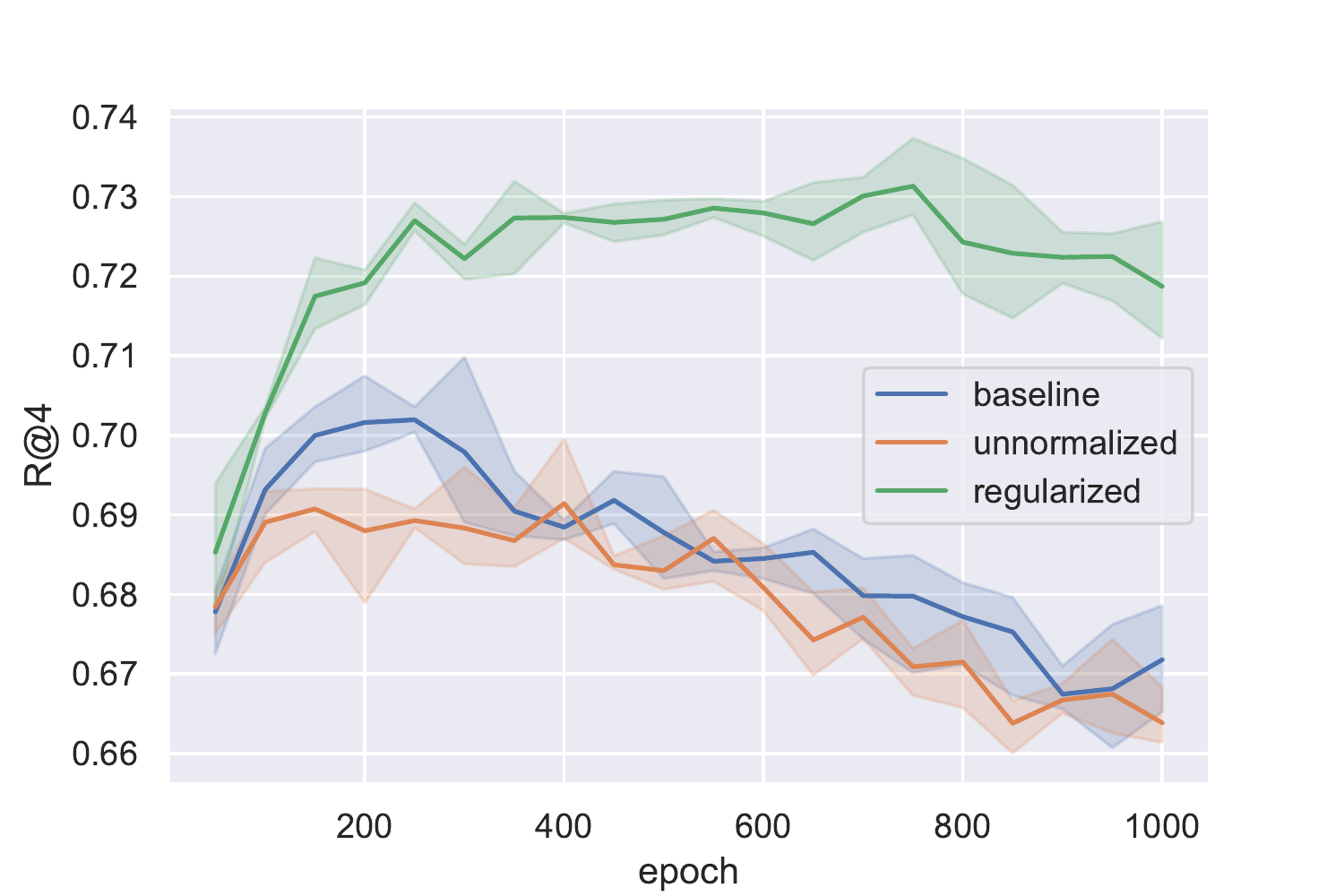} & \includegraphics[width=.5\textwidth]{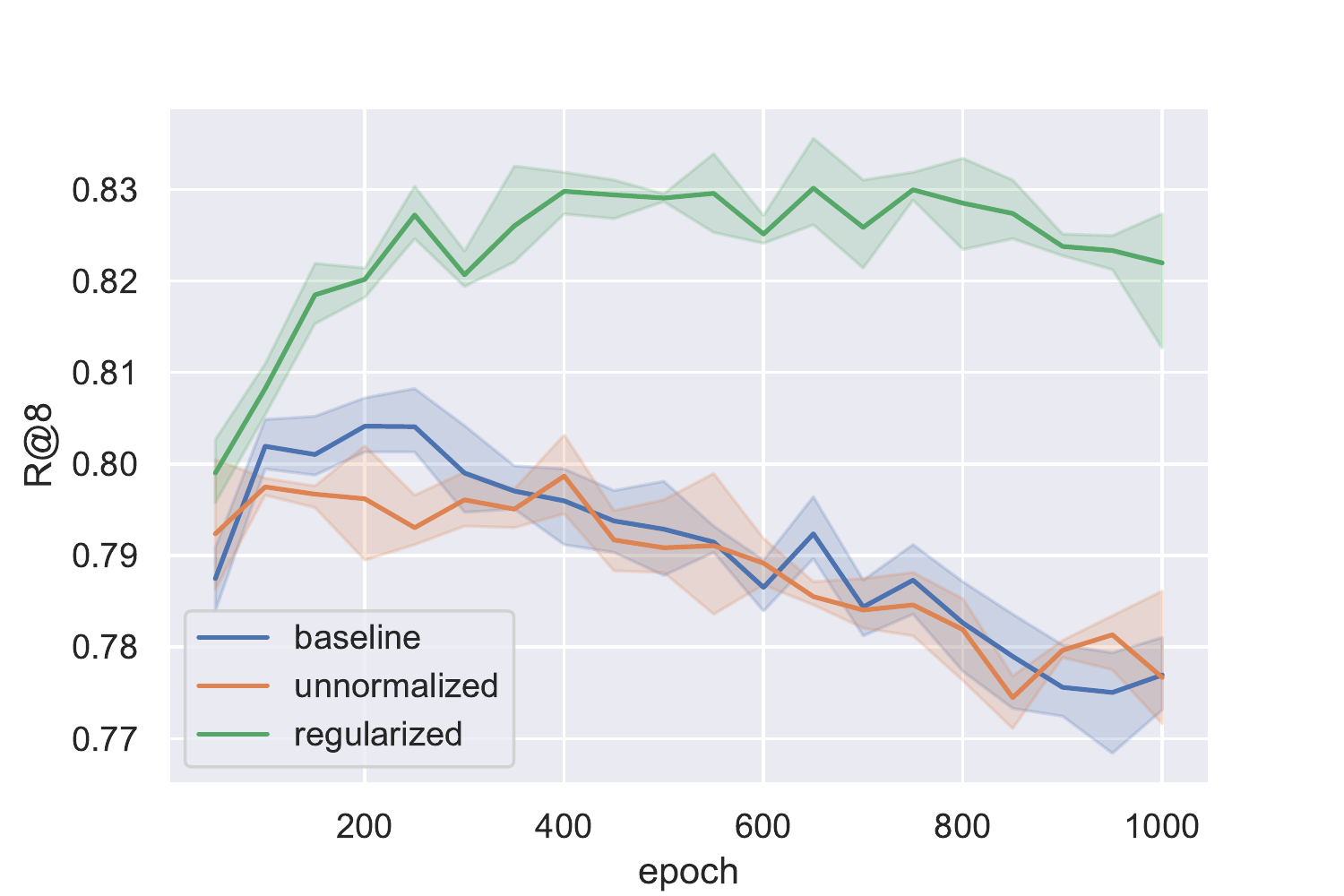}
\end{tabular}
\caption{Test set performance vs. number of epochs trained, on the CUB-200 dataset.} \label{cub-reg}
\end{figure}
\begin{figure}[h!]
\begin{tabular}{cc}
\includegraphics[width=.5\textwidth]{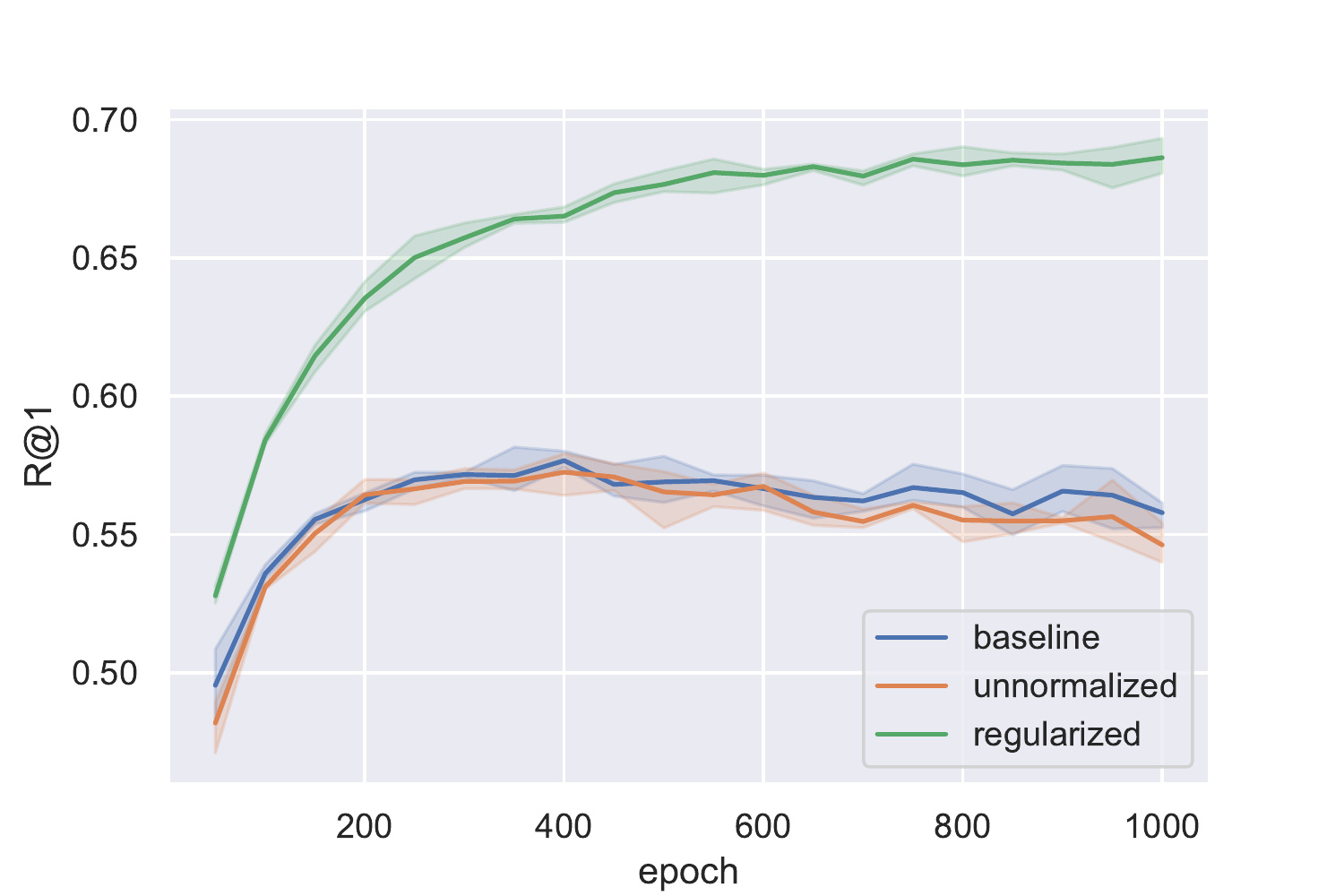} & \includegraphics[width=.5\textwidth]{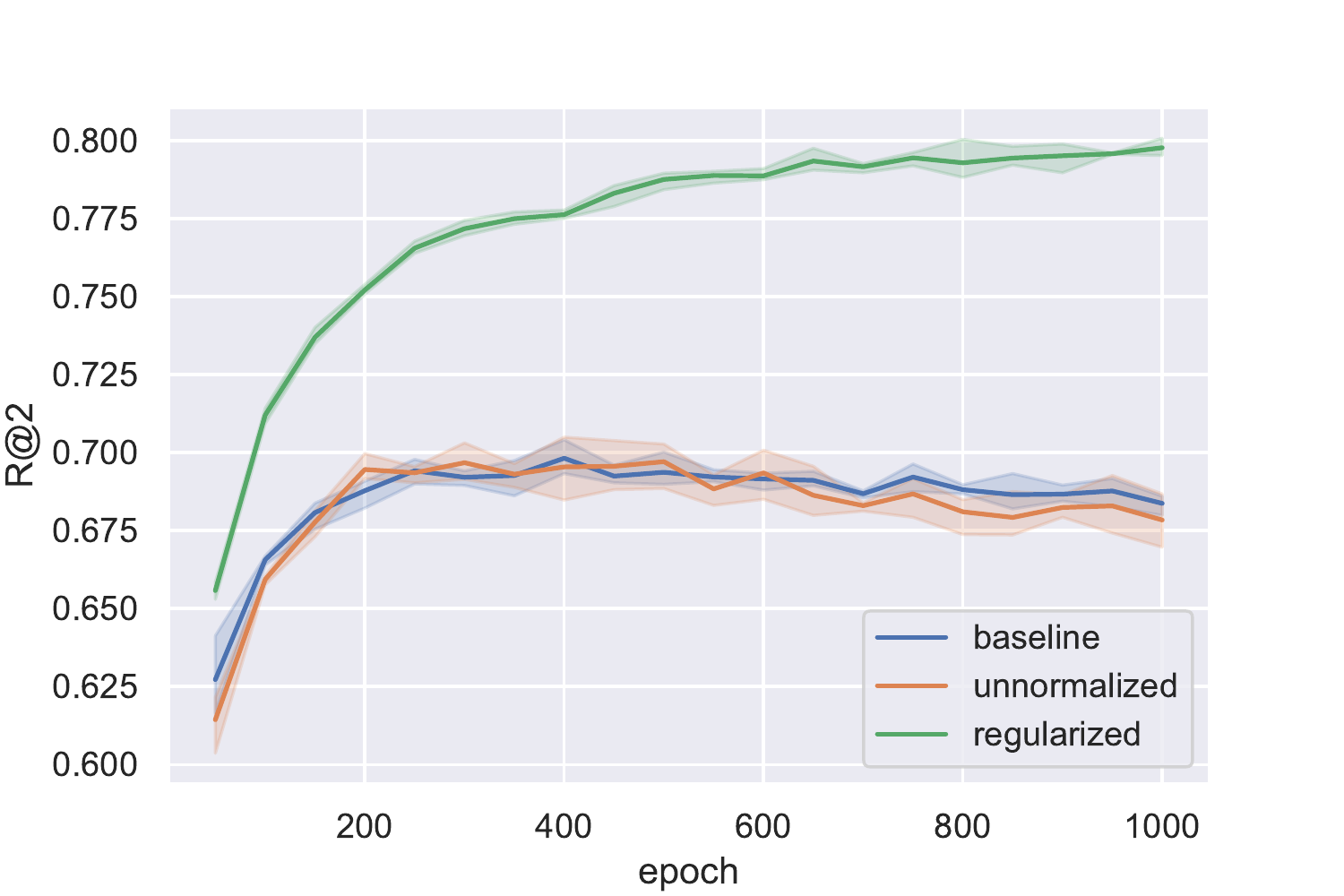} \\\includegraphics[width=.5\textwidth]{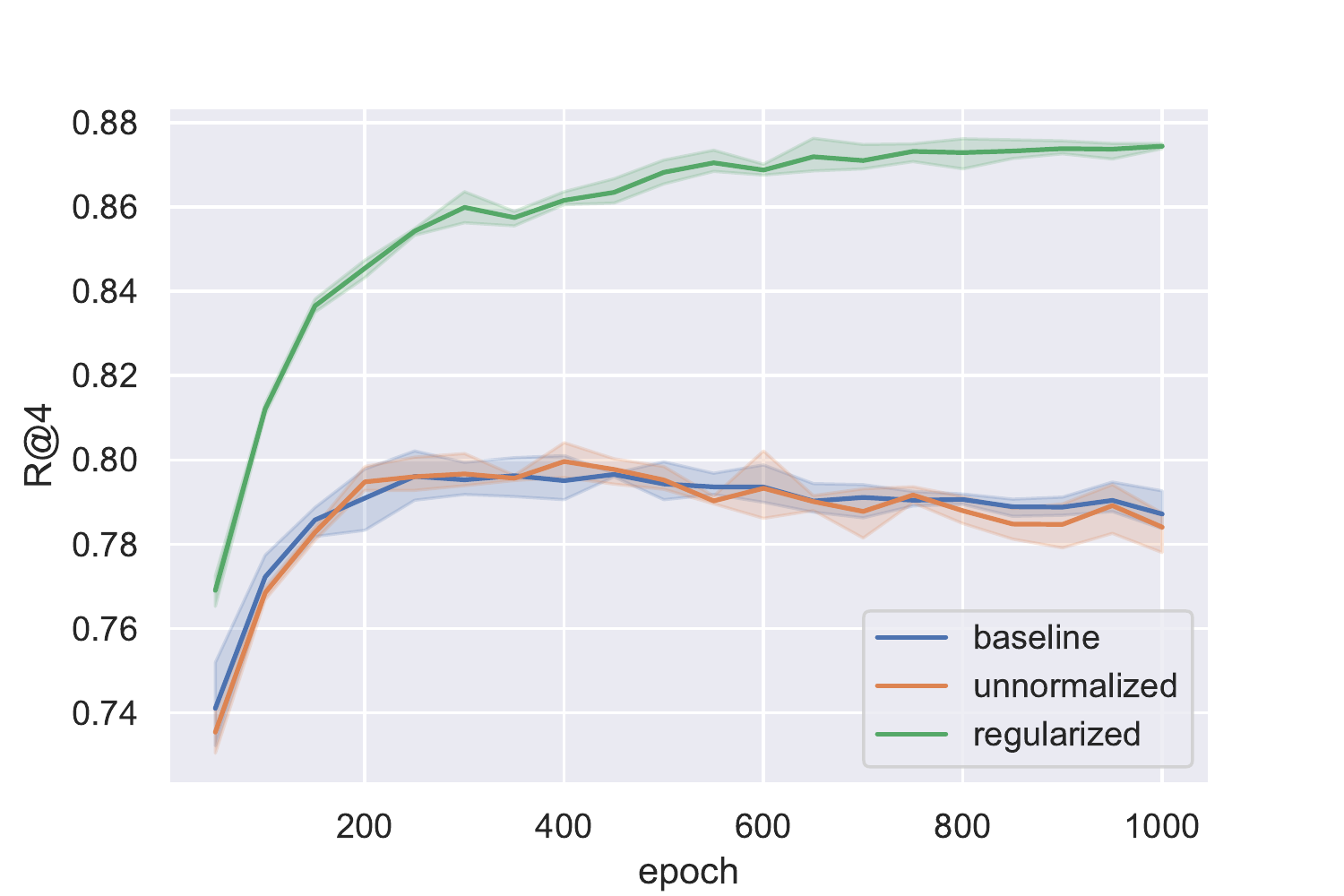} & \includegraphics[width=.5\textwidth]{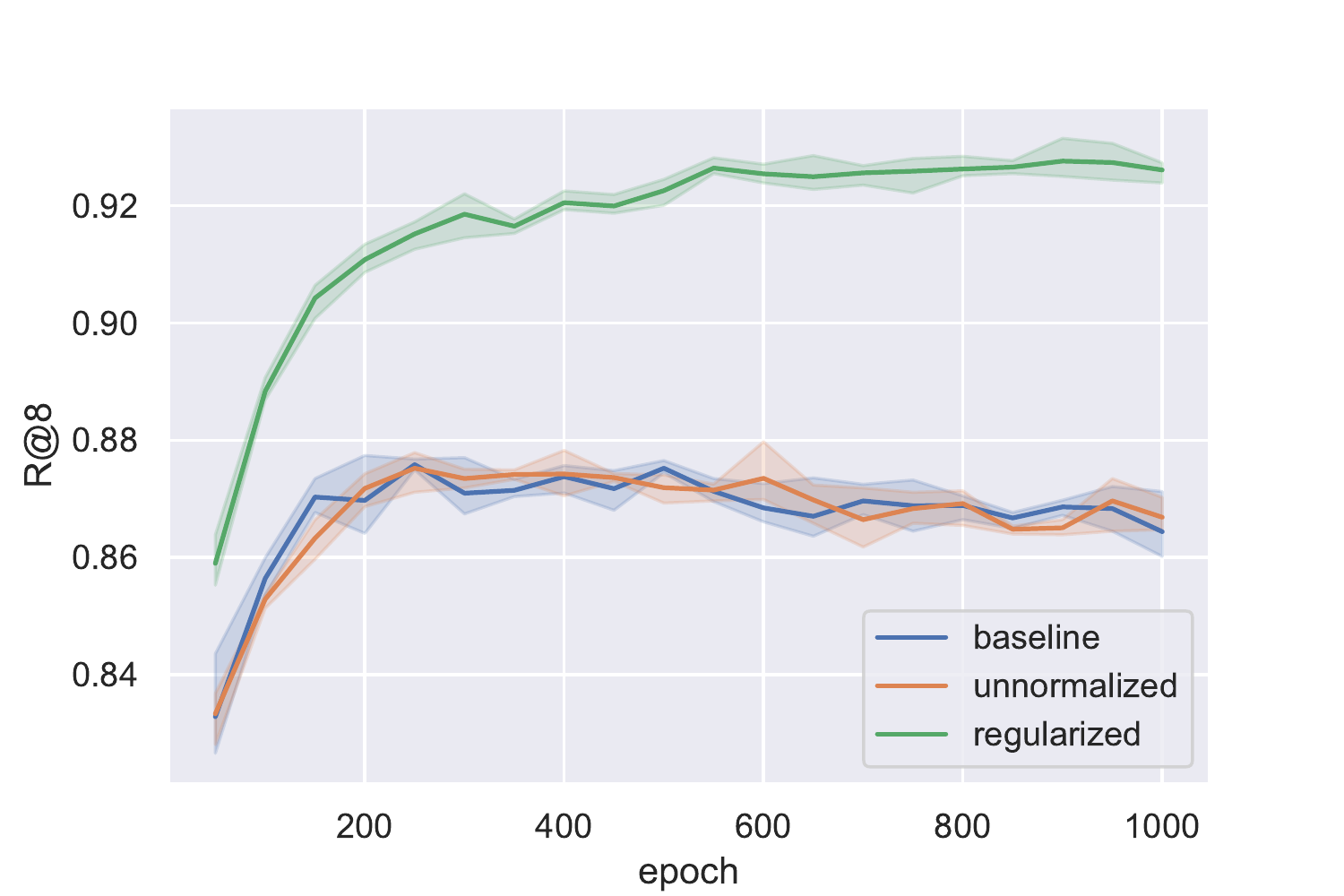}
\end{tabular}
\caption{Test set performance vs. number of epochs trained, on the Stanford Cars dataset.} \label{cars-reg}
\end{figure}

\endgroup

\pagebreak

\section{Conclusion}
We presented a method for adjusting representation learning algorithms to learn an atlas of a manifold instead of mapping into a Euclidean space. This allows, for a given encoding dimension, more interesting geometries of the embedding space. Our experiments with SimCLR and triplet-loss based metric learning training show that for low encoding dimensions our approach gives much more powerful representations over the baseline. Independent of manifold learning, we also showed that triplet loss training is significantly enhanced by adding a regularizing MMD loss function encouraging the embedding space to be uniform. We hope that this work leads to further research into atlas based manifold learning techniques.


\bibliography{bibliography}
\bibliographystyle{iclr2021_workshop}

\appendix
\section{Appendix}
\subsection{Experiments setup} \label{sec-setup}
We ran our experiments using PyTorch \cite{paszke2019pytorch}. In all cases we use a batch size of 128 and the Adam optimizer with learning rate $10^{-4}$ and $\beta_1 = 0.9, \beta_2 = 0.999$ (the PyTorch defaults). For MNIST and FashionMNIST we use a ResNet18 \cite{he2016deep} backbone and train for 100 epochs while for CIFAR10 we use a ResNet50 backbone and train for 1,000 epochs.  We use the same data augmentation used in the CIFAR10 experiments in the original work \cite{chen2020simple}: color jitter with strength 0.5 (and a probability of 0.8 of applying), random grayscale with probability 0.2, and a random resized crop.

Each of the datasets comes with a standard train/test split. For hyperparameter selection of  $\lambda_1, \lambda_2$, and $\tau$ (the temperature used in the contrastive loss function) we do a grid search (with $d = 2, n=16$) over $\lambda_1 \in \{0.1, 1, 5, 10, 20\}, \lambda_2 \in \{0.05, 0.1, 0.2\}$, and $\tau \in \{0.1, 0.5, 1\}$ by training on a random selection of 80\% of the training data and then computing the piece-wise linear evaluation on the remaining 20\%. For MNIST and FashionMNIST we chose the hyperparameters that give the best average rank across the two datasets. This yielded $\lambda_1 = 20, \lambda_2 = 0.1$, and $\tau = 1$ for MNIST and FashionMNIST and $\lambda_1 = 20, \lambda_2 = 0.1$, and $\tau = 0.5$ for CIFAR10. Using these parameters we train on the entire train dataset across $d \in \{2, 4, 8\}$ (we noticed for higher $d$ there is not much improvement over the baseline) and $n \in \{1, 4, 16, 32\}$. We also train baseline SimCLR for comparison (but were unable to get convergence in dimension two for CIFAR10). We repeat each training configuration three times and report the mean $\pm$ std of accuracy on the holdout test set in Table \ref{results-table}.

\pagebreak

\subsection{Visualizations} \label{sec-visualizations}
In this section we display visualizations for each of the datasets used in our MSimCLR experiments, using a model with $d=2, n=16$. Each image corresponds to a chart and every image in the test set gets plotted at its $(x, y)$ coordinate in the chart with highest probability. The images are best viewed by zooming in.
\vspace{-10pt}
\subsubsection{CIFAR10}
\begin{table}[h]
\begin{tabular}{cccc}
\includegraphics[width=.25\textwidth]{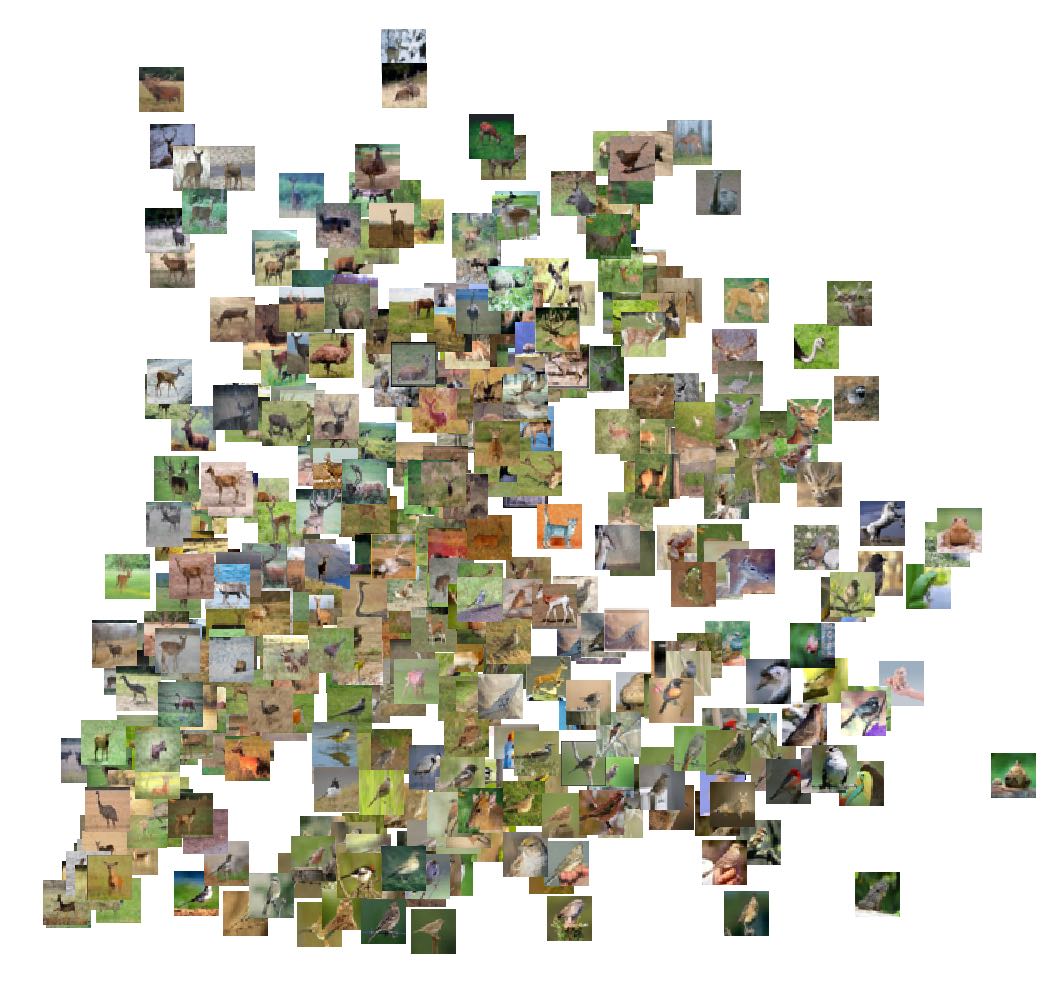} & \includegraphics[width=.25\textwidth]{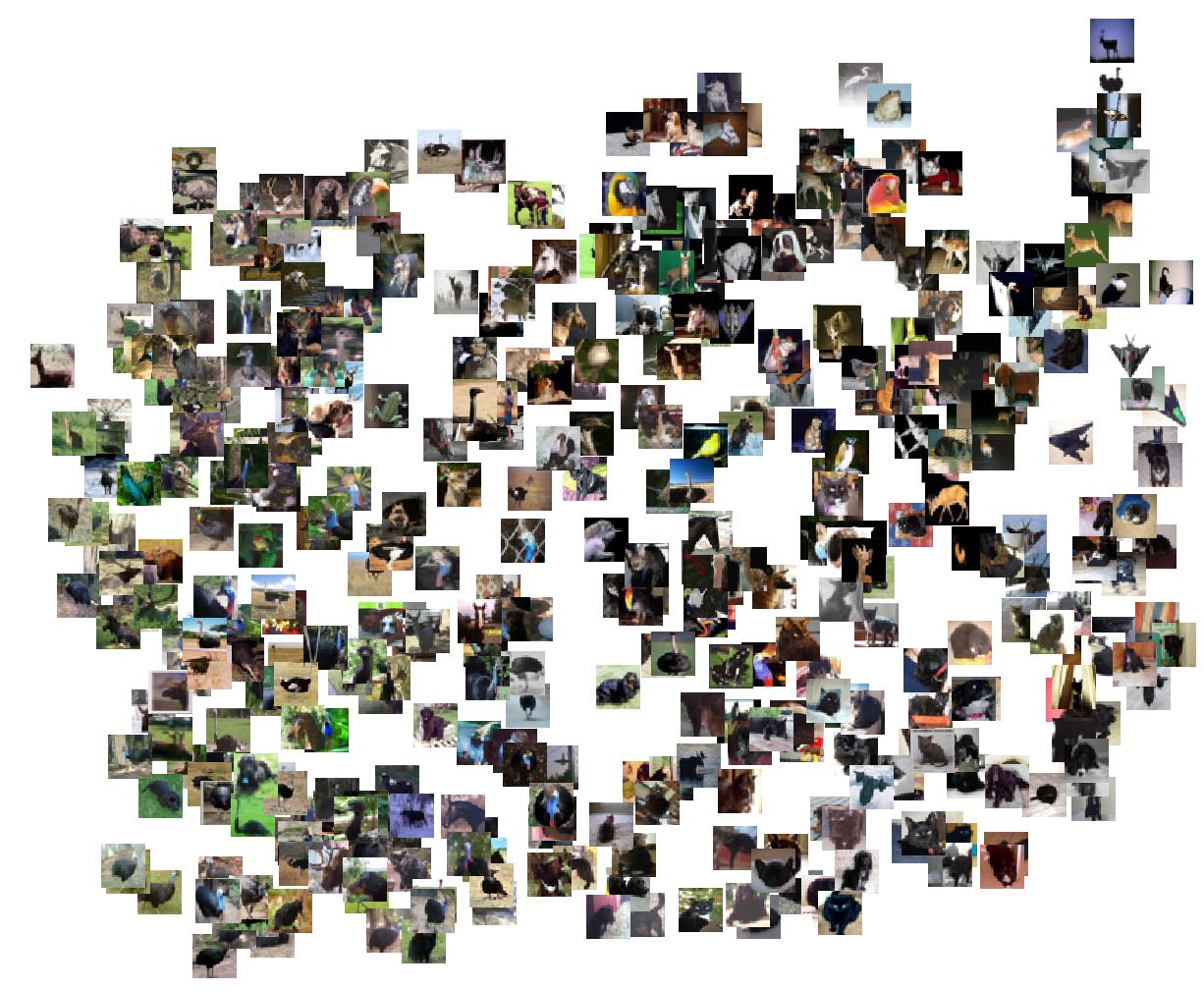} & \includegraphics[width=.25\textwidth]{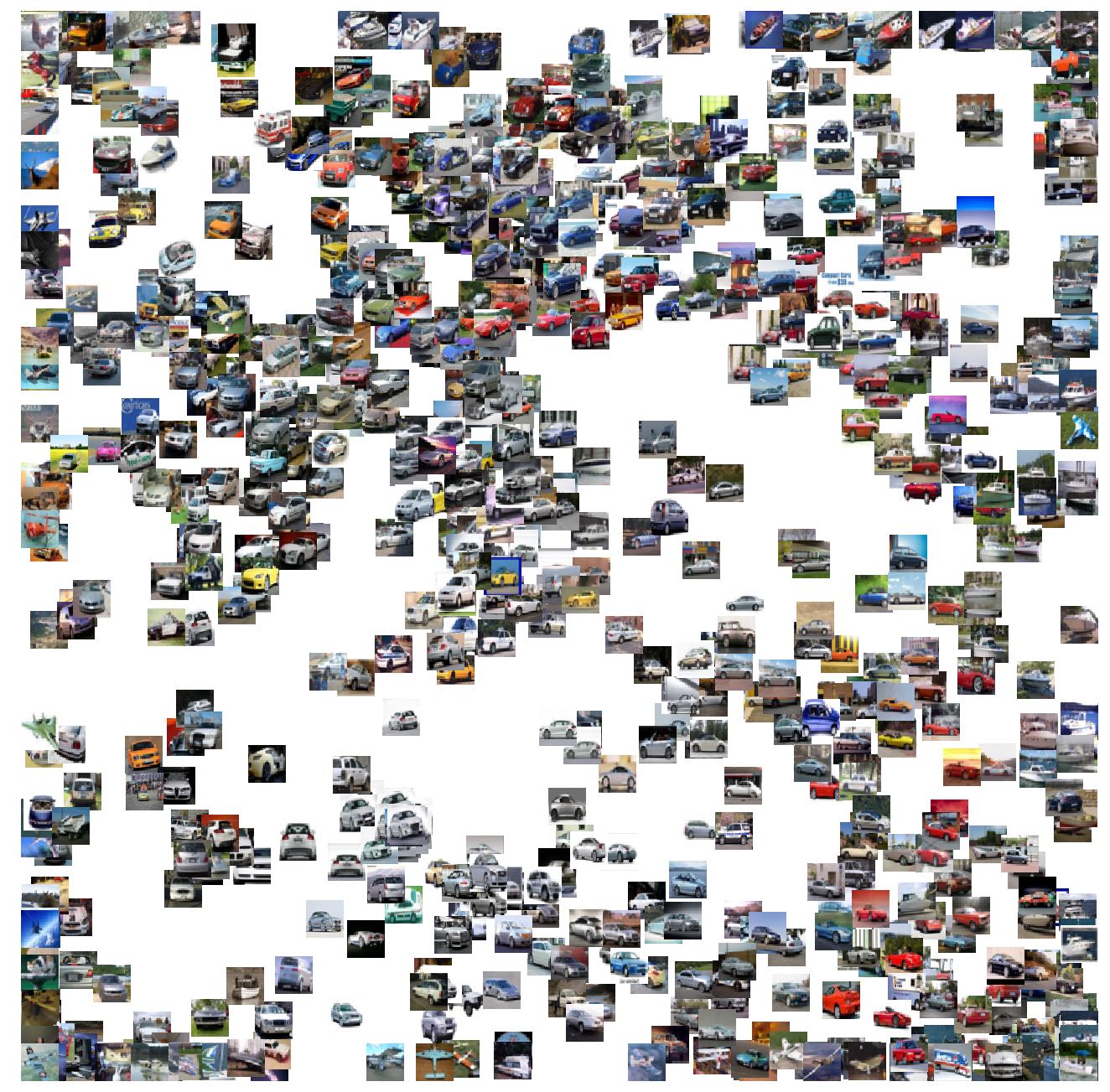} & \includegraphics[width=.25\textwidth]{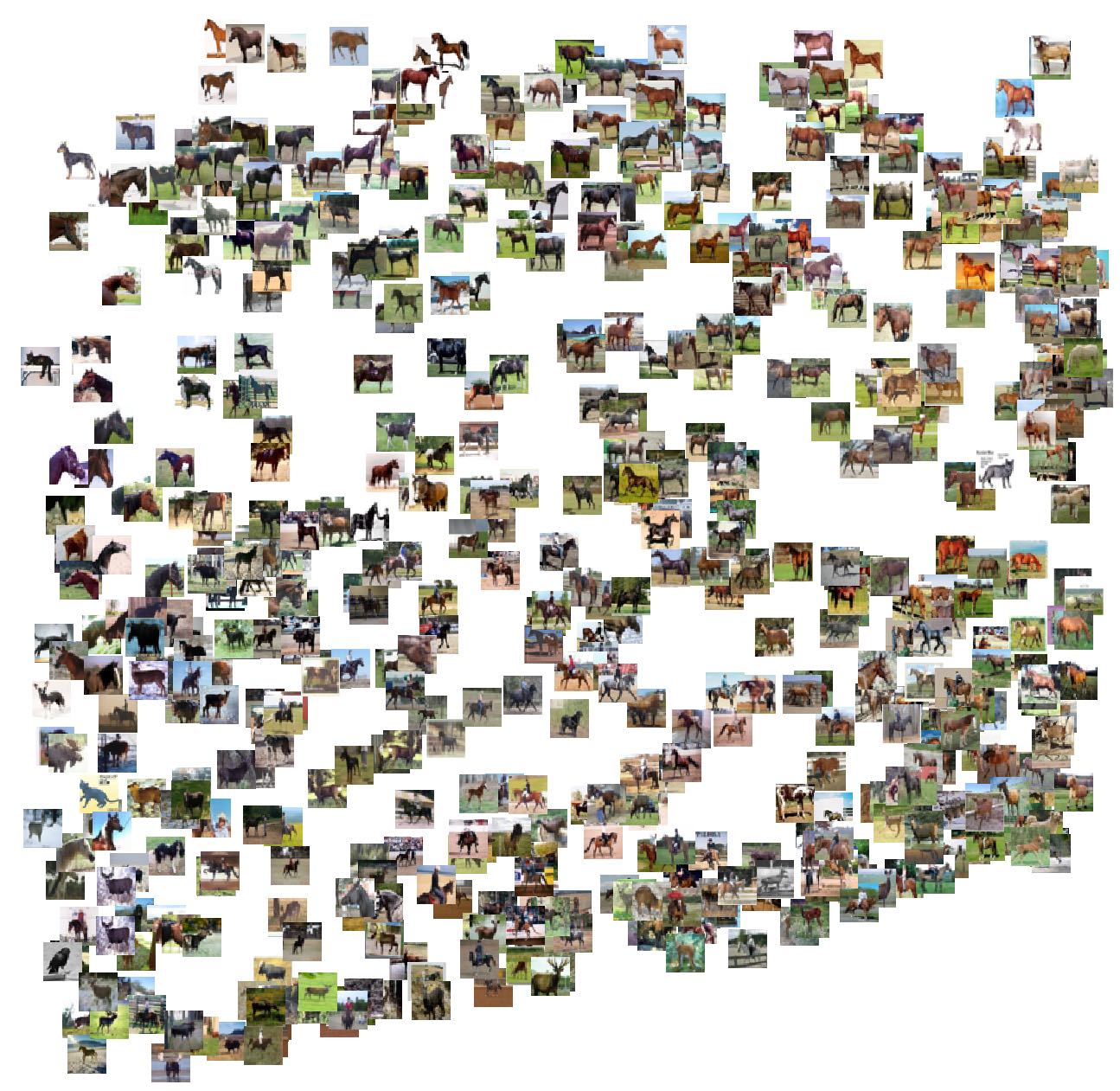}\\
\includegraphics[width=.25\textwidth]{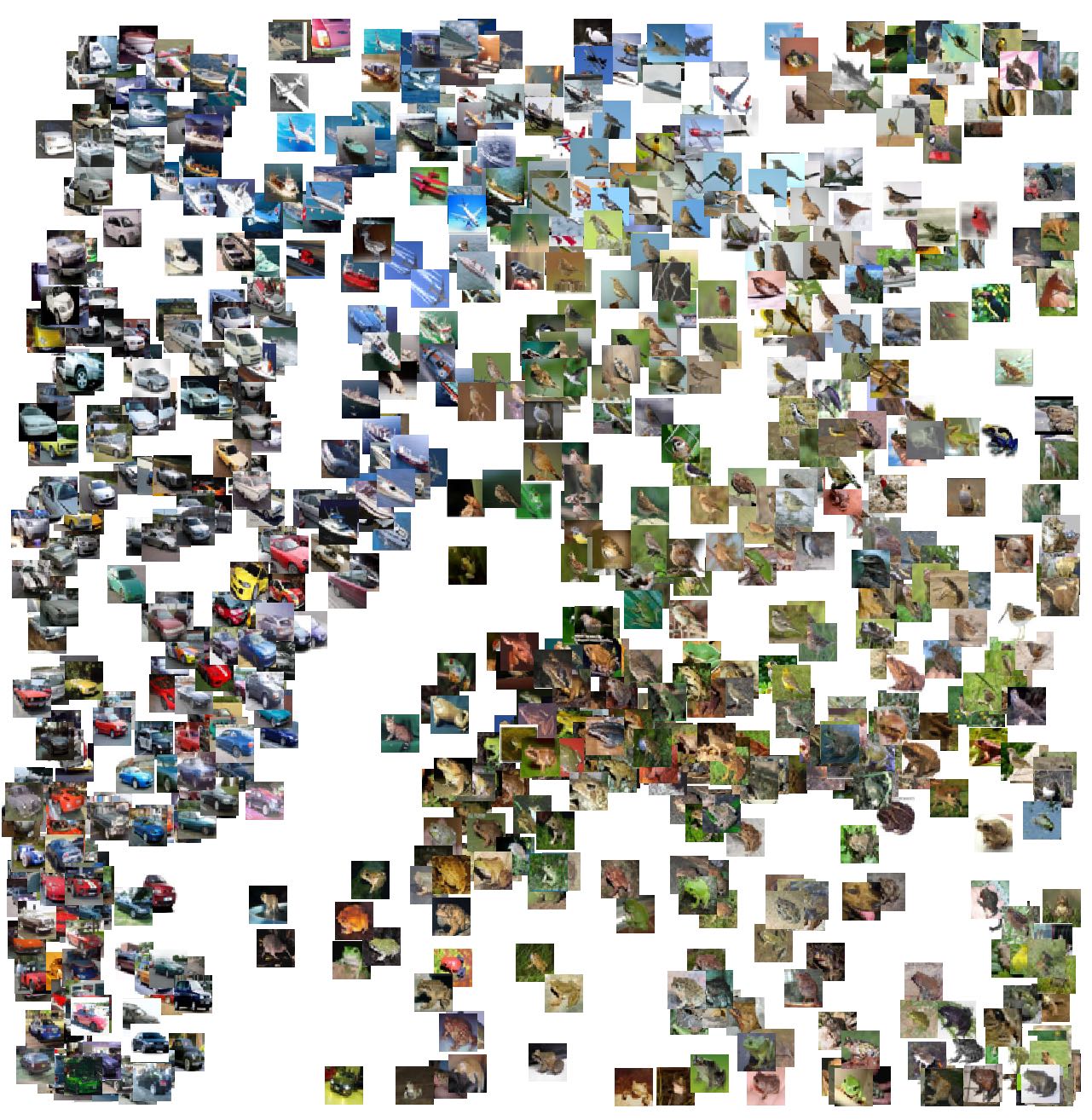} & \includegraphics[width=.25\textwidth]{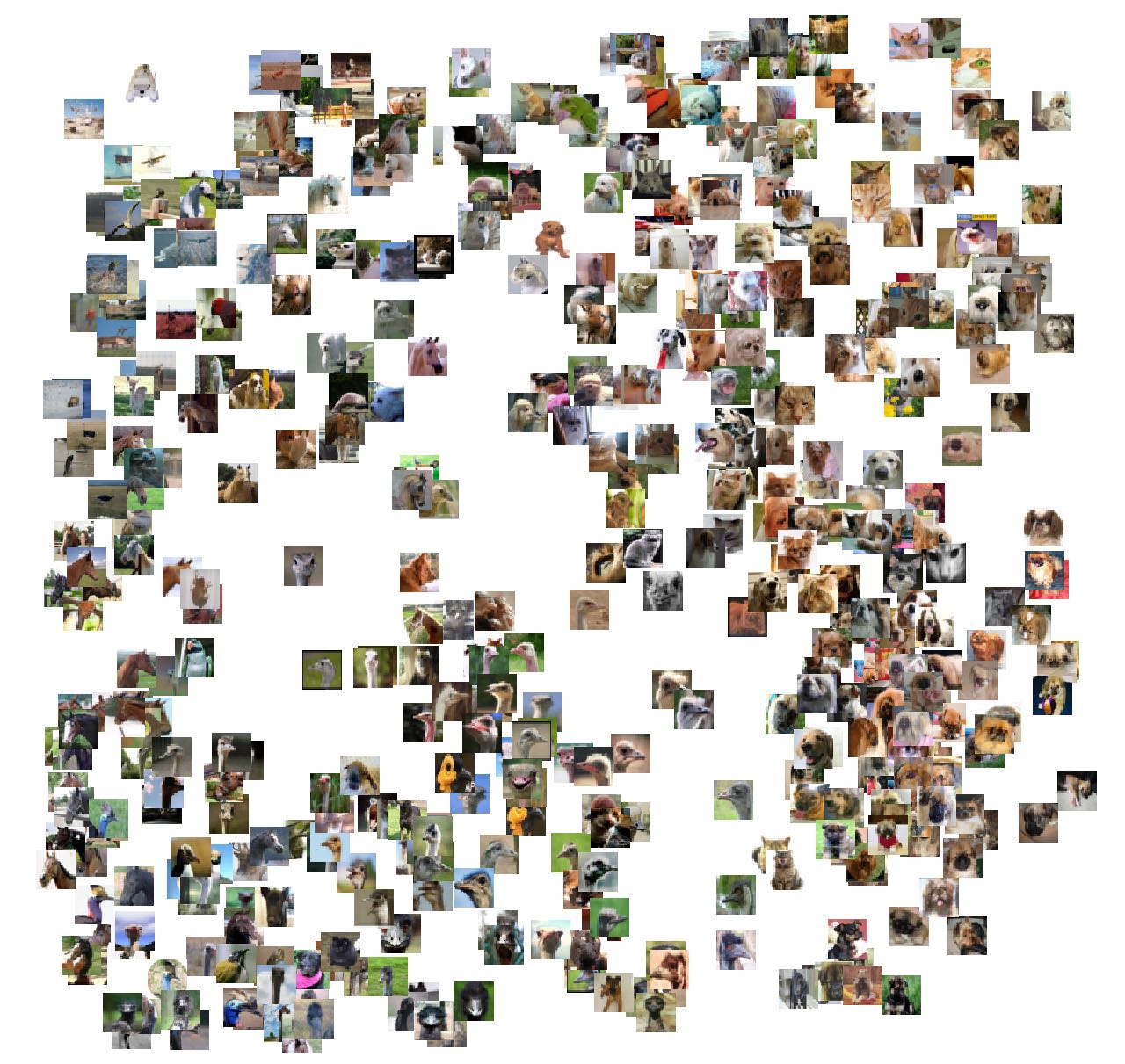} & \includegraphics[width=.25\textwidth]{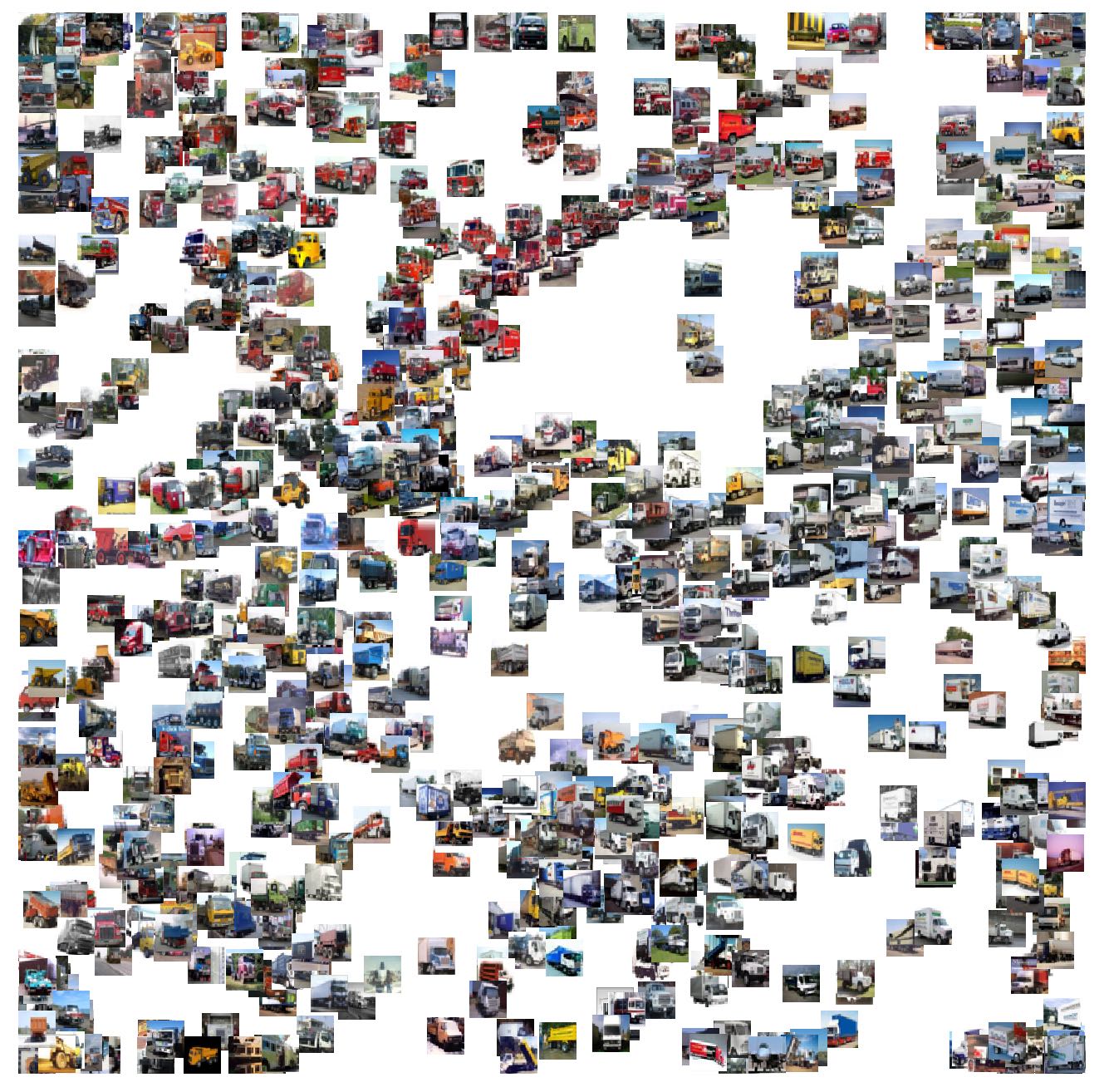} & \includegraphics[width=.25\textwidth]{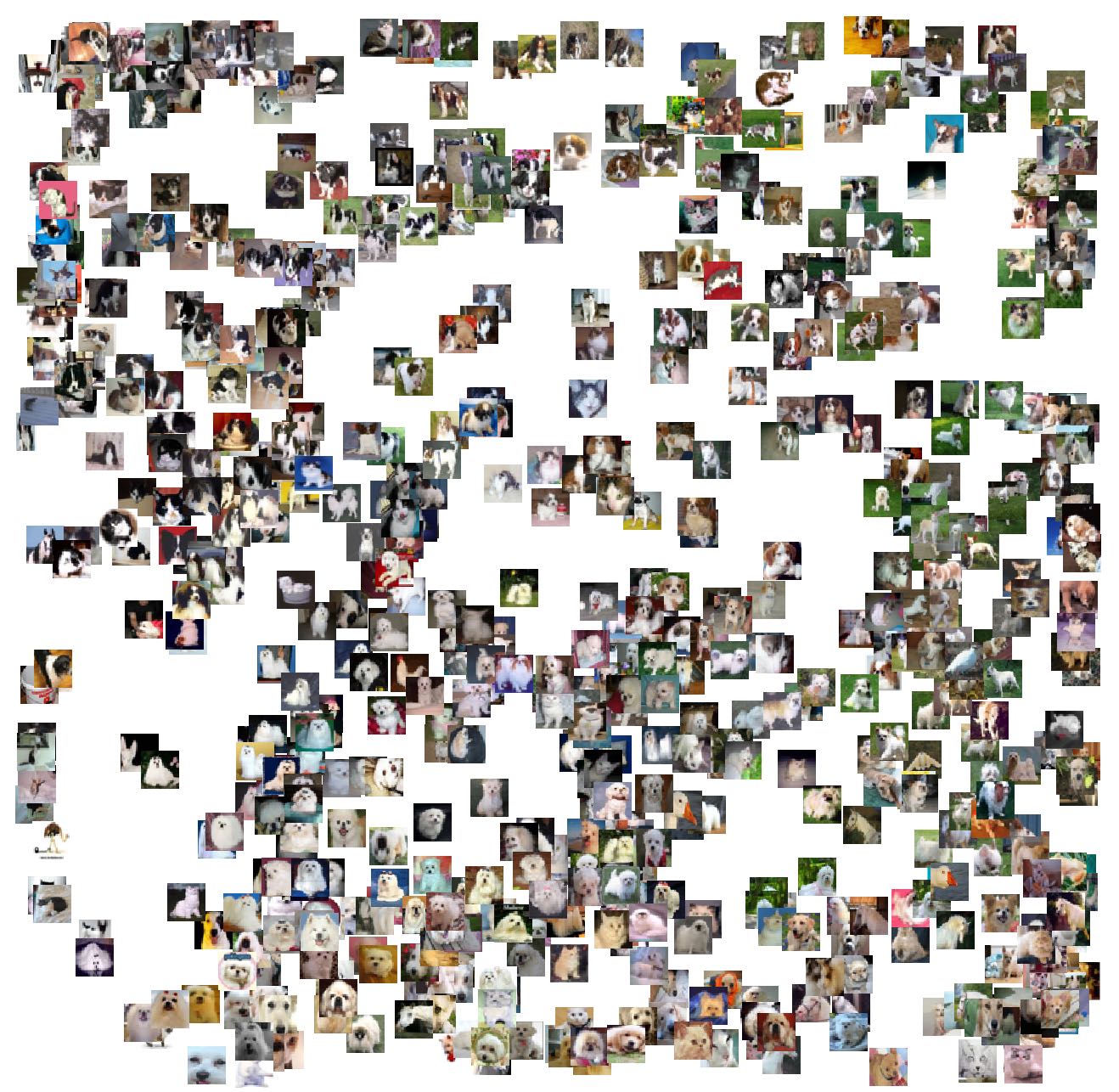}\\
\includegraphics[width=.25\textwidth]{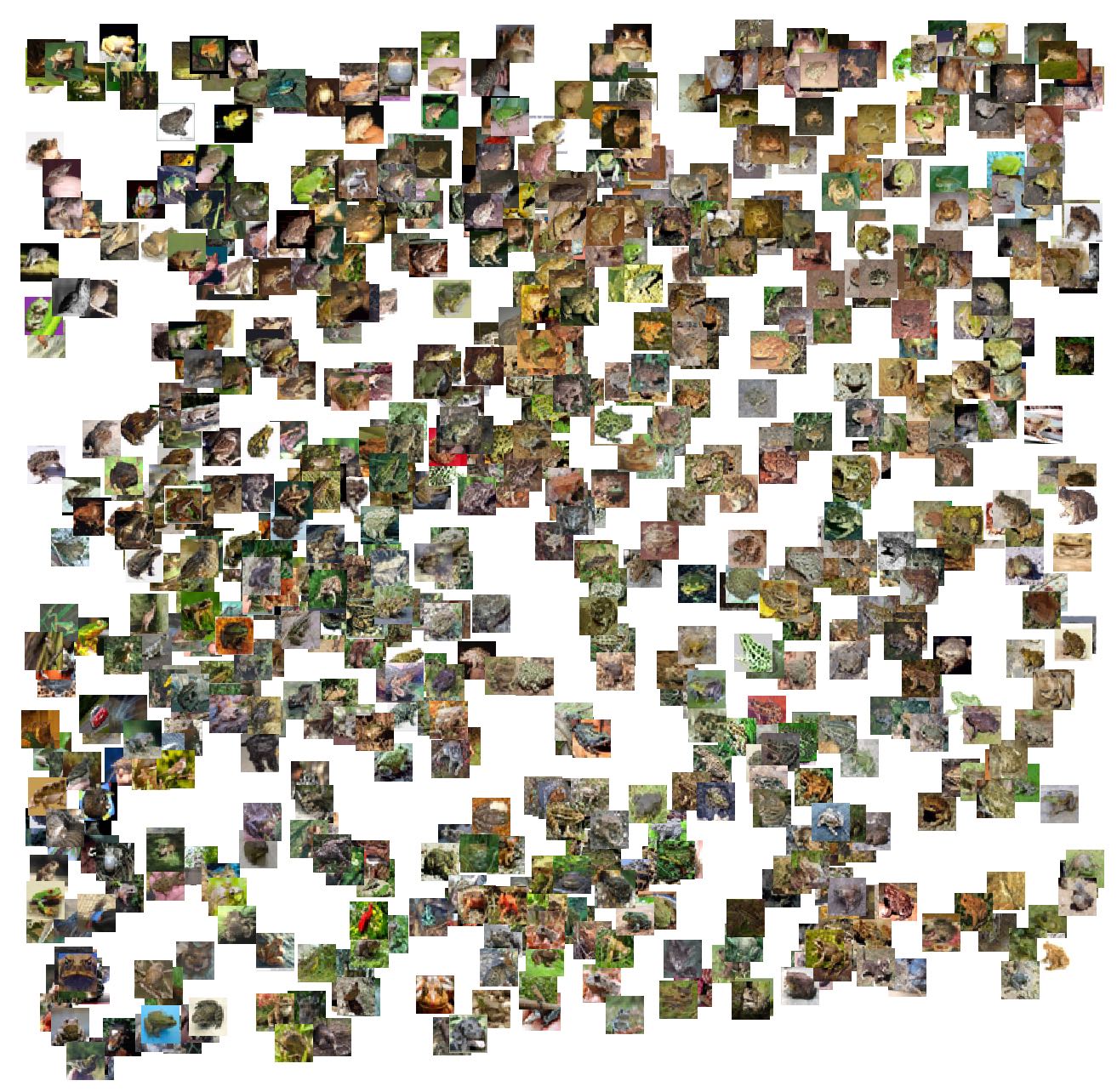} & \includegraphics[width=.25\textwidth]{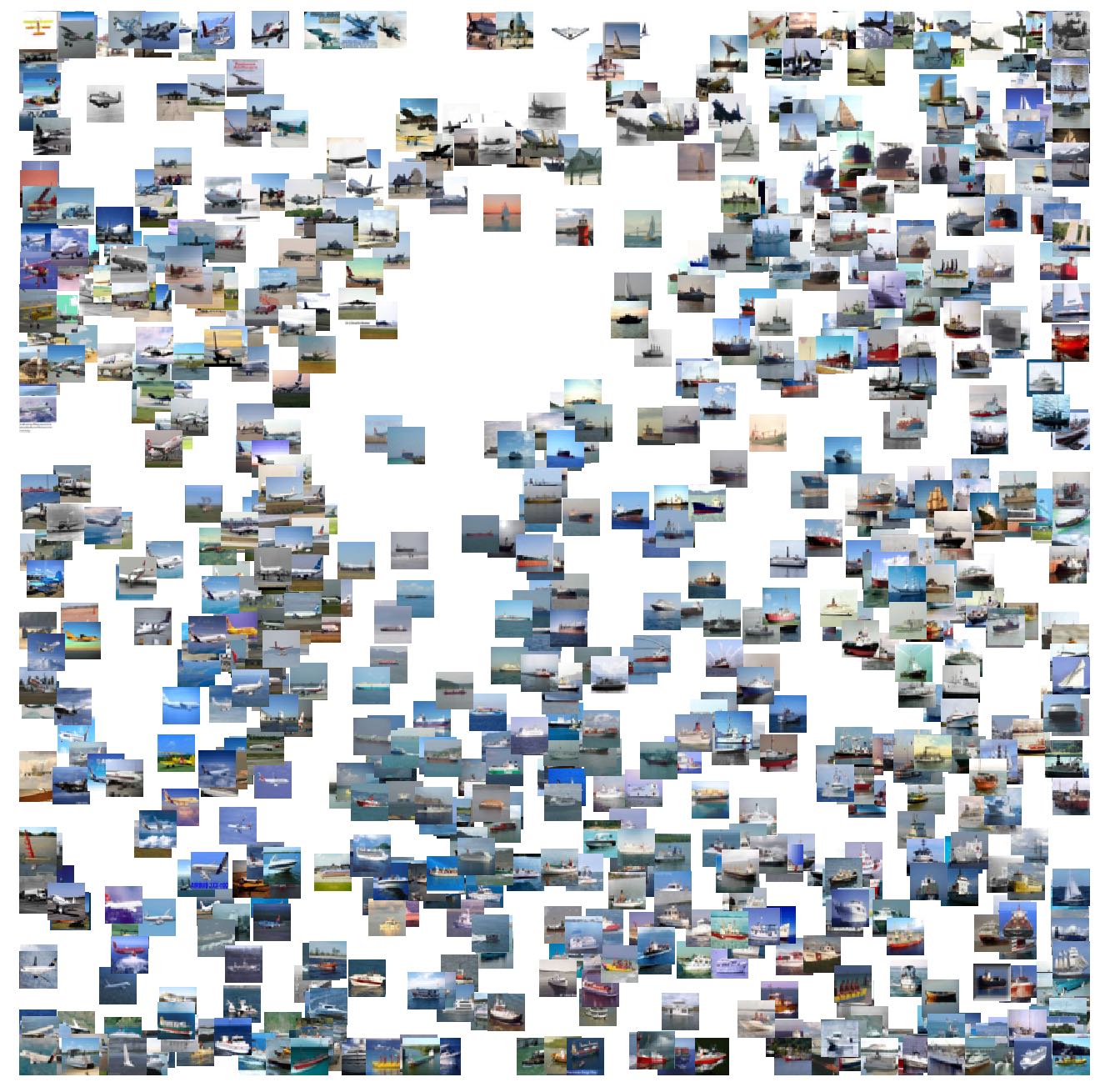} & \includegraphics[width=.25\textwidth]{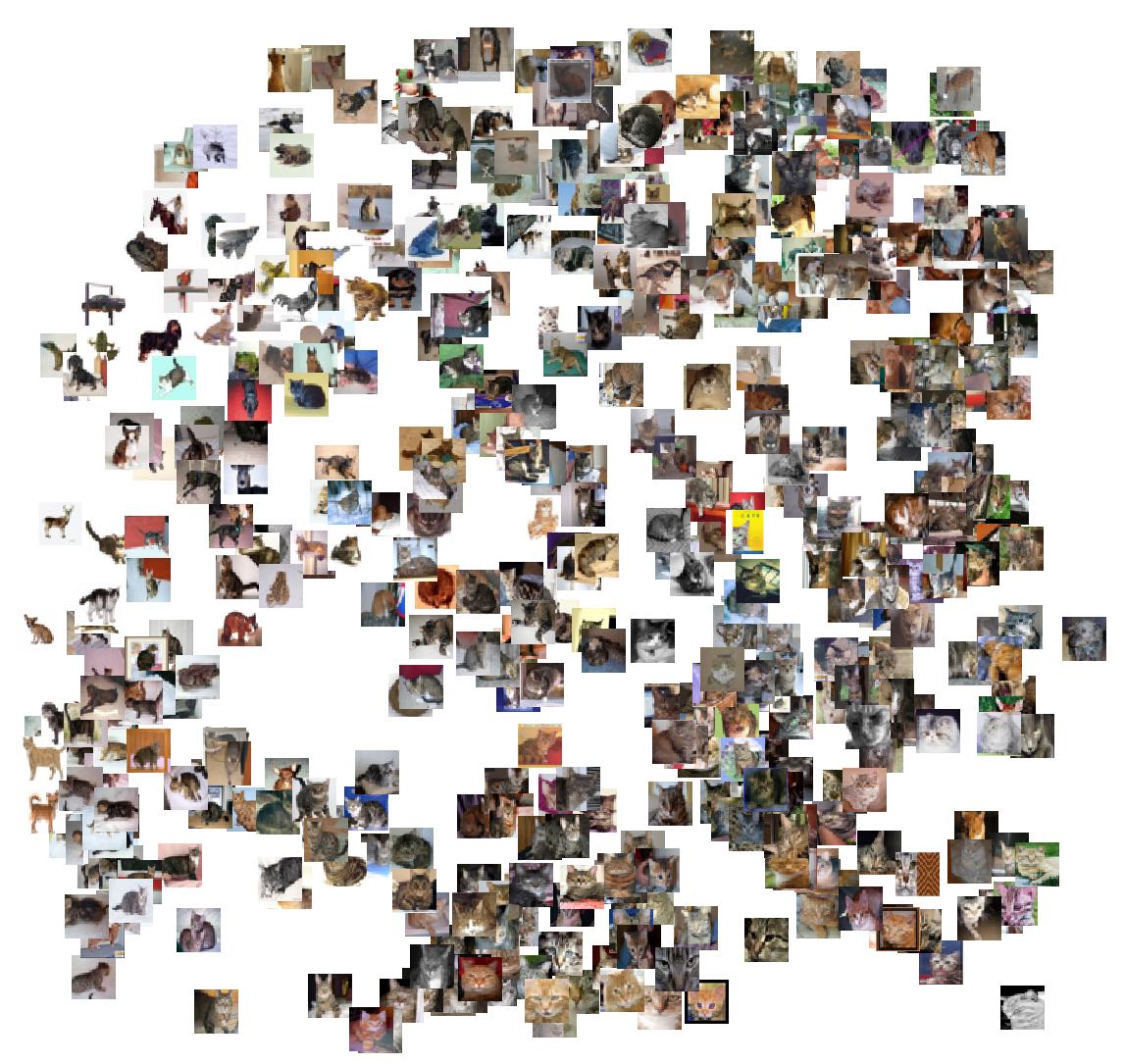} & \includegraphics[width=.25\textwidth]{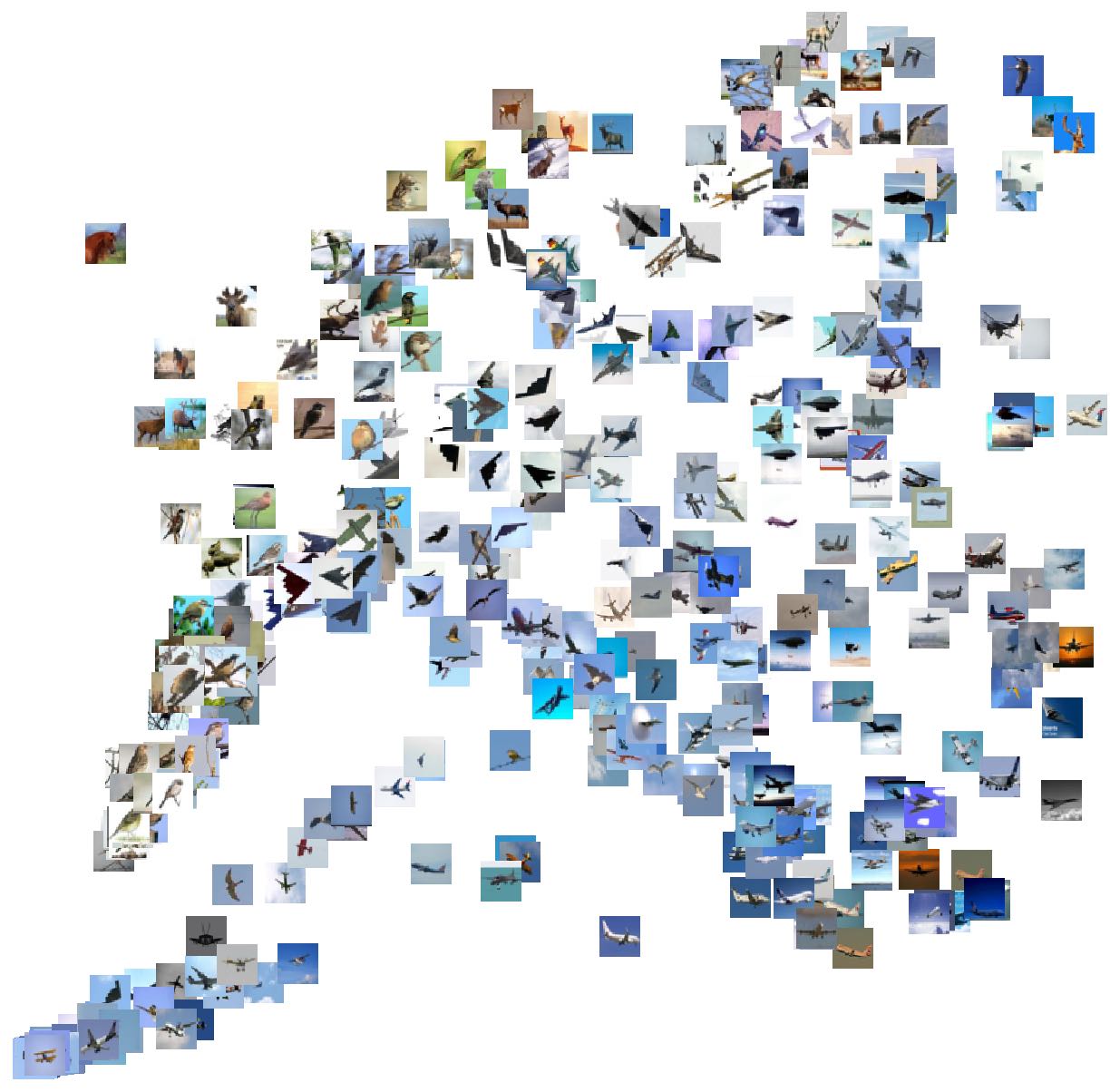}\\
\includegraphics[width=.25\textwidth]{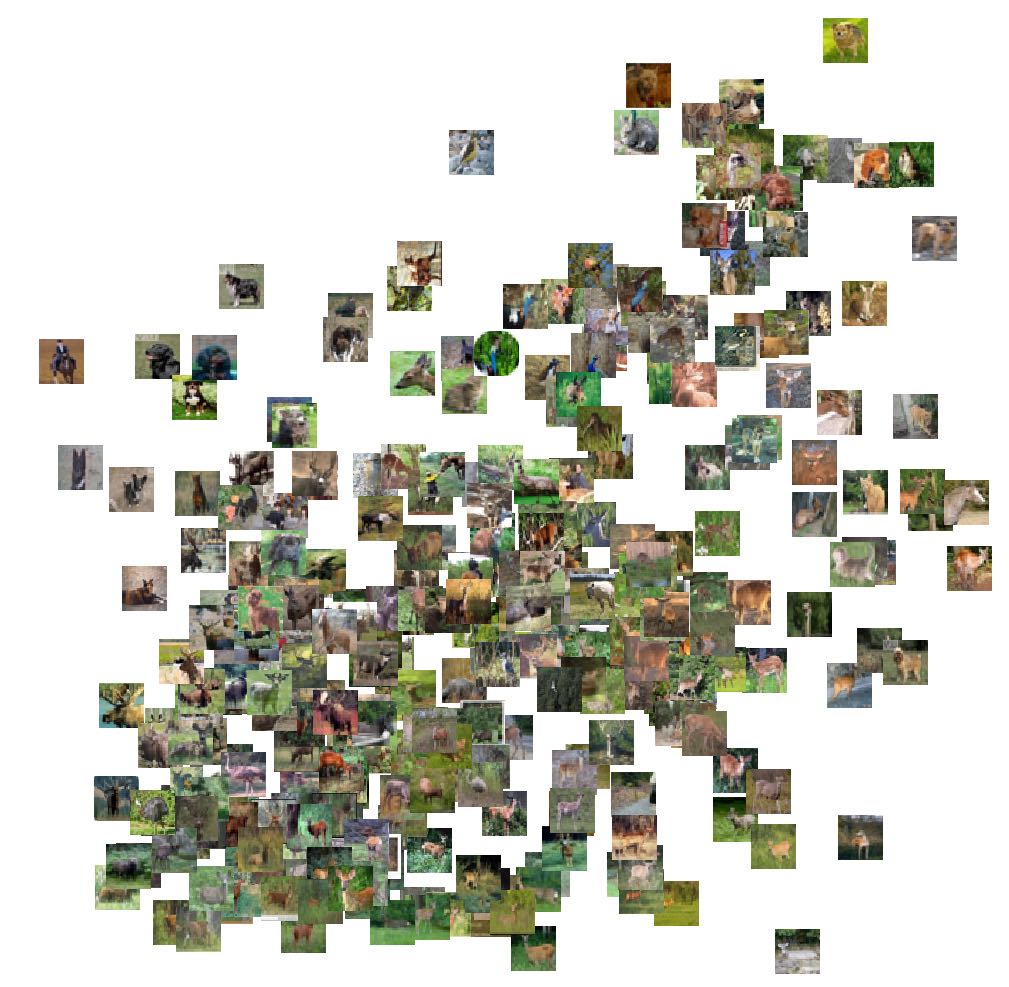} & \includegraphics[width=.25\textwidth]{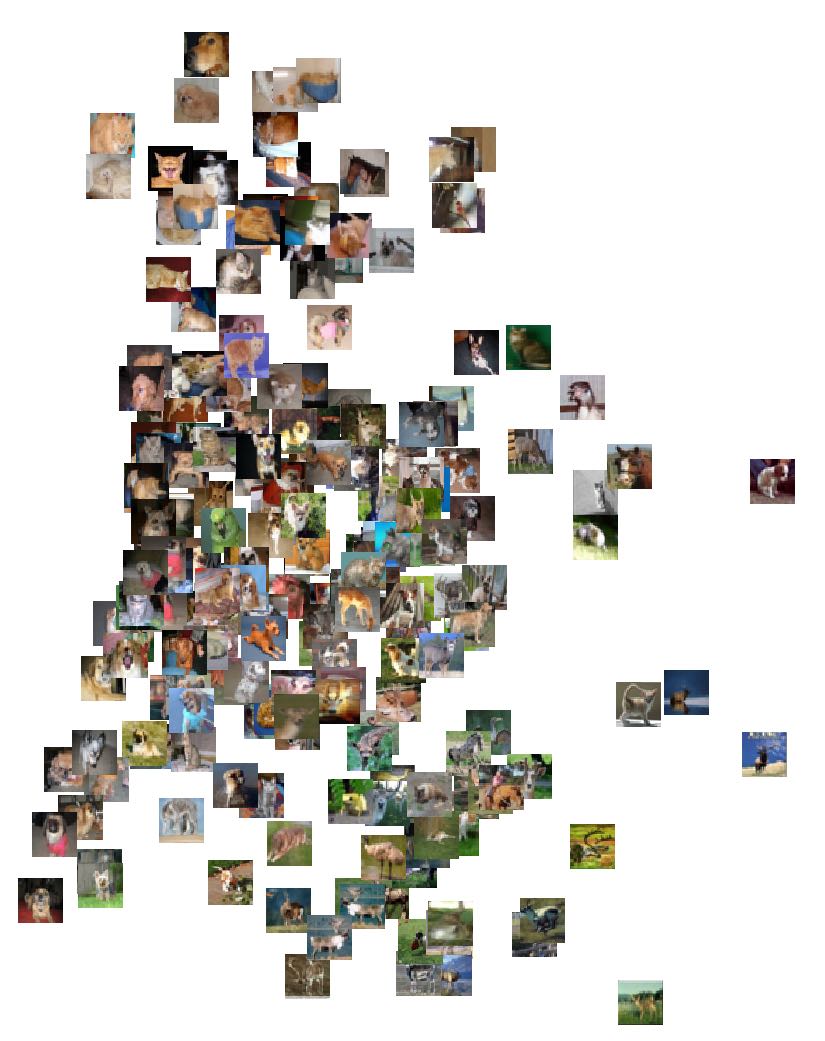} & \includegraphics[width=.25\textwidth]{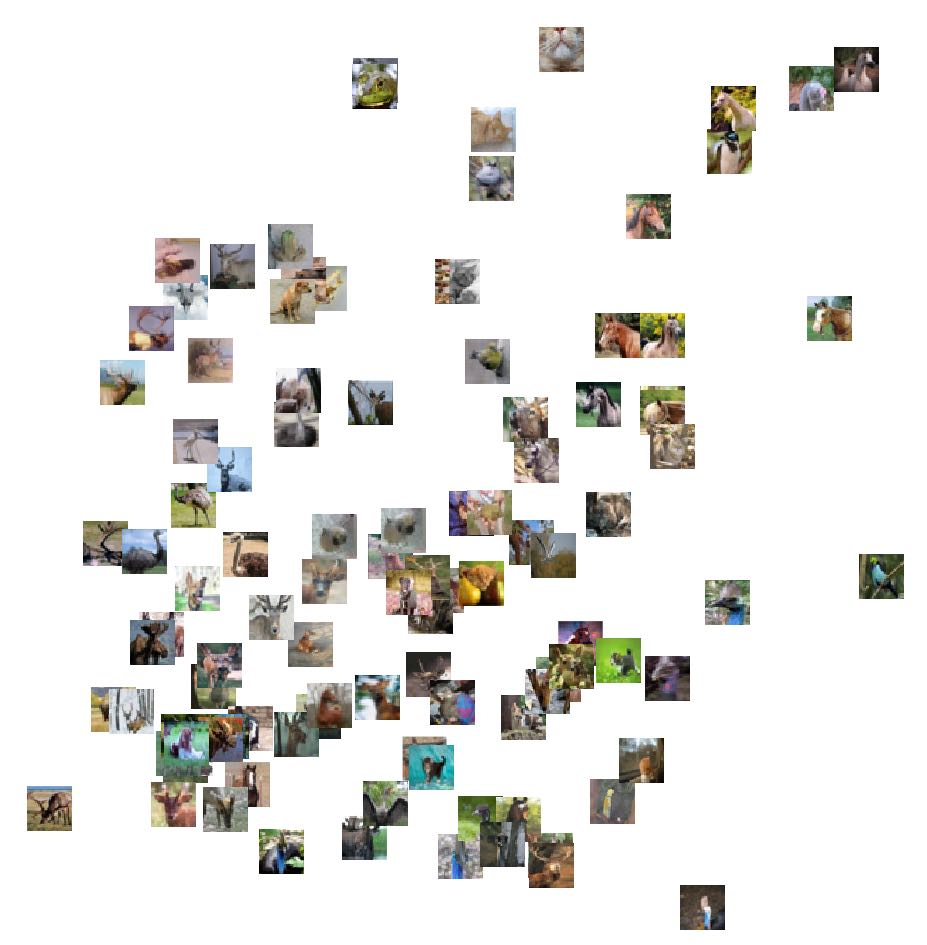} & \includegraphics[width=.25\textwidth]{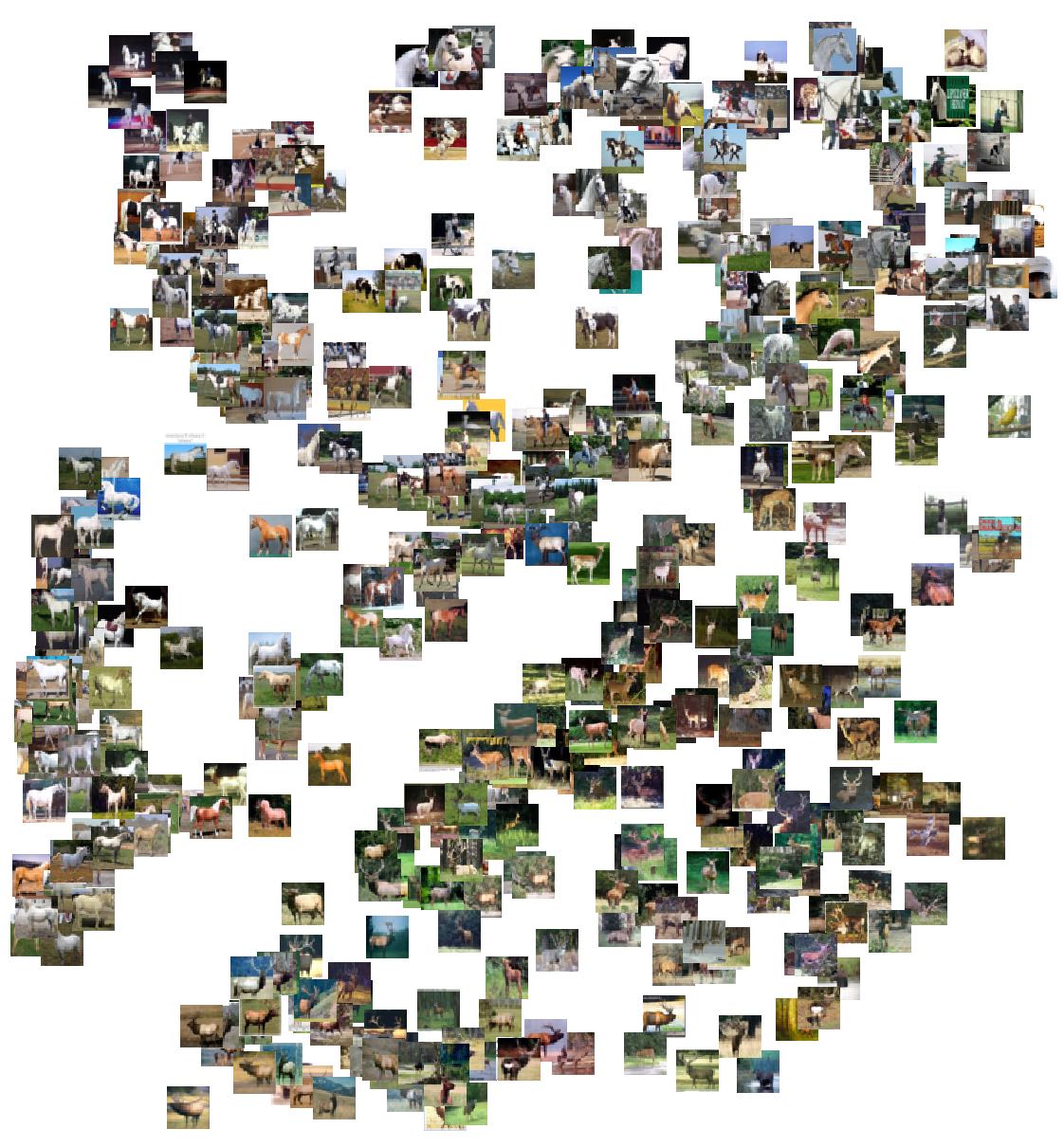}\\
\end{tabular}
\end{table}

\pagebreak

\subsubsection{Fashion MNIIST}
\vspace{-50pt}
We note that in this case only 15 charts are displayed since the network did not end up assigning any points to one of them.
\vspace{-100pt}
\begin{table}[h!]
\begin{tabular}{cccc}
\includegraphics[width=.25\textwidth]{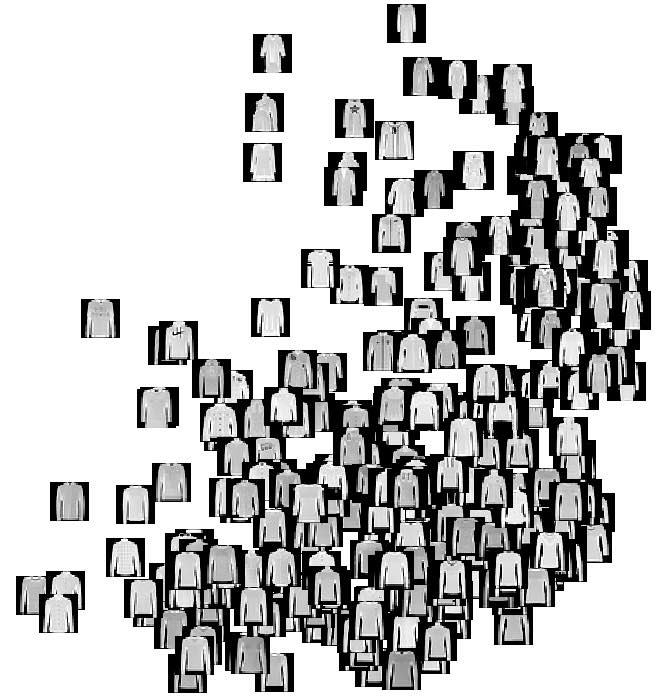} & \includegraphics[width=.25\textwidth]{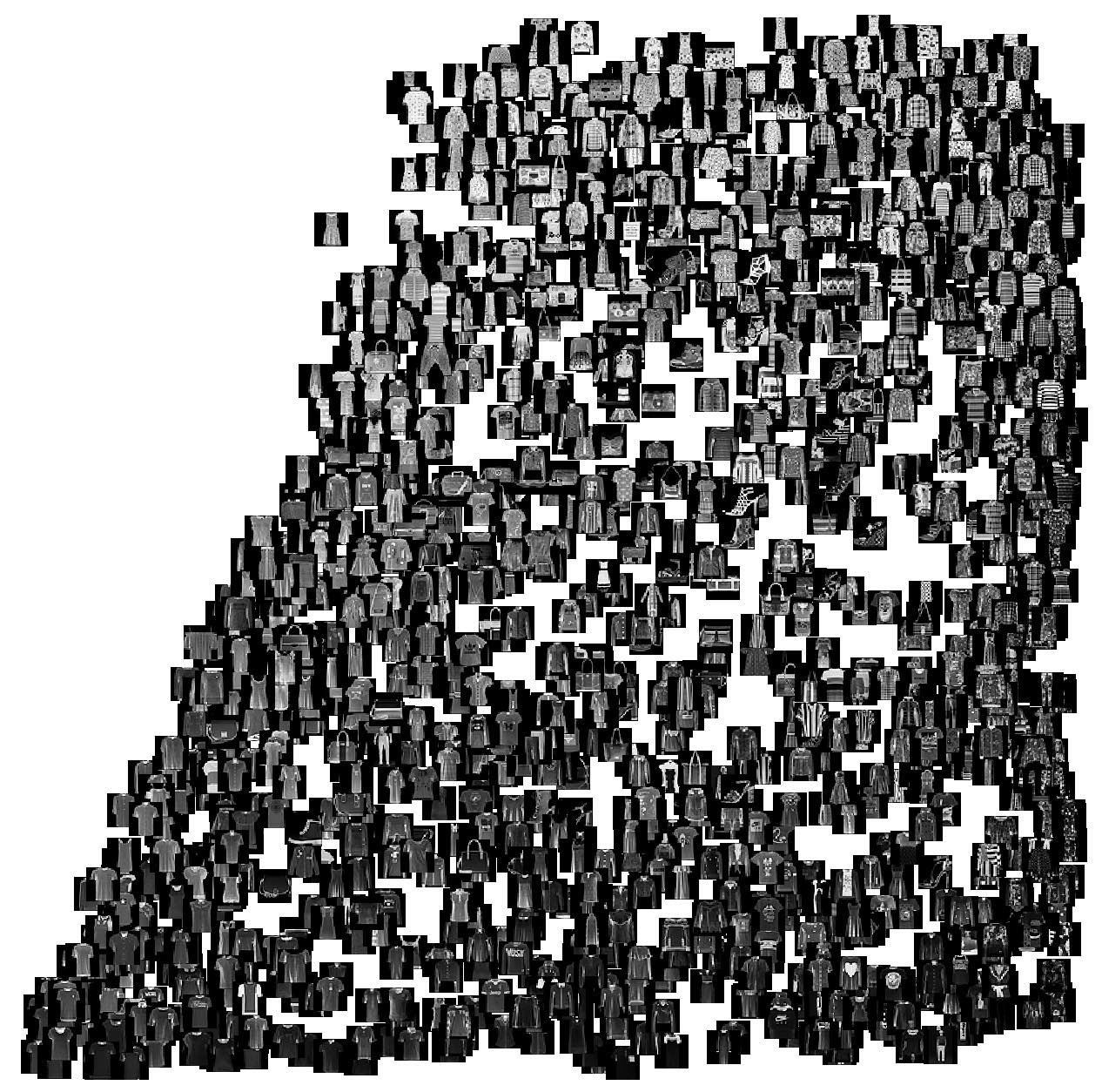} & \includegraphics[width=.25\textwidth]{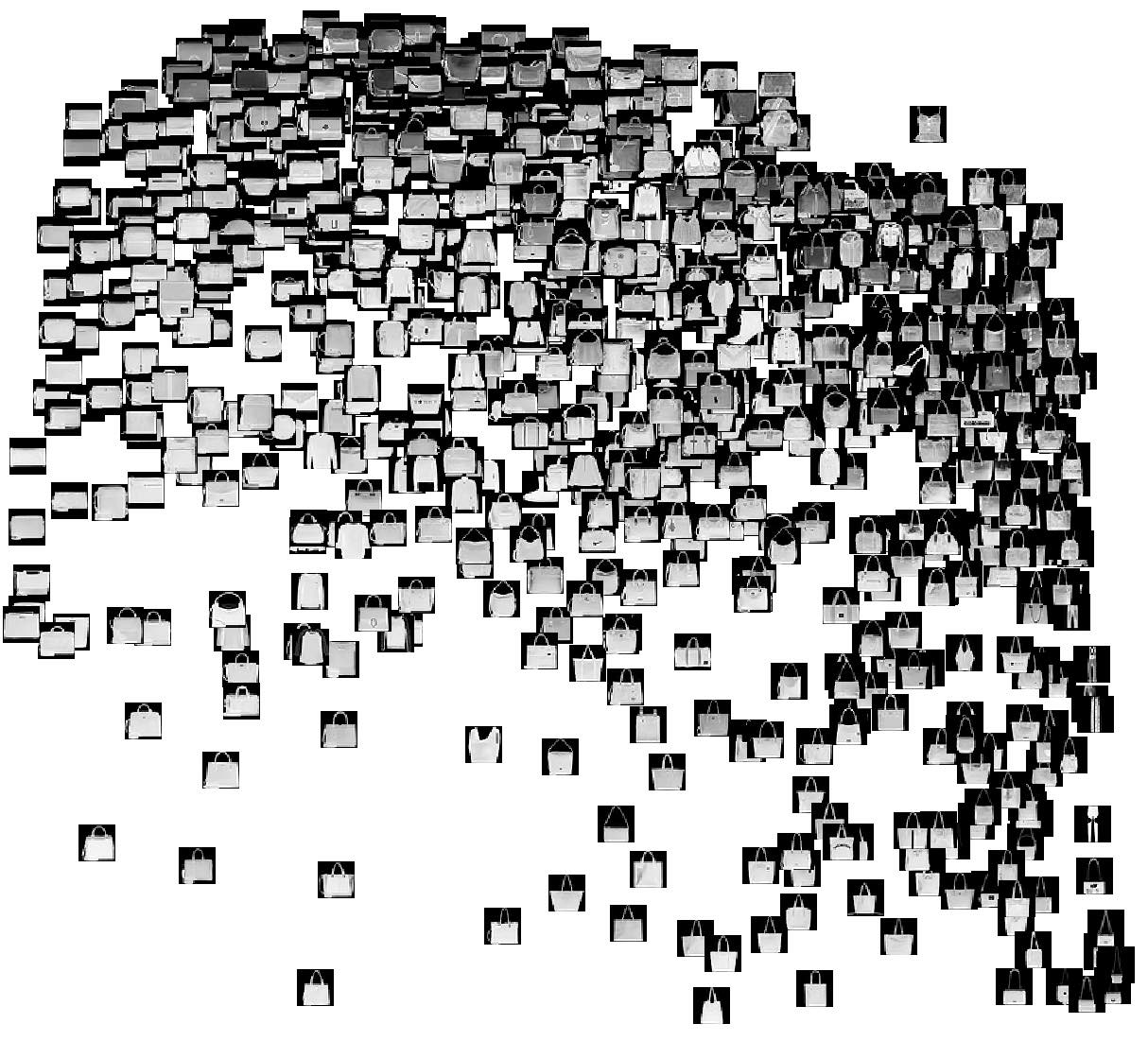} & \includegraphics[width=.25\textwidth]{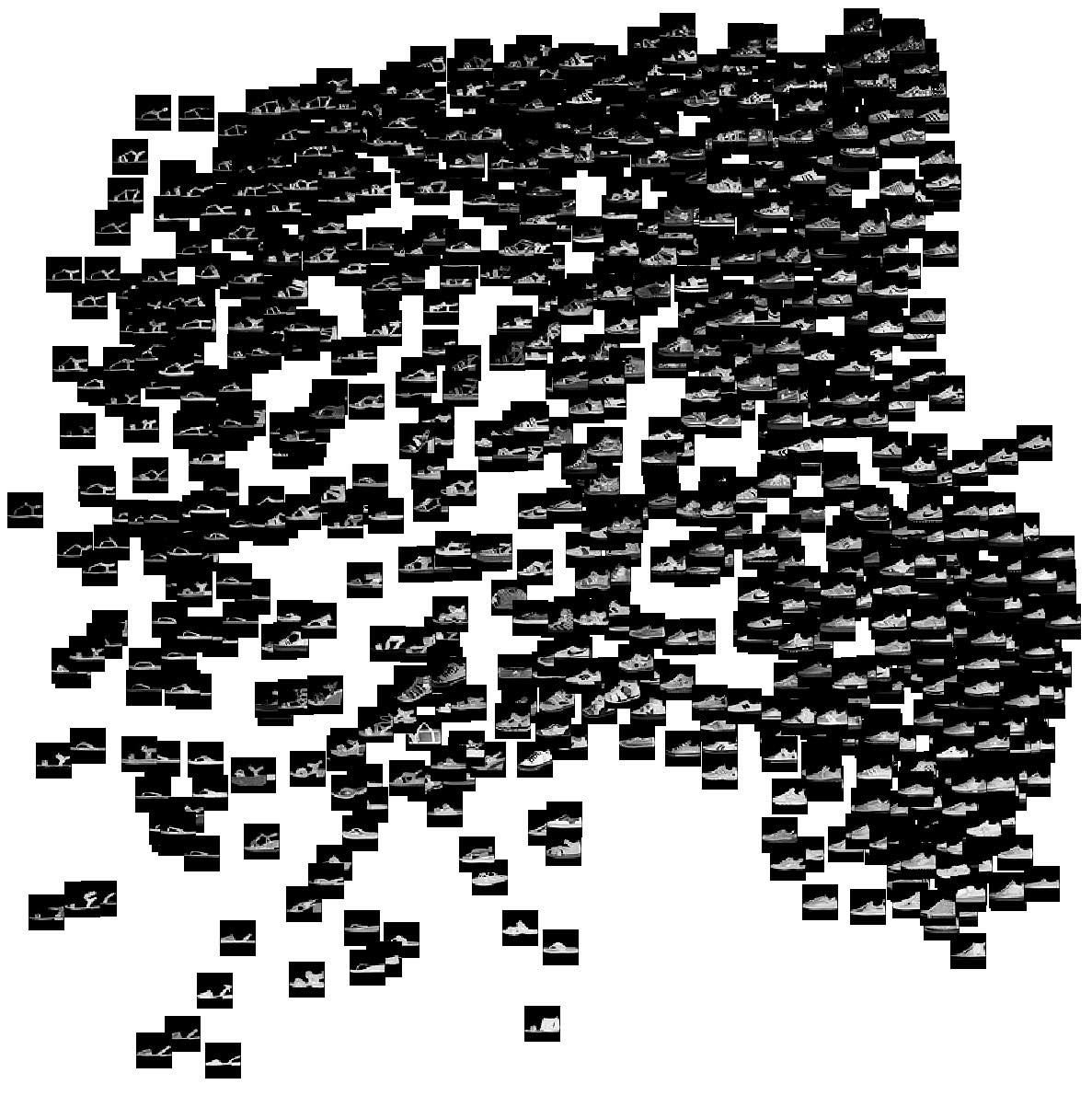}\\
\includegraphics[width=.25\textwidth]{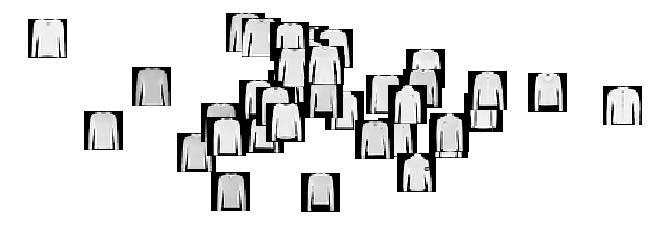} & \includegraphics[width=.25\textwidth]{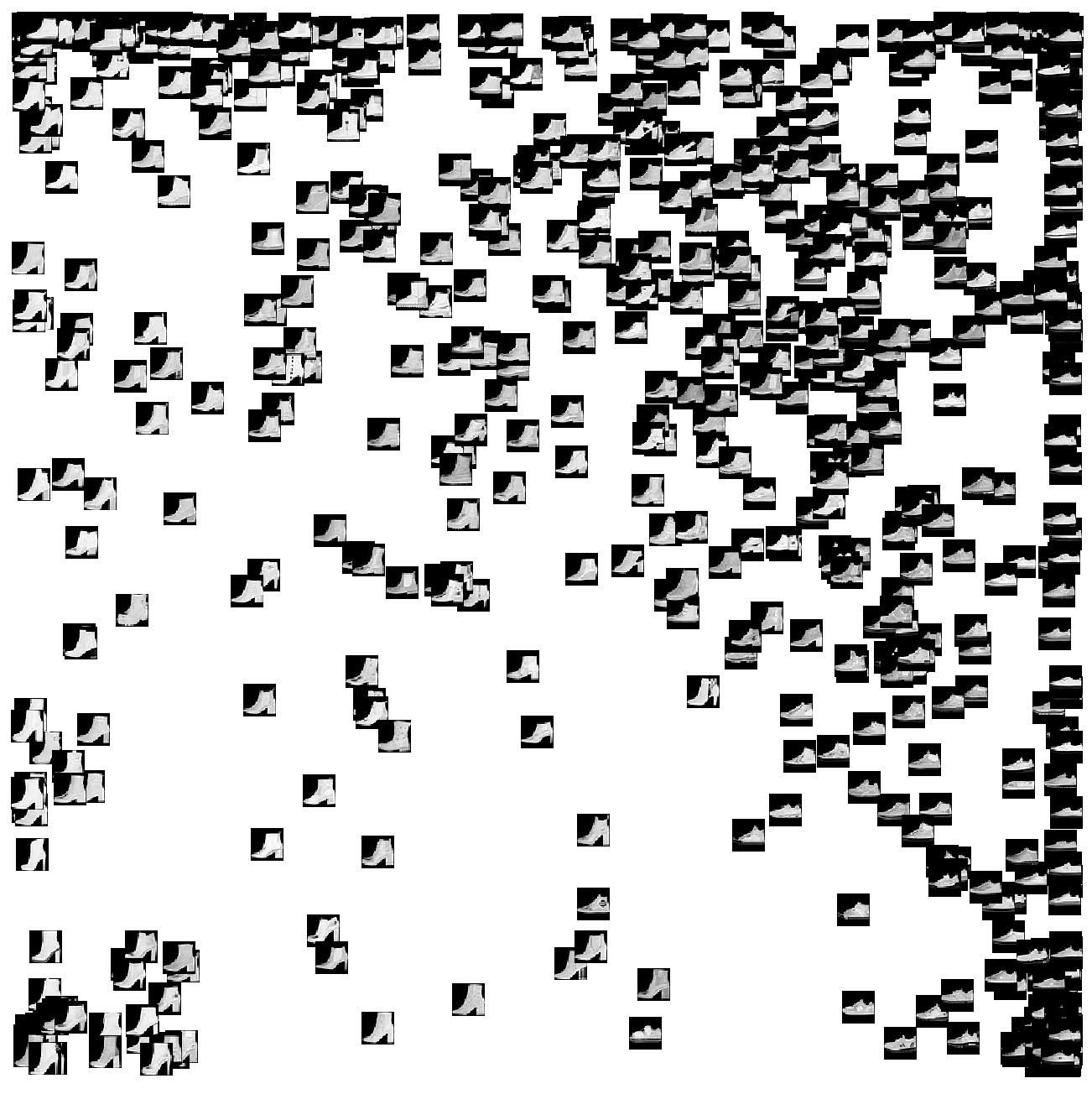} & \includegraphics[width=.25\textwidth]{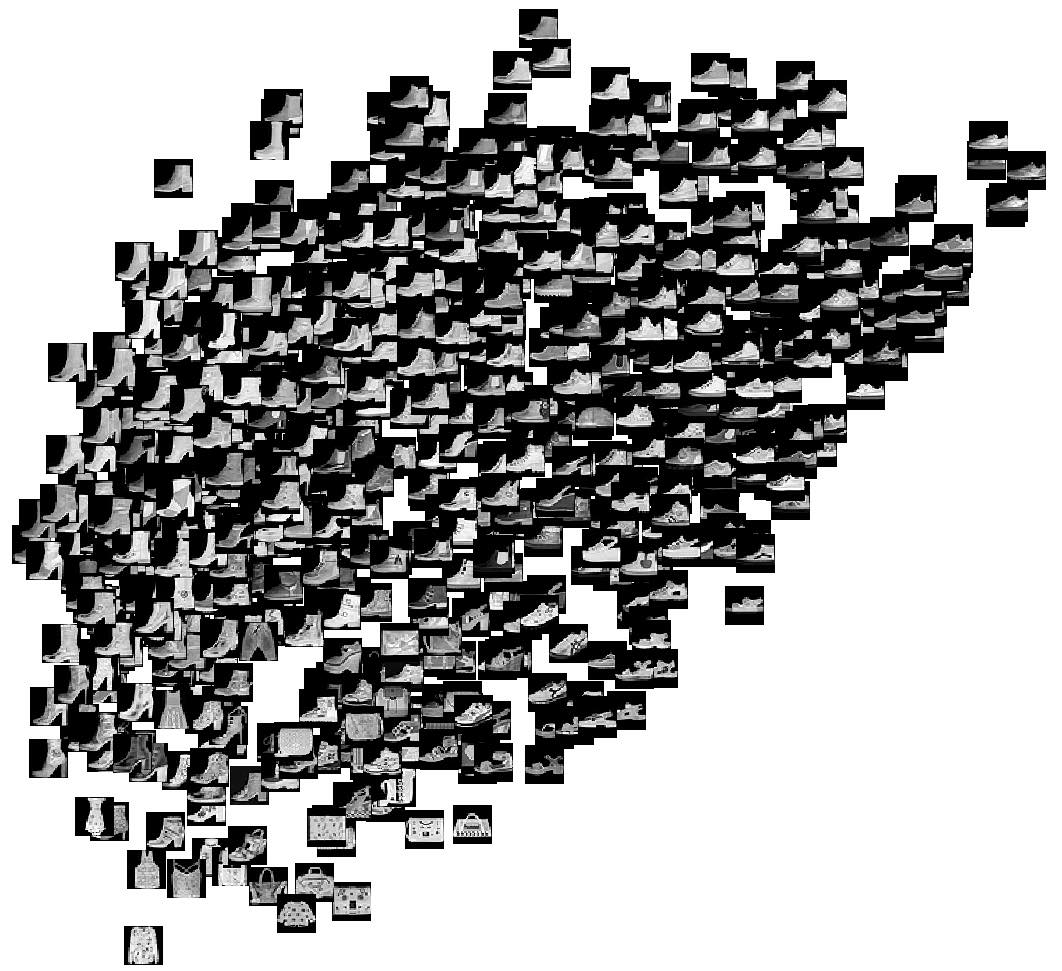} & \includegraphics[width=.25\textwidth]{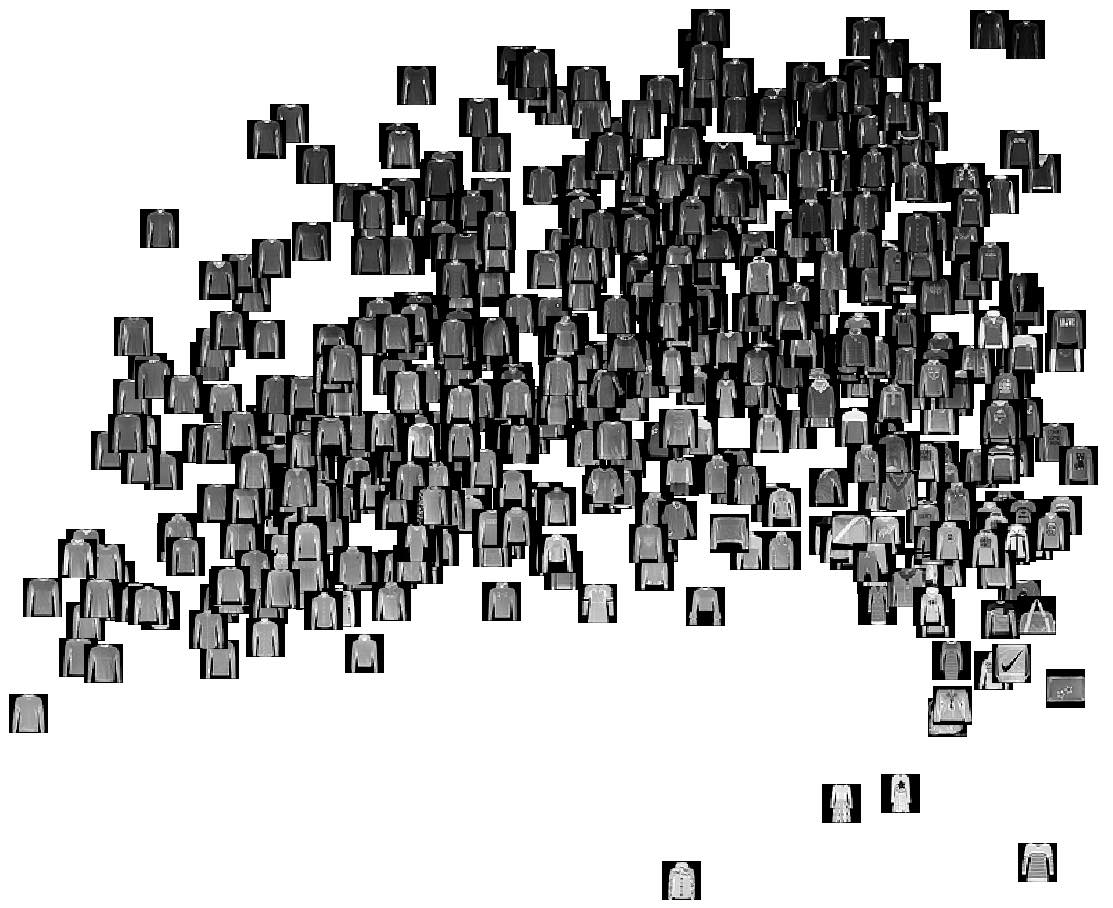}\\
\includegraphics[width=.25\textwidth]{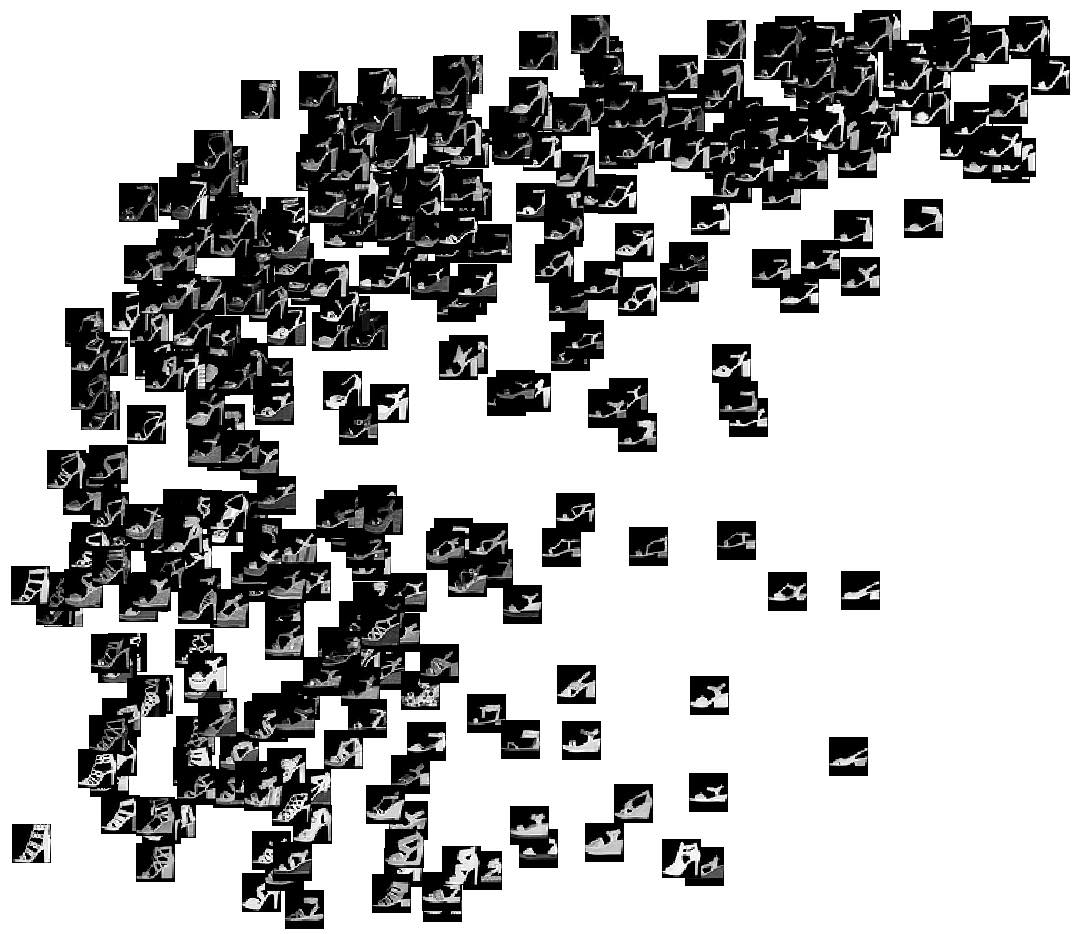} & \includegraphics[width=.25\textwidth]{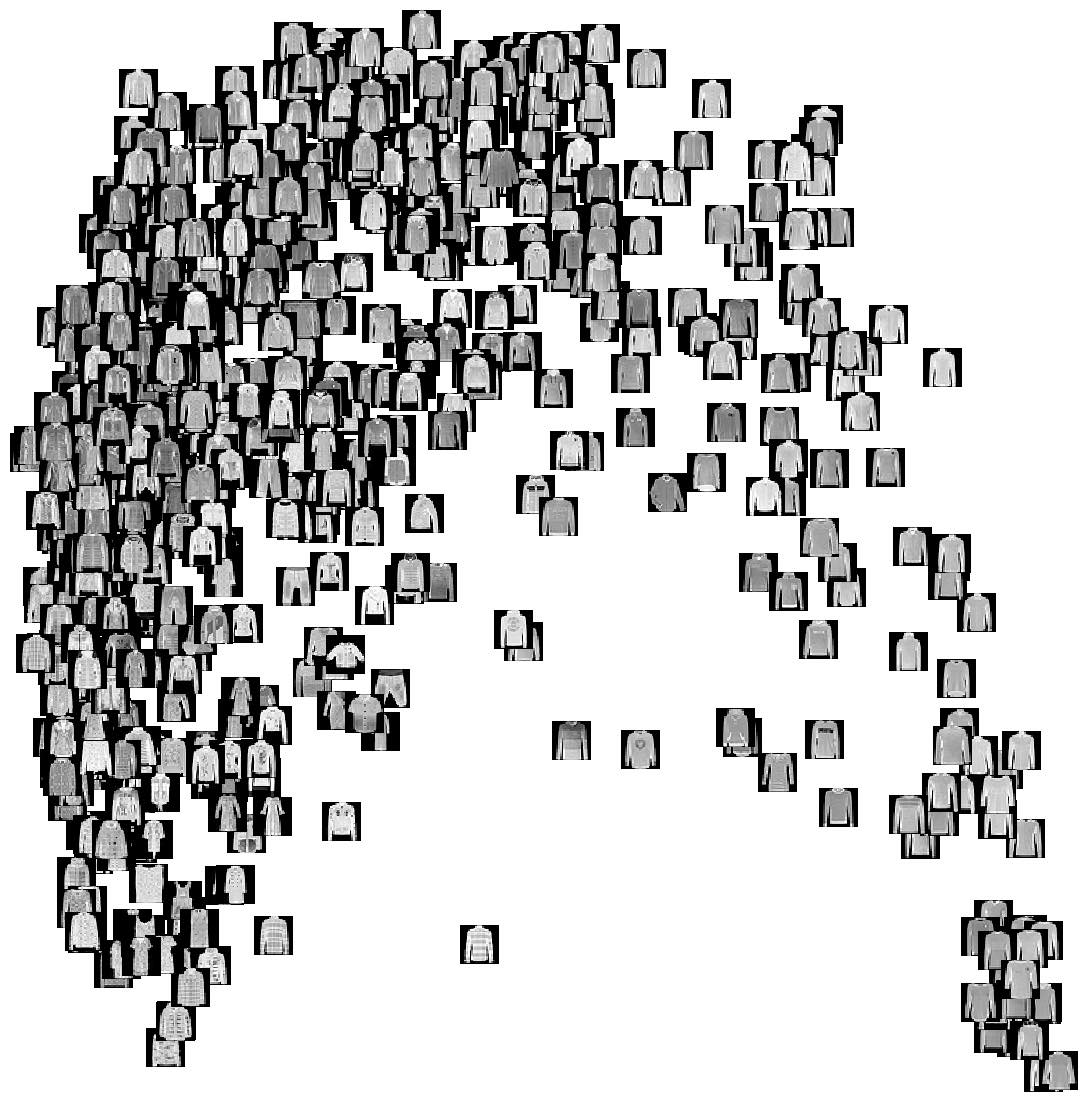} & \includegraphics[width=.25\textwidth]{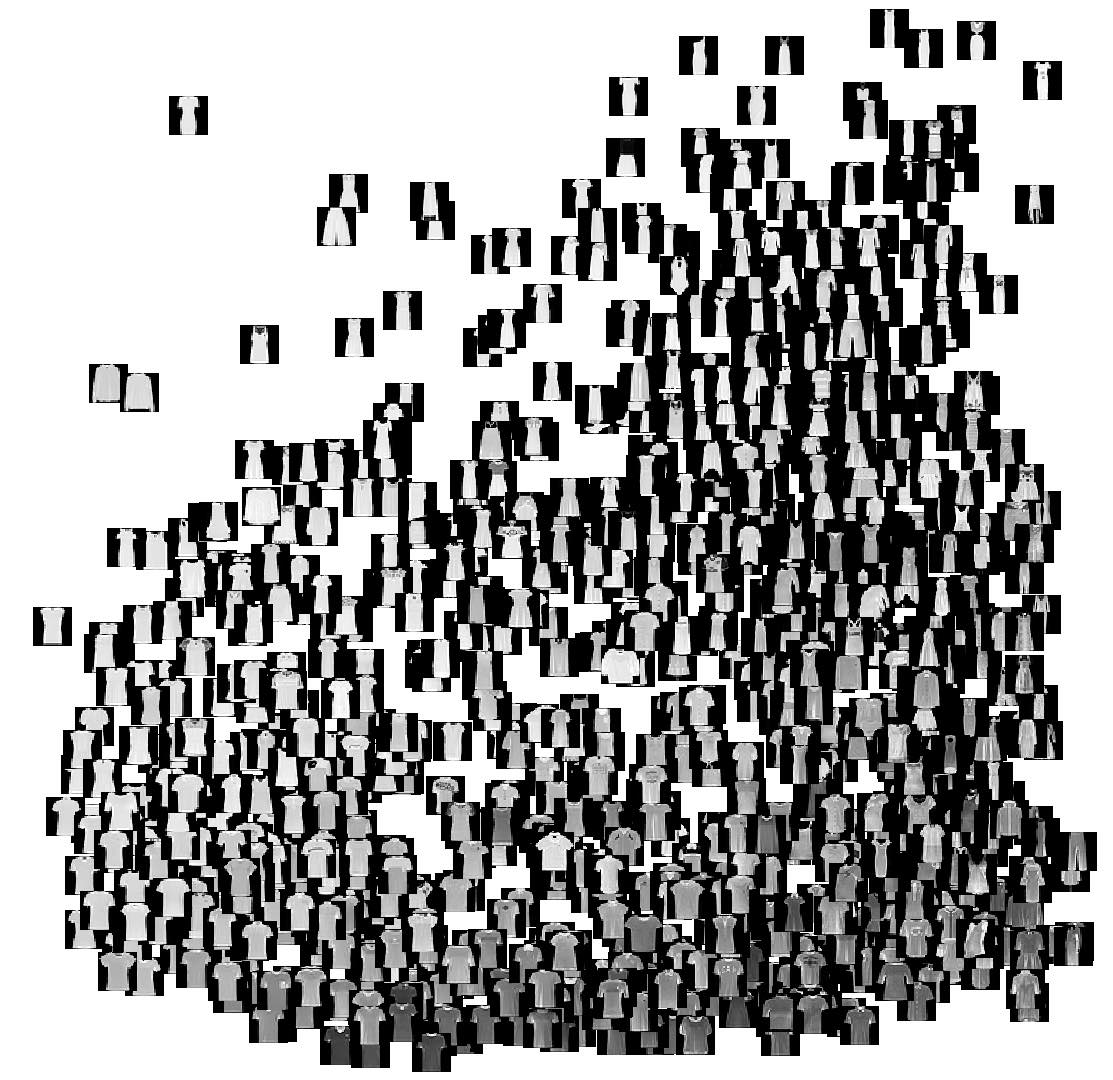} & \includegraphics[width=.25\textwidth]{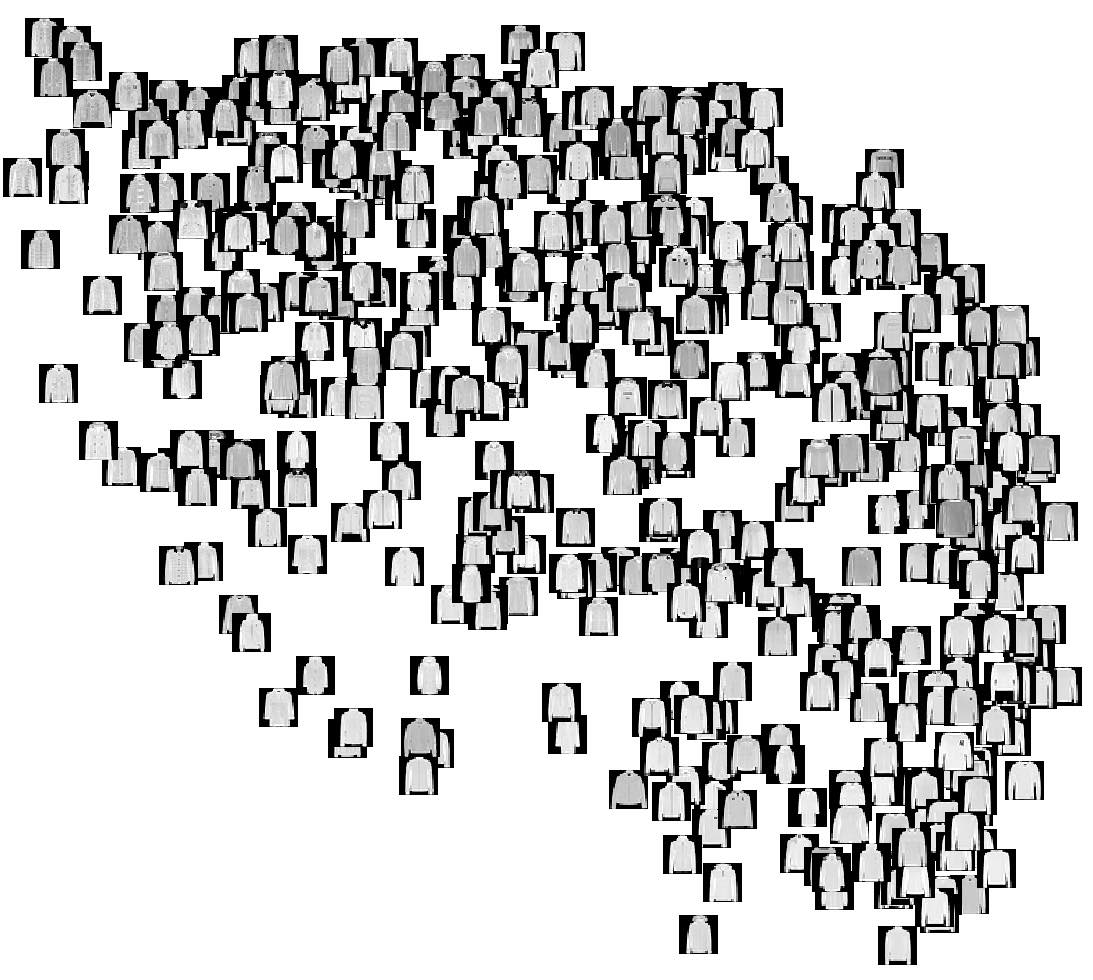}\\
\includegraphics[width=.25\textwidth]{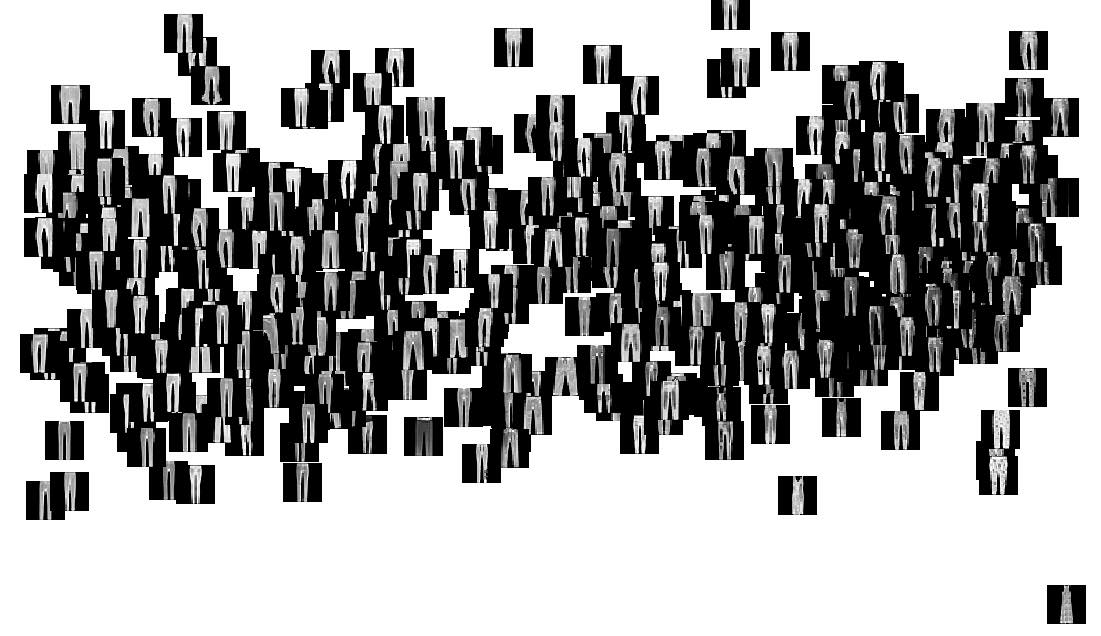} & \includegraphics[width=.25\textwidth]{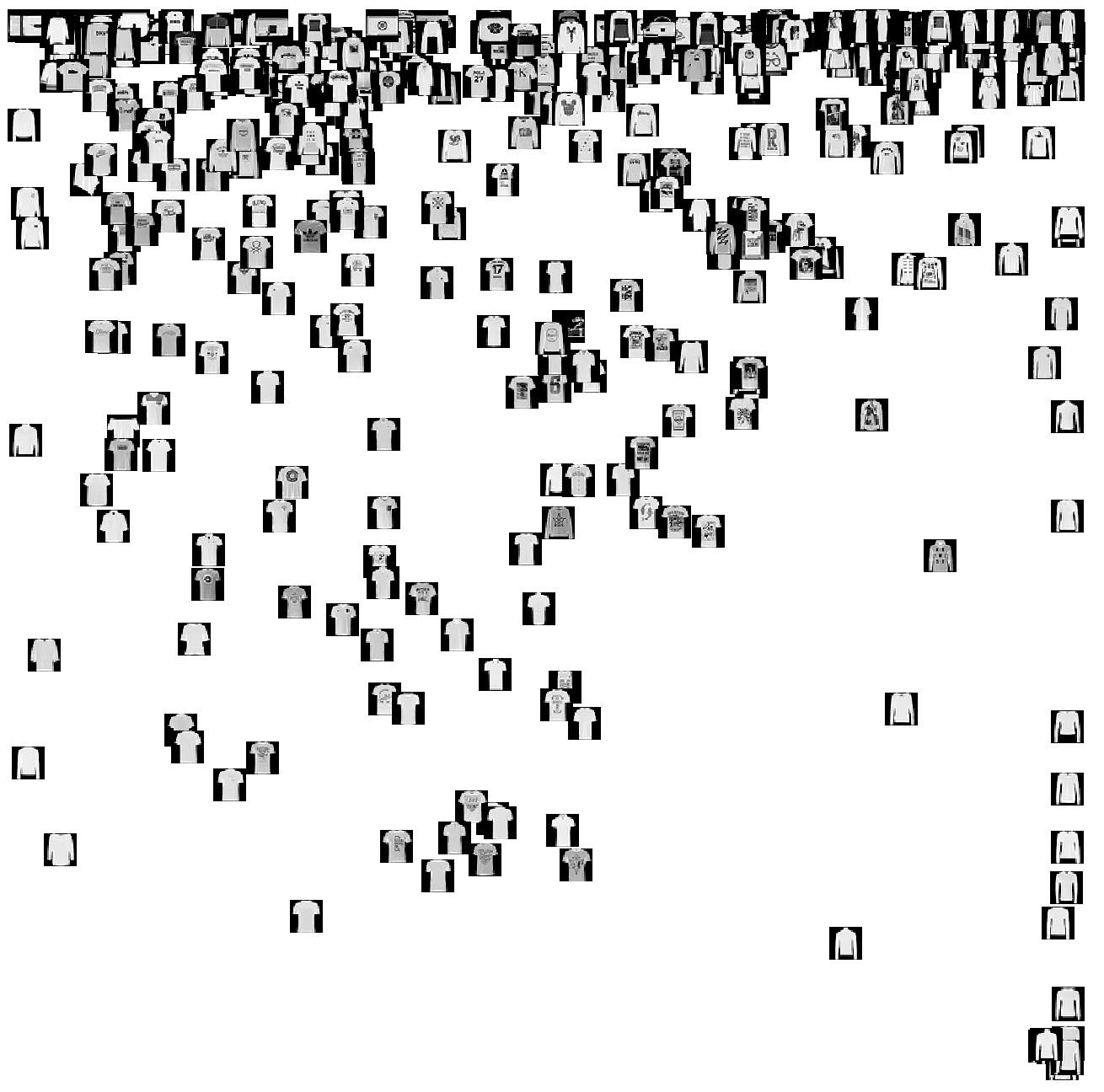} & \includegraphics[width=.25\textwidth]{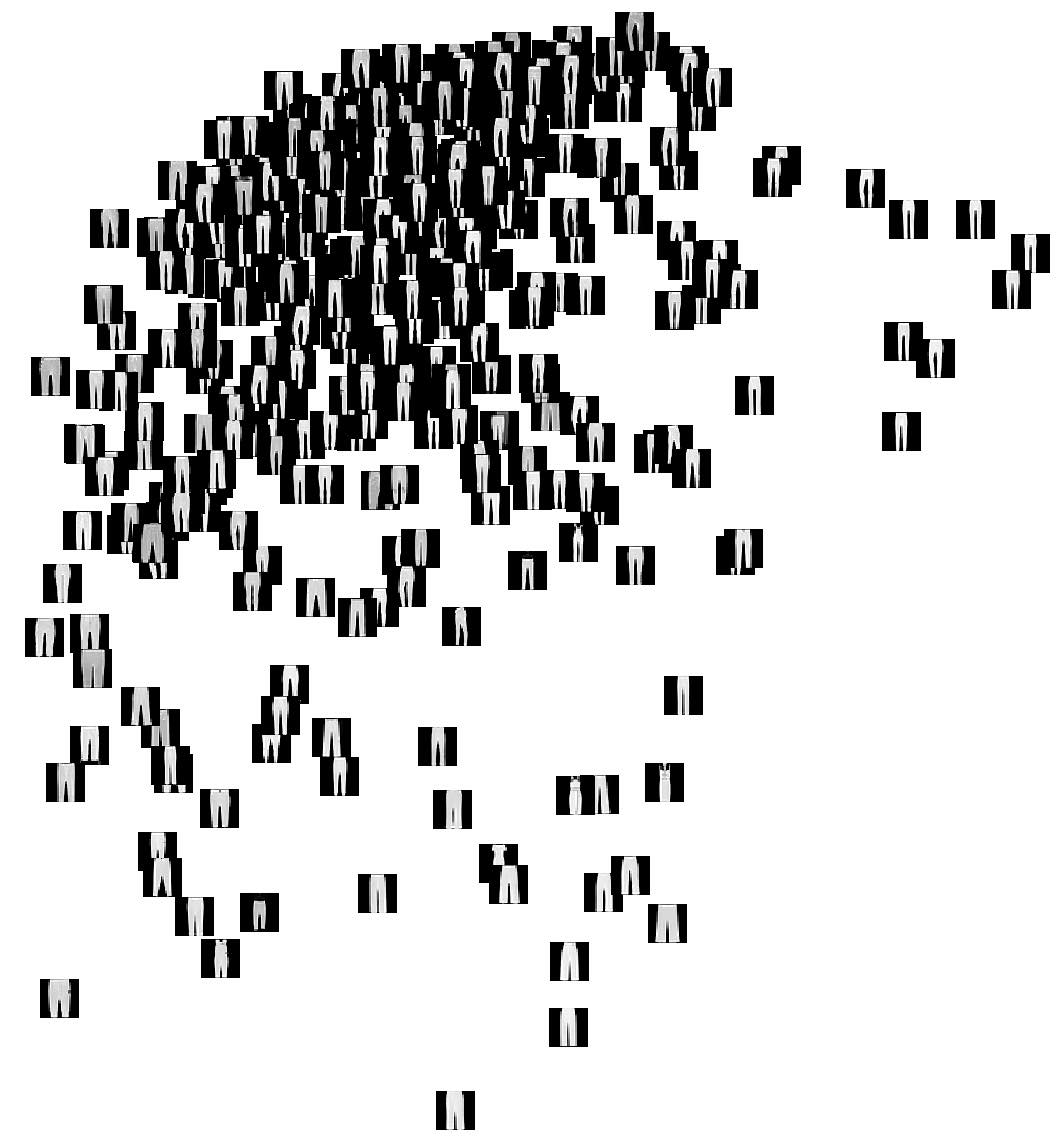} & \\
\end{tabular}
\end{table}

\pagebreak

\subsubsection{MNIST}
\vspace{-210pt}
\begin{table}[h!]
\begin{tabular}{cccc}
\includegraphics[width=.25\textwidth]{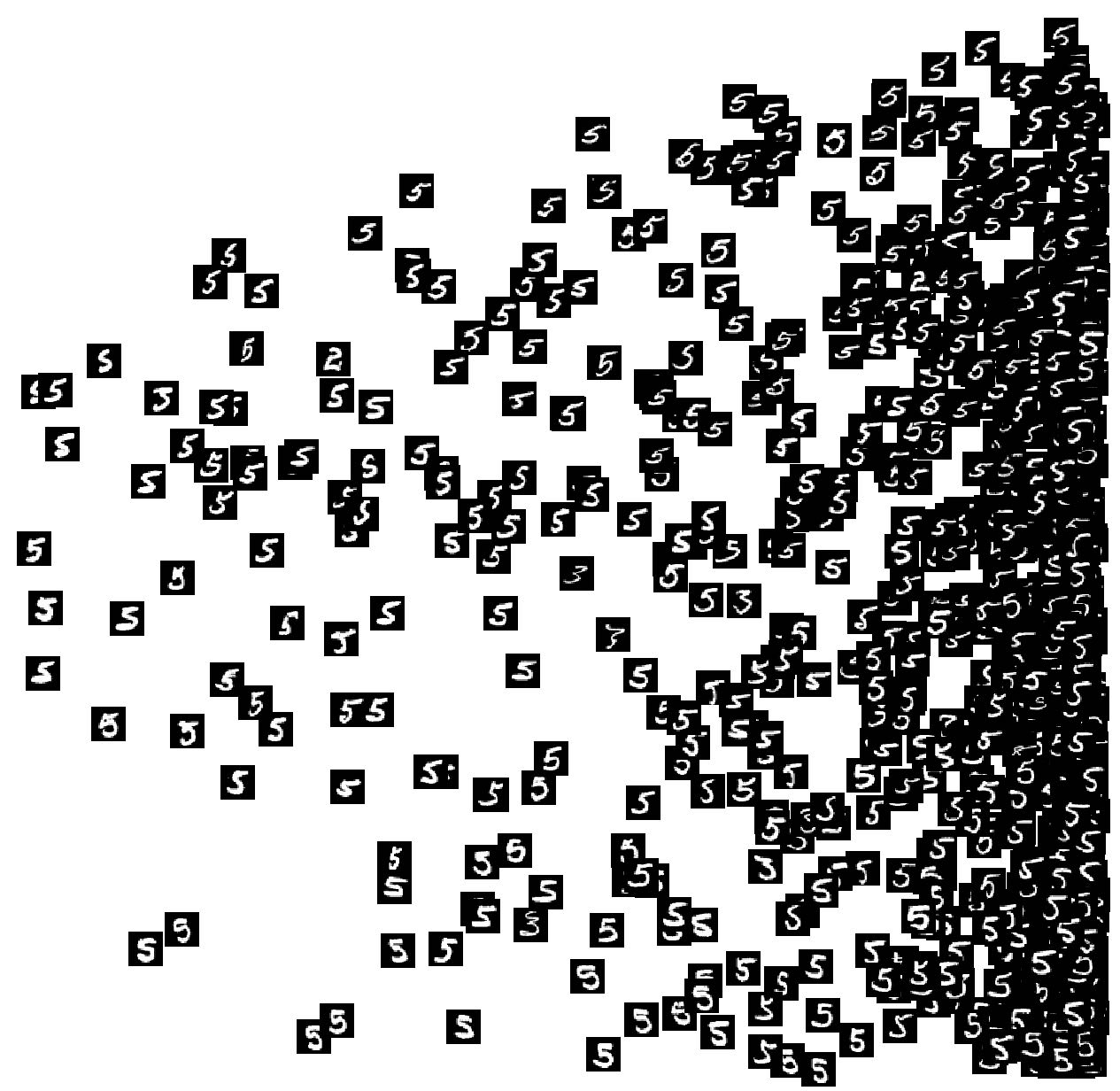} & \includegraphics[width=.25\textwidth]{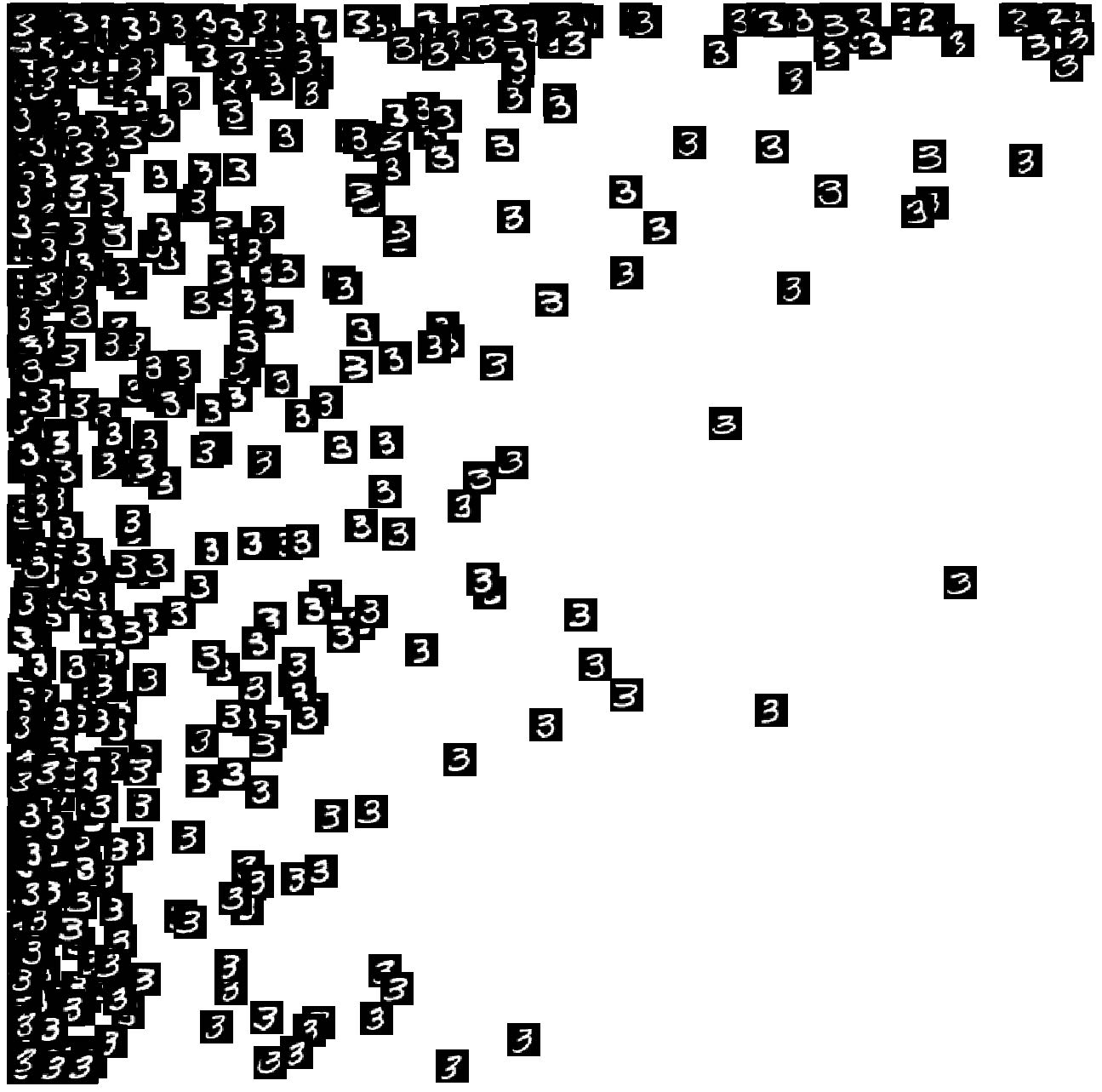} & \includegraphics[width=.25\textwidth]{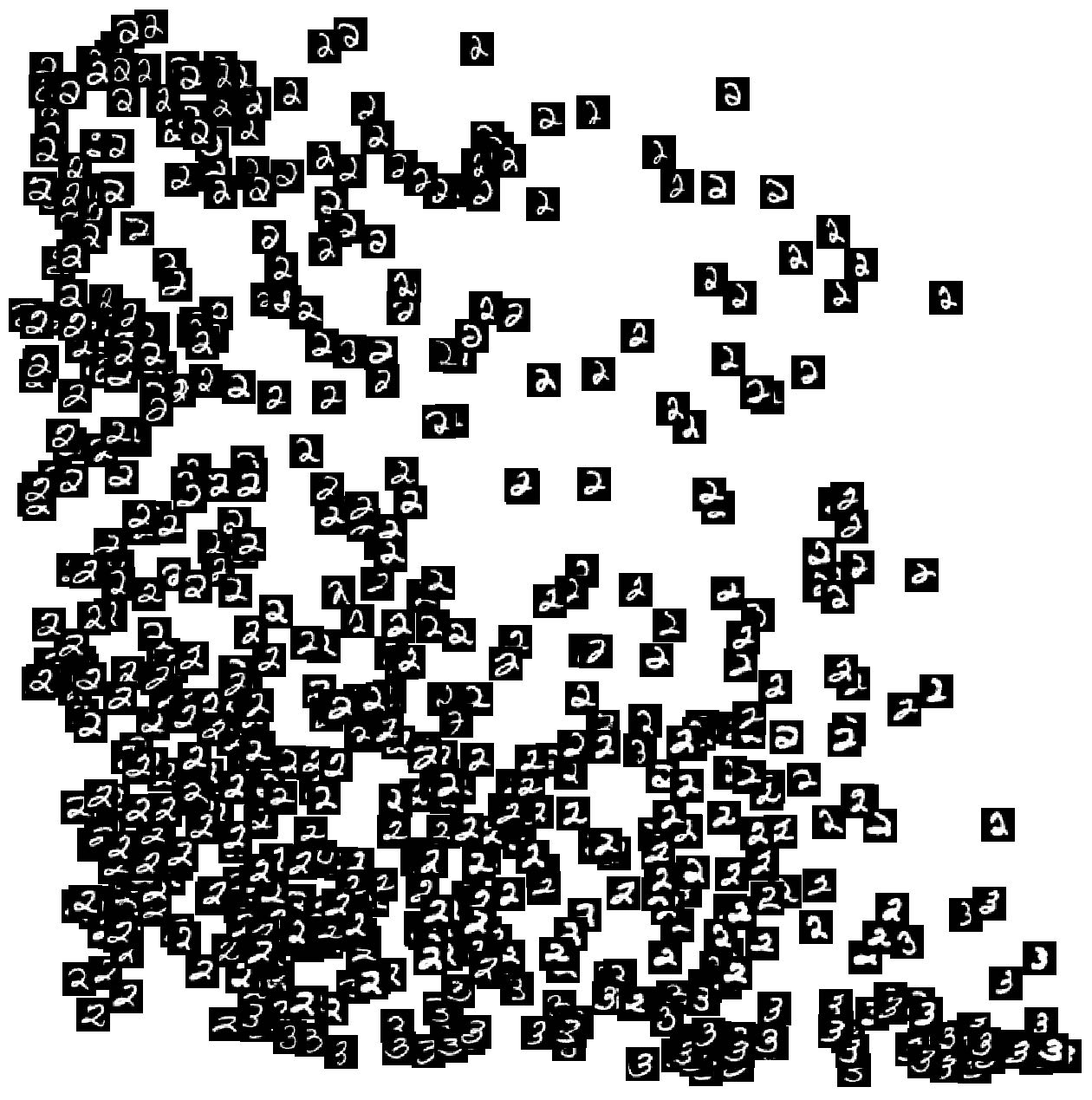} & \includegraphics[width=.25\textwidth]{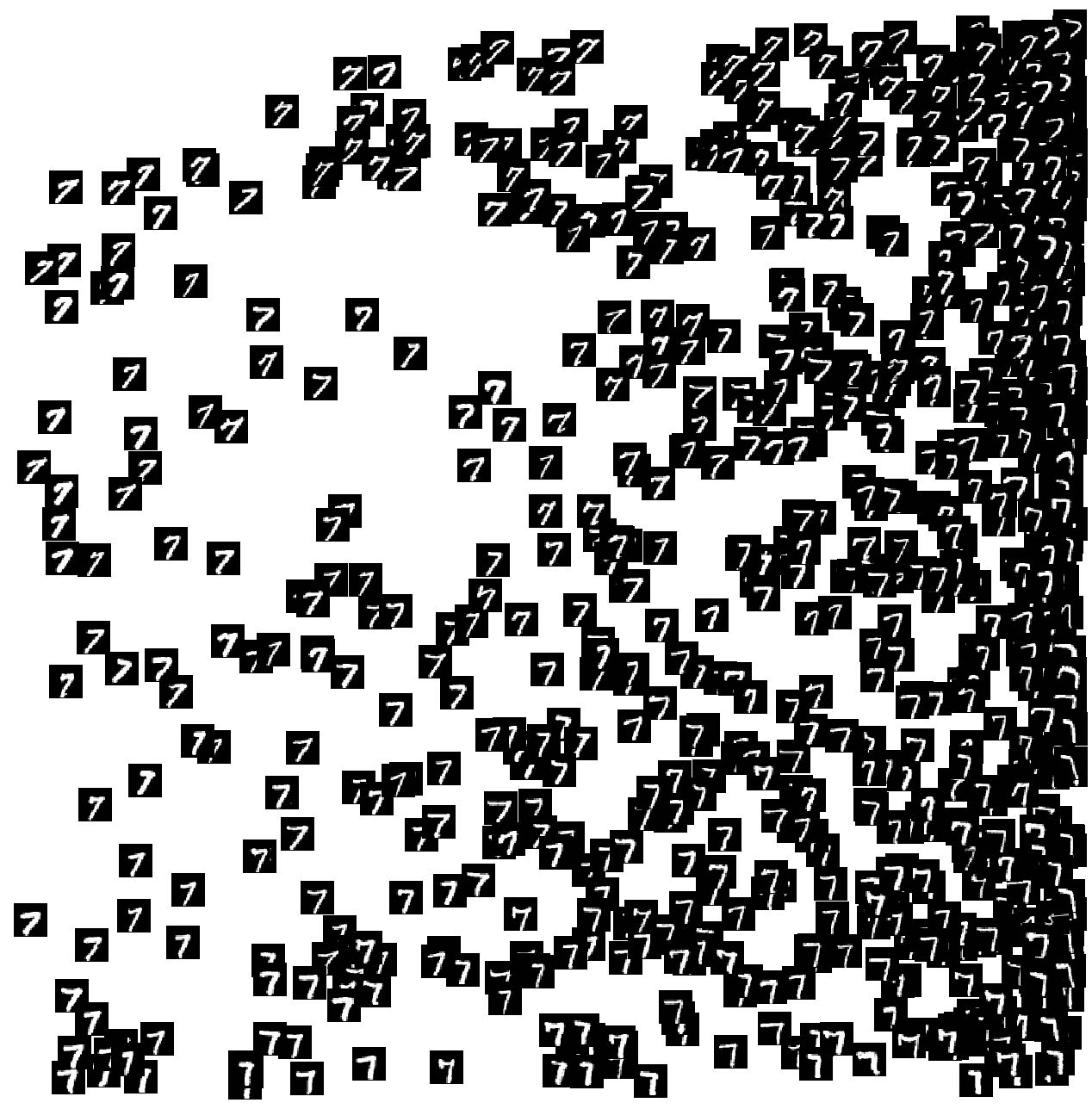}\\
\includegraphics[width=.25\textwidth]{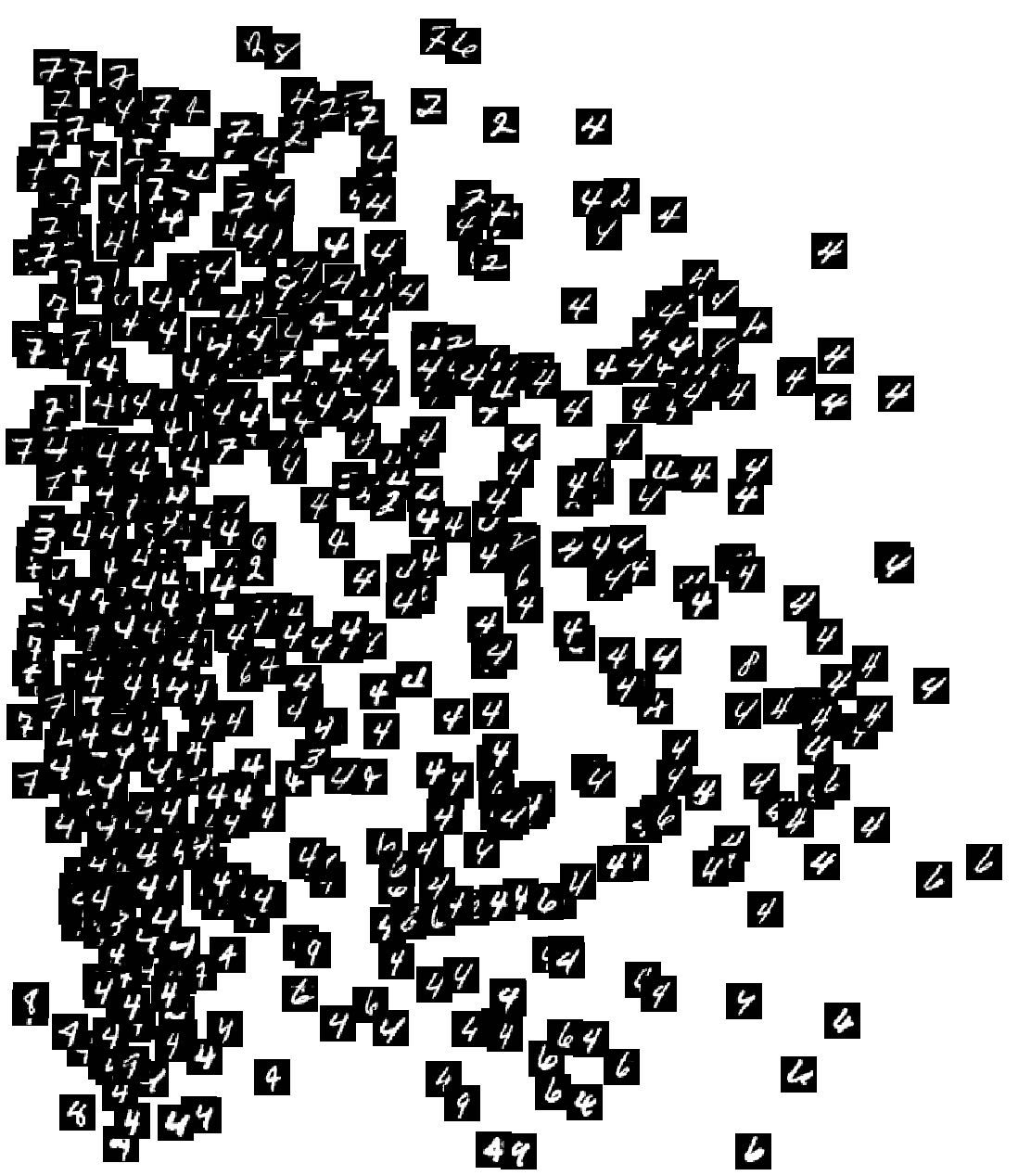} & \includegraphics[width=.25\textwidth]{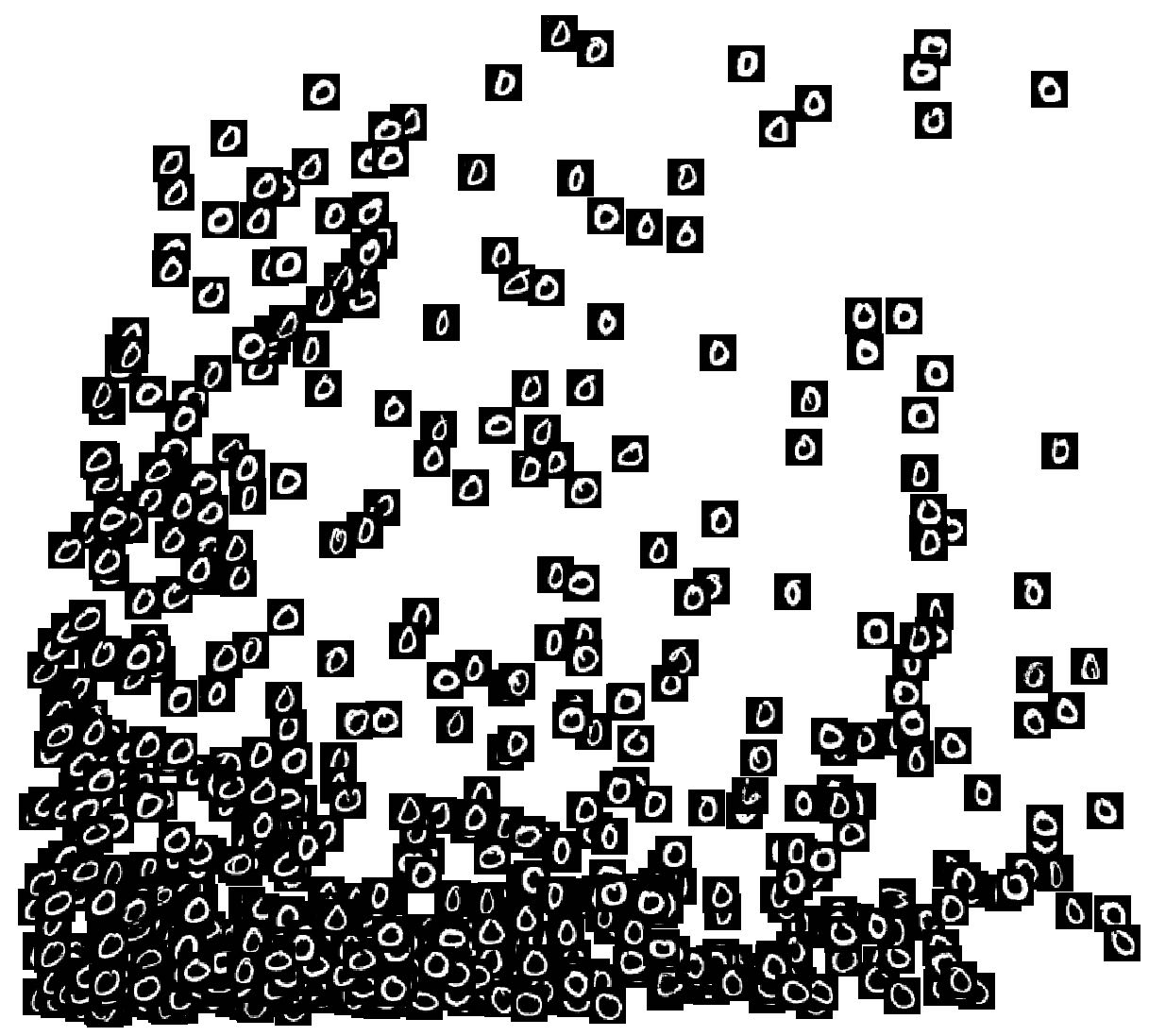} & \includegraphics[width=.25\textwidth]{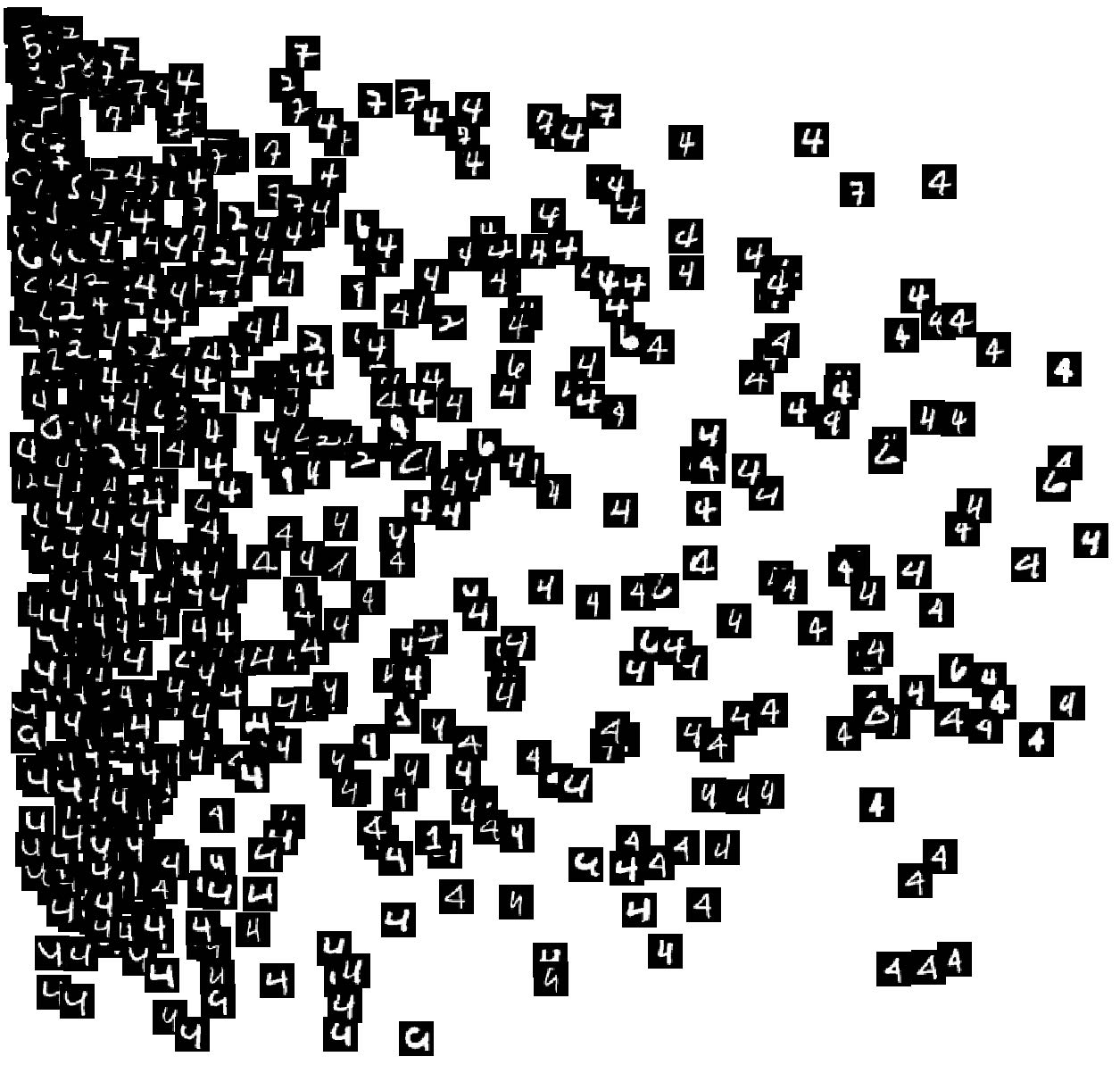} & \includegraphics[width=.25\textwidth]{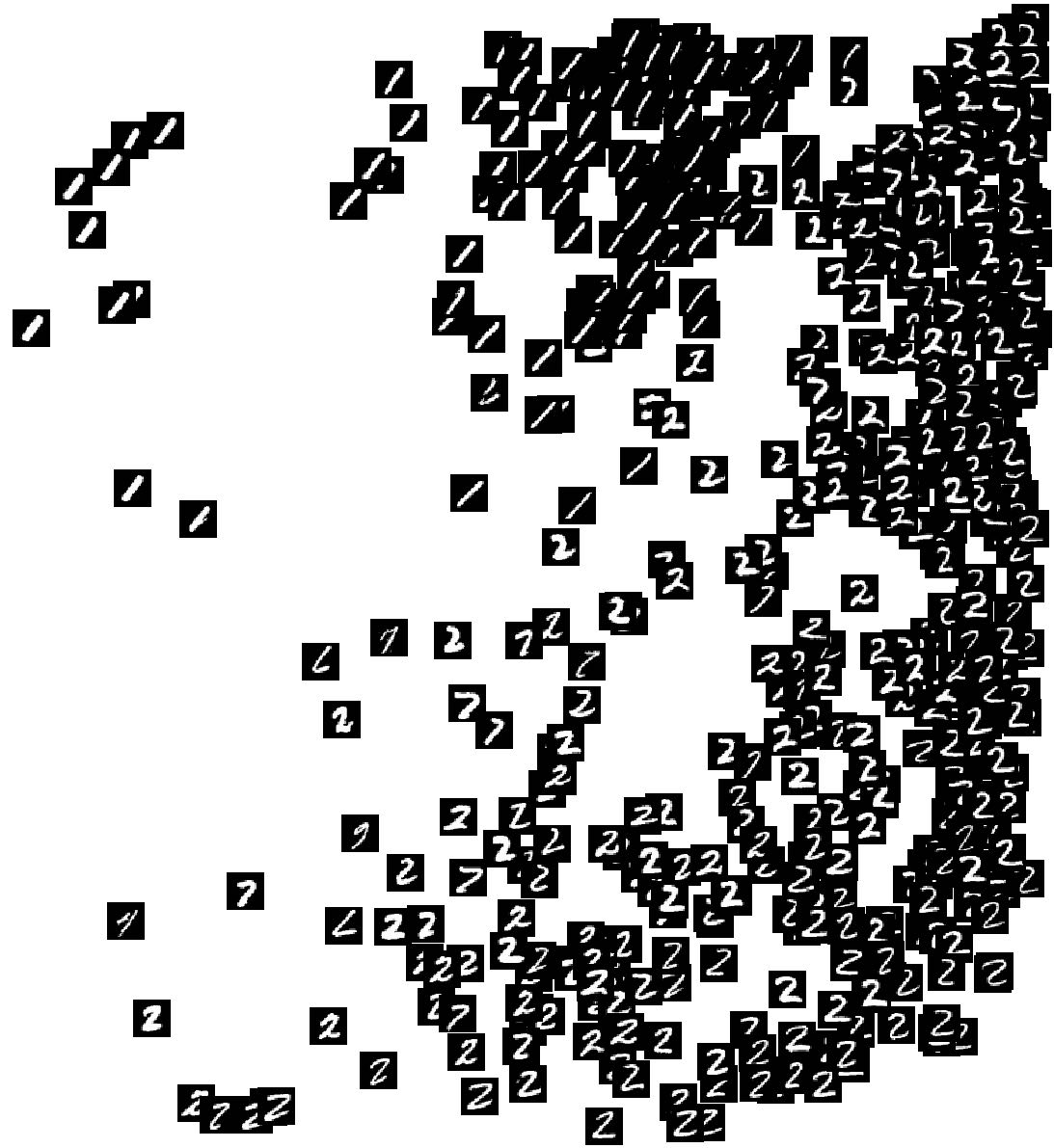}\\
\includegraphics[width=.25\textwidth]{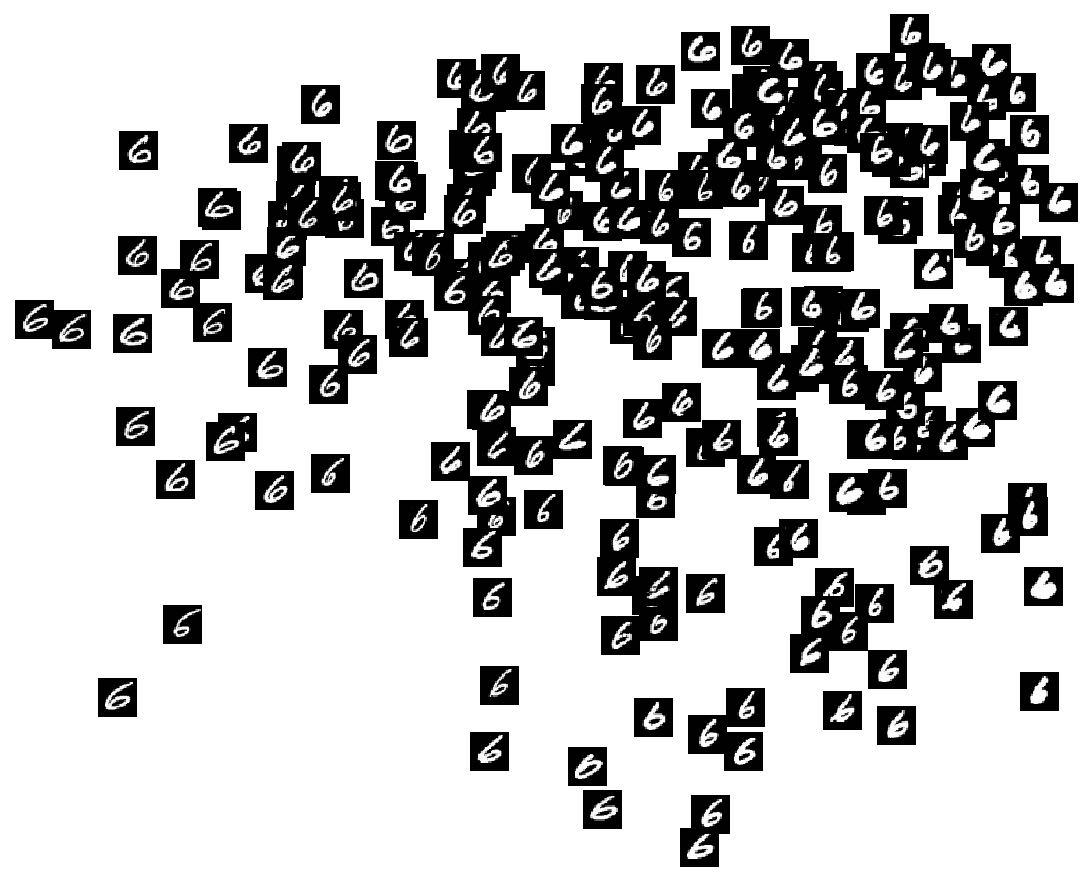} & \includegraphics[width=.25\textwidth]{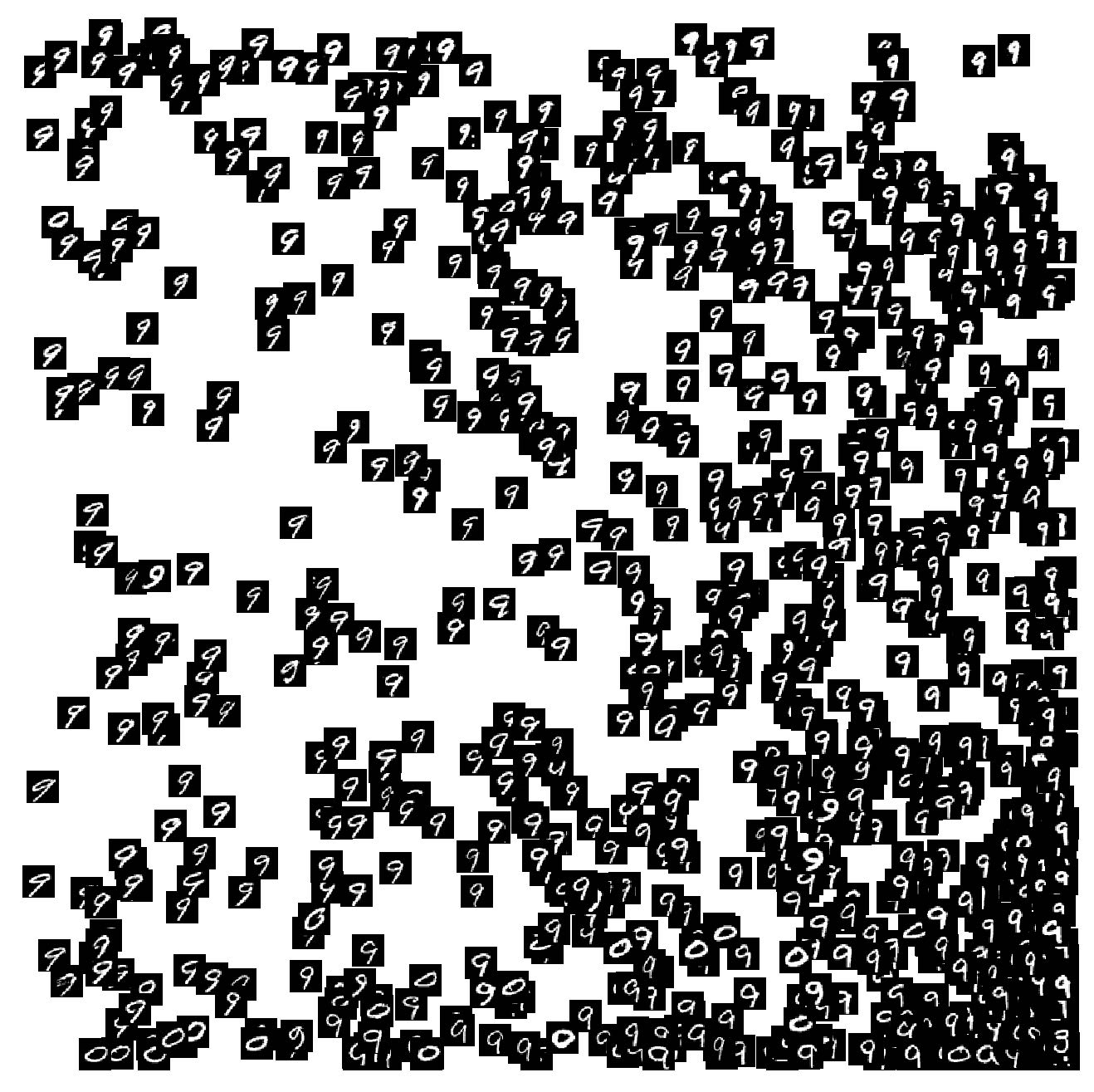} & \includegraphics[width=.25\textwidth]{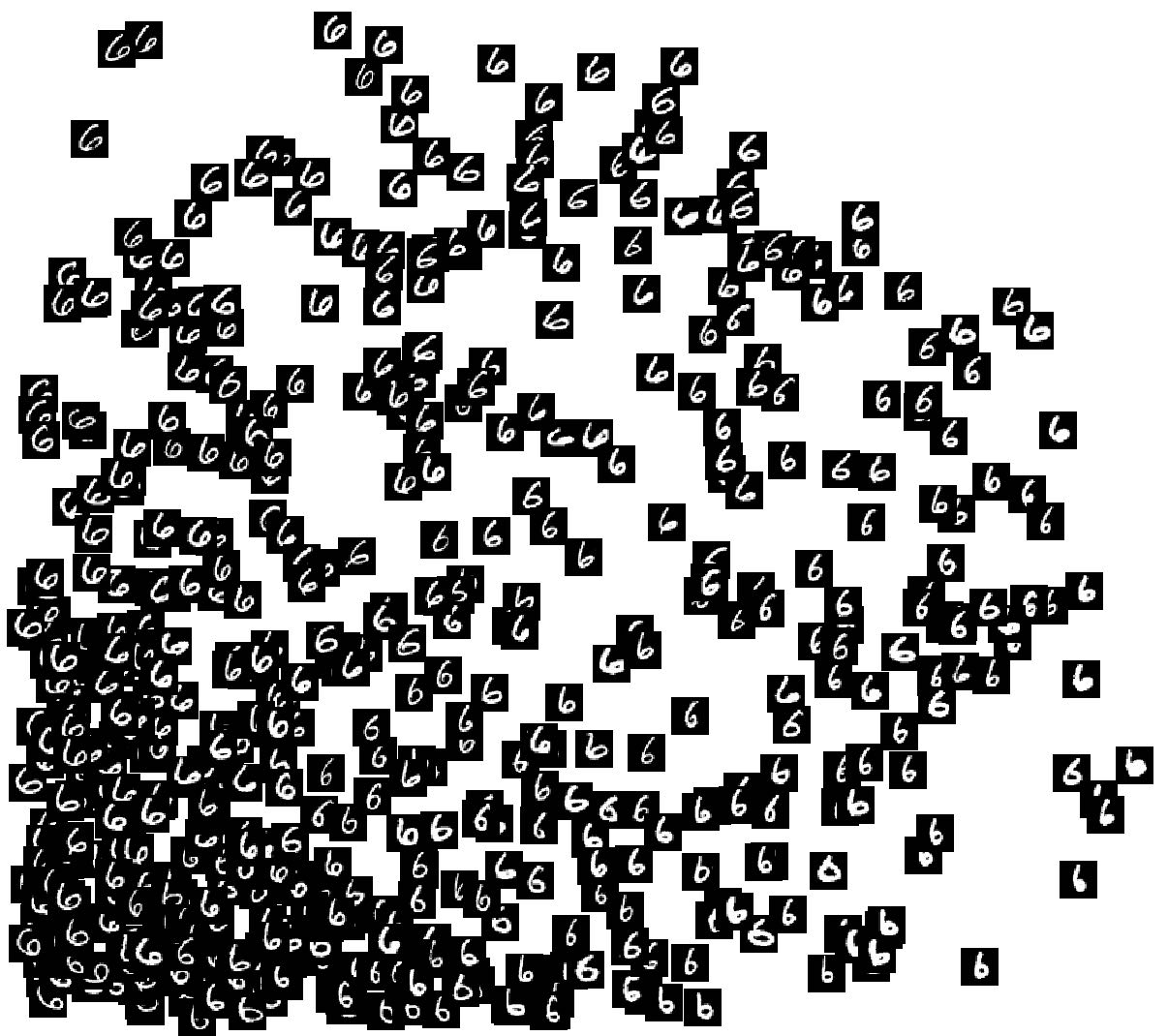} & \includegraphics[width=.25\textwidth]{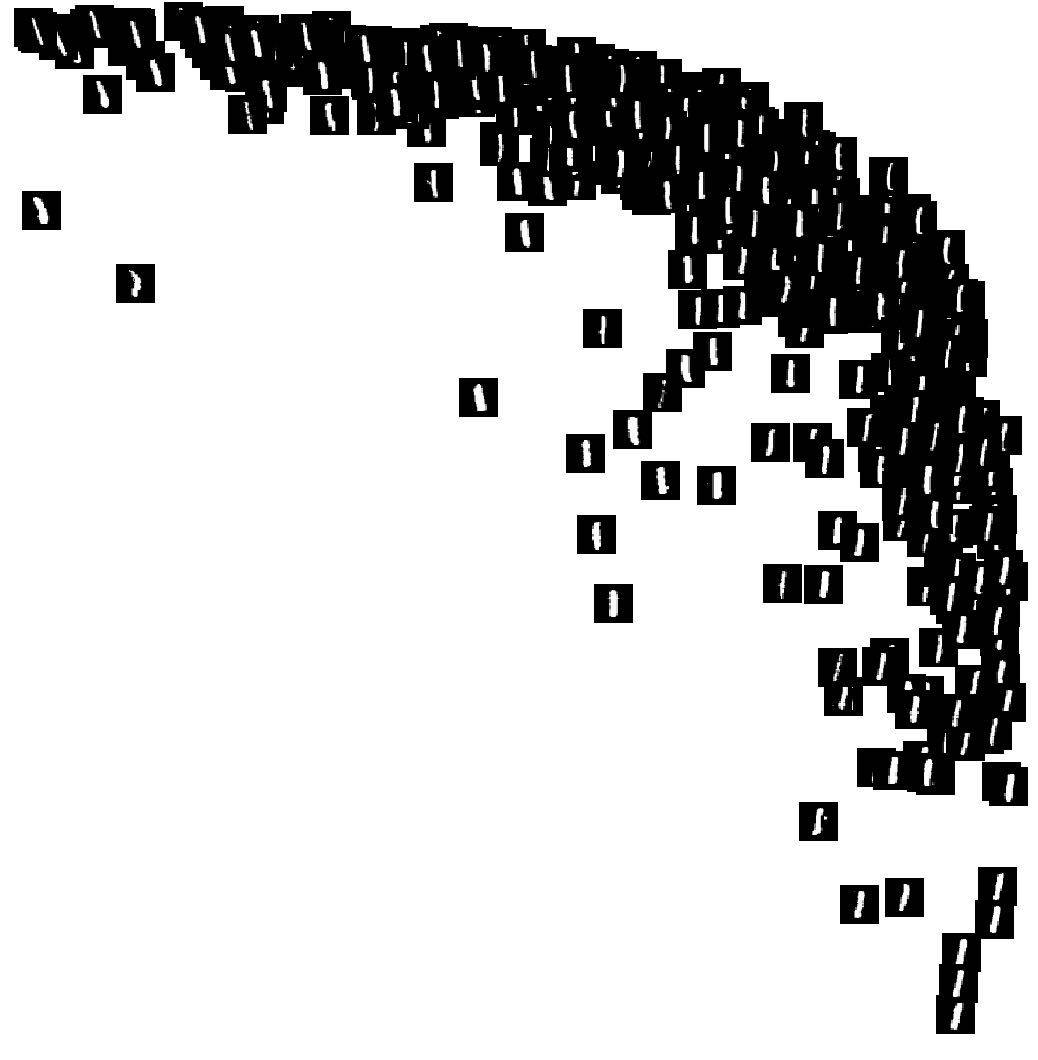}\\
\includegraphics[width=.25\textwidth]{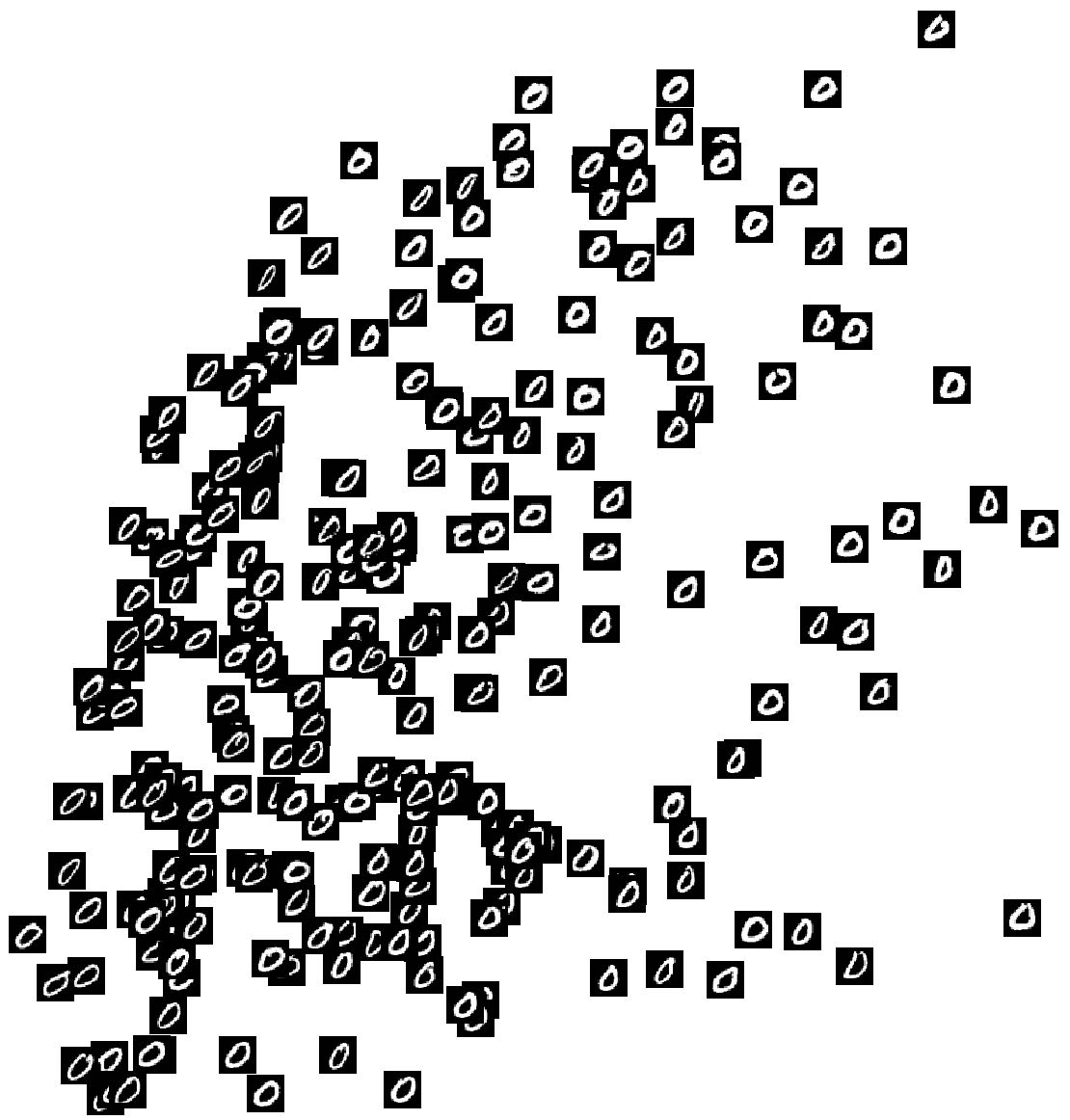} & \includegraphics[width=.25\textwidth]{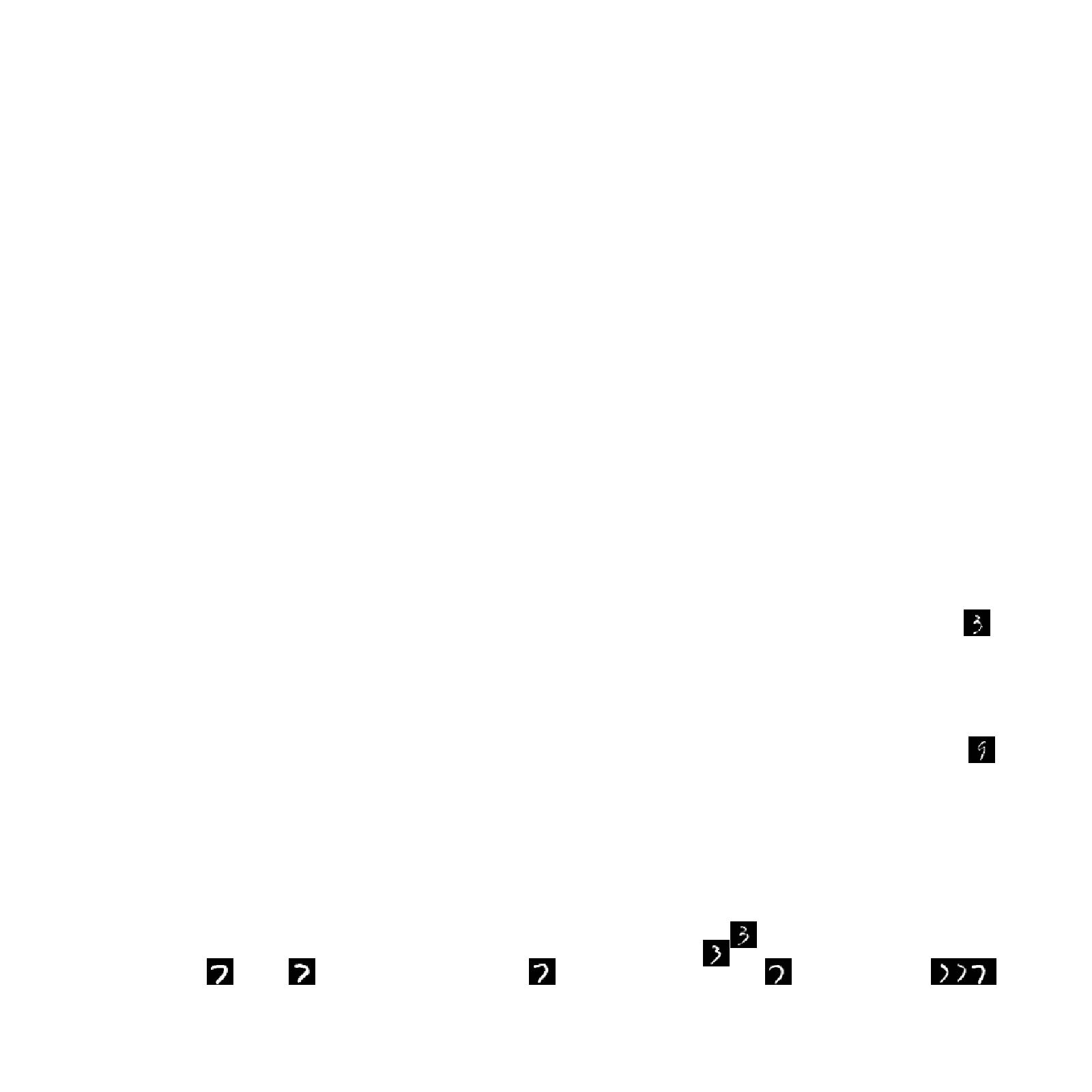} & \includegraphics[width=.25\textwidth]{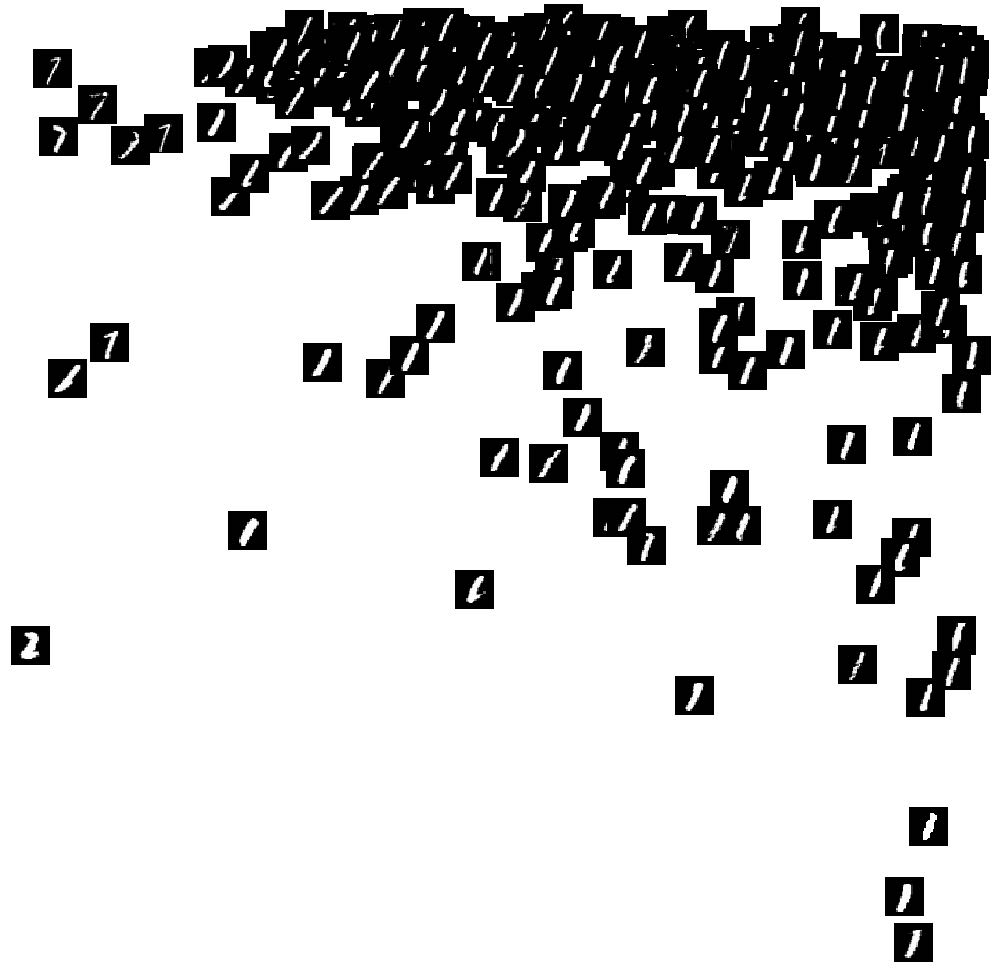} & \includegraphics[width=.25\textwidth]{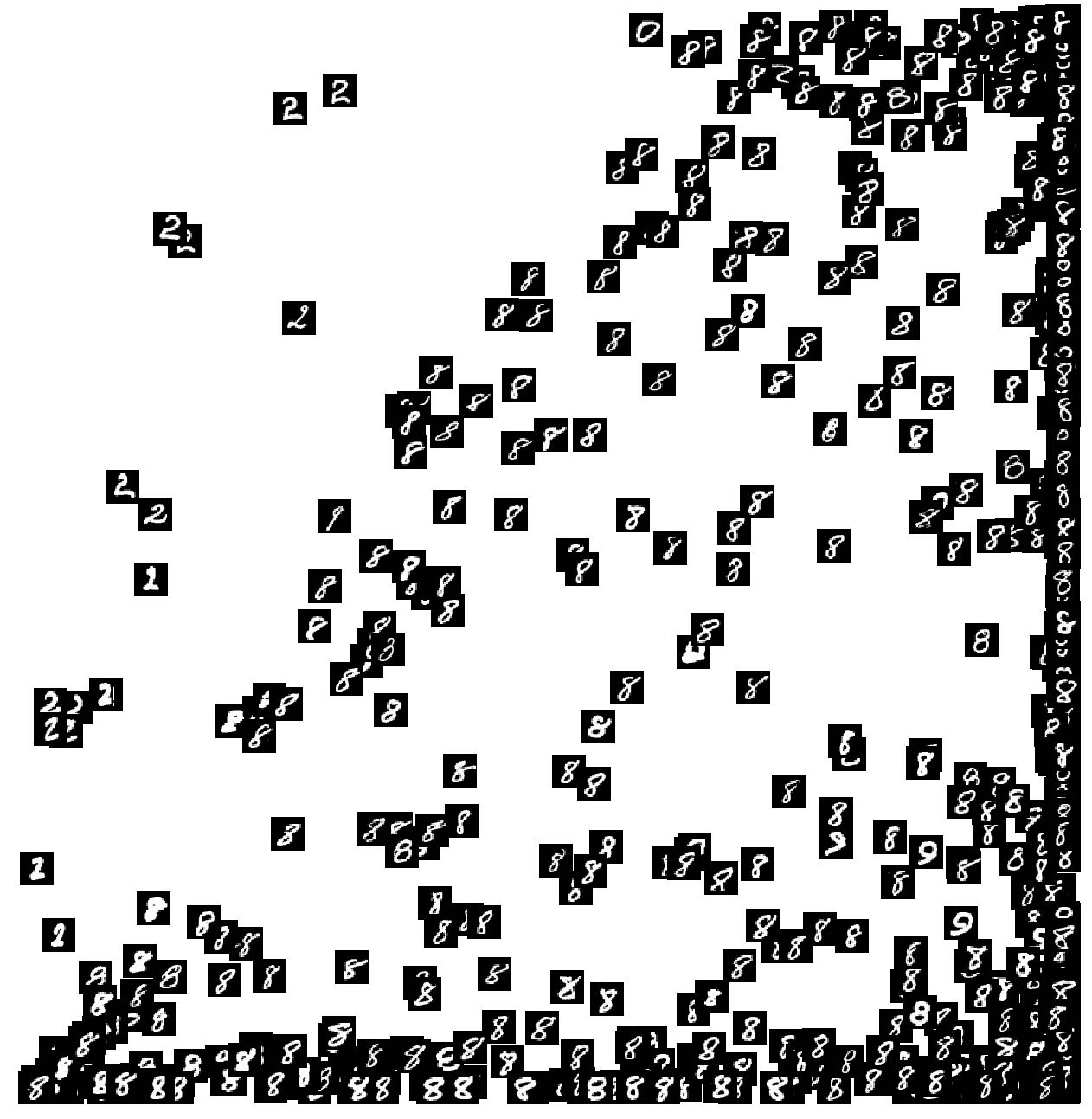}\\
\end{tabular}
\end{table}

\pagebreak

\subsection{Proof of proposition \ref{prop-mmd}} \label{prop-proof}
\begin{proof}
We breakdown the three terms defining $\MMD$ (\ref{mmd-def}). For the first term, we have:
\begin{align*}
\E_{(z_1, i), (z_2, j) \sim p\times p} & k_\Z((z_1, i), (z_2, j)) \\
&= \E_{x, \tilde x \sim p_{\text{data}} \times p_{\text{data}}} \sum_{i, j=1}^n q_i(x) q_j(\tilde x) k_\Z((\phi_i(x), i), (\phi_j(\tilde x), j)) \\
&= \E_{x, \tilde x \sim p_{\text{data}} \times p_{\text{data}}} \sum_{i=1}^n q_i(x) q_i(\tilde x) k_0(\phi_i(x), \phi_i(\tilde x)) \\
&\approx \frac{1}{N(N - 1)} \sum_{\substack{j, k = 1 \\ j \ne k}}^N \sum_{i=1}^n q_i(x_j) q_i(x_k) k_0(\phi_i(x_j), \phi_i(x_k)),
\end{align*}
where the first equality comes from (applying twice) the fact that
\begin{equation}
\E_{z, i \sim p} f(z, i) = \E_{x \sim p_{\text{data}}} \E_{z, i \sim q(\cdot | x)} f(z, i) = \E_{x \sim p_{\text{data}}} \sum_i q_i(x) f(\phi_i(x), i), \text{for any $f: \Z \to \R$} \label{z-to-x}
\end{equation}
and the approximation in the last line uses the U-statistic estimate
\begin{equation}
\E_{y, \tilde y \sim Q \times Q} h(y, \tilde y) \approx \frac{1}{N(N-1)} \sum_{\substack{j, k = 1 \\ j \ne k}}^N h(y_j, y_k) ~~ \text{for any $h:\X \times \X \to \R$} \label{U-stat},
\end{equation}
which holds for any distribution $Q$ and random sample $\{y_1, \ldots, y_N\}$  drawn from $Q$.

For the second term we have, letting $\U_{[0,1]^d}$ denote the uniform distribution on $[0, 1]^d$,
\begin{align*}
\E_{(z_1, i), (z_2, j) \sim p\times \U_\Z} & k_\Z((z_1, i), (z_2, j)) \\
&= \E_{x \sim p_{\text{data}}} \sum_{i=1}^n q_i(x) \E_{z_2, j \sim \U_\Z} k_\Z((\phi_i(x), i), (z_2, j)) \\
&= \E_{x \sim p_{\text{data}}} \sum_{i=1}^n q_i(x) \frac{1}{n} \E_{w \sim \U_{[0, 1]^d}} k_0(\phi_i(x), w) \\
&\approx \frac{1}{n N^2}\sum_{j, k = 1}^N \sum_{i=1}^n q_i(x) k_0(\phi_i(x_j), w_k),
\end{align*}
where the first equality again follows from (\ref{z-to-x}) and the approximation in the last line uses the standard expected value estimator applied to each factor. Finally for the third term we again use the estimate (\ref{U-stat}) to obtain
\begin{align*}
\E_{(z_1, i), (z_2, j) \sim \U_\Z\times \U_\Z} & k_\Z((z_1, i), (z_2, j)) \\
&= \E_{w \sim \U_{[0, 1]^d}} \E_{\tilde w \sim \U_{[0, 1]^d}} \sum_{i, j = 1}^n \frac{1}{n^2} k_\Z((w, i), (\tilde w, j)) \\
&= \E_{w \sim \U_{[0, 1]^d}} \E_{\tilde w \sim \U_{[0, 1]^d}} \frac{1}{n} k_0(w, \tilde w) \\
&\approx \frac{1}{n N(N-1)} \sum_{\substack{j, k = 1 \\ j \ne k}}^N k_0(w_j, w_k).
\end{align*}
Combining these gives the right hand side of (\ref{ellZ-def}).
\end{proof}

\end{document}